\documentclass{article} 
\PassOptionsToPackage{table}{xcolor} 
\usepackage{iclr2027_conference,times}

\usepackage{amsmath,amsfonts,bm}

\def\eqref#1{equation~\ref{#1}}

\def\1{\bm{1}}

\DeclareMathAlphabet{\mathsfit}{\encodingdefault}{\sfdefault}{m}{sl}
\SetMathAlphabet{\mathsfit}{bold}{\encodingdefault}{\sfdefault}{bx}{n}

\usepackage{hyperref}
\usepackage{url}
\usepackage{booktabs}
\usepackage{xspace}
\usepackage{pifont}
\usepackage{multirow}
\usepackage{graphicx}
\usepackage{wrapfig}
\usepackage[most]{tcolorbox}
\usepackage[T1]{fontenc}
\usepackage{subcaption}
\usepackage{adjustbox}
\usepackage{amsmath}
\usepackage{float}

\usepackage{xcolor}

\definecolor{humancolor}{HTML}{3F7FC6}   
\definecolor{agentcolor}{HTML}{D2762F}   
\definecolor{adoptcolor}{HTML}{8A6BBE}   
\definecolor{modifycolor}{HTML}{3D8FD1}  
\definecolor{rejectcolor}{HTML}{D96882}  
\definecolor{geminiprocolor}{HTML}{4C72B0}
\definecolor{geminiflashcolor}{HTML}{55A868}
\definecolor{gptcolor}{HTML}{C44E52}
\definecolor{claudecolor}{HTML}{DD8452}
\definecolor{qwenmoecolor}{HTML}{8172B3}
\definecolor{qwendensecolor}{HTML}{CCB974}
\definecolor{kolabblue}{HTML}{2B5C9E}
\definecolor{kolabred}{HTML}{A33A2B}
\definecolor{kolabpurple}{HTML}{6A4C93}
\definecolor{kolabgreen}{HTML}{2E8B57}
\definecolor{kolabpink}{HTML}{D81B60}
\definecolor{kolabteal}{HTML}{168A8A}
\definecolor{kolabyellow}{HTML}{B88700}
\colorlet{headerbg}{kolabblue!9}         
\colorlet{bestbg}{kolabblue!15}          
\colorlet{oursbg}{kolabyellow!10}         
\colorlet{baselinefg}{black!55}          

\definecolor{softlink}{HTML}{B0626B}  
\definecolor{softcite}{HTML}{5A86B0}  
\definecolor{softurl}{HTML}{8A72AE}   
\hypersetup{colorlinks=true, linkcolor=softlink, citecolor=black, urlcolor=softurl, pdfborder={0 0 0}}

\newcommand{\cmark}{\textcolor{kolabgreen}{\ding{51}}}
\newcommand{\xmark}{\textcolor{kolabred!80}{\ding{55}}}
\newcommand{\dataset}{\textsc{Kolab}\xspace}

\newif\ifshownotes
\shownotestrue
\ifshownotes
  \newcommand{\dk}[1]{\textcolor{blue}{[@DK: #1]}}
  \newcommand{\shirley}[1]{\textcolor{magenta}{\bf\small [#1 --Shirley]}}
  \newcommand{\note}[1]{\textcolor{magenta}{\small [Note: #1]}}
\else
  \newcommand{\dk}[1]{}
  \newcommand{\shirley}[1]{}
  \newcommand{\note}[1]{}
\fi

\newcommand{\adopt}{\textsc{Adopt}\xspace}
\newcommand{\modify}{\textsc{Modify}\xspace}
\newcommand{\reject}{\textsc{Reject}\xspace}
\newcommand{\adoptc}{\textcolor{adoptcolor}{\textbf{\textsc{Adopt}}}\xspace}
\newcommand{\modifyc}{\textcolor{modifycolor}{\textbf{\textsc{Modify}}}\xspace}
\newcommand{\rejectc}{\textcolor{rejectcolor}{\textbf{\textsc{Reject}}}\xspace}
\newcommand{\hact}{\textcolor{humancolor}{\textbf{H}}}
\newcommand{\aact}{\textcolor{agentcolor}{\textbf{A}}}

\newcommand{\best}[1]{\cellcolor{bestbg}\textbf{#1}}          
\newcommand{\sd}[1]{{\scriptsize$\pm$#1}}                     
\newcommand{\na}{\textcolor{black!35}{--}}                    
\newcommand{\mdot}[1]{\textcolor{#1}{\raisebox{0.1ex}{\scalebox{1.1}{$\bullet$}}}\hspace{0.3em}}
\newcommand{\mGemPro}{\mdot{geminiprocolor}Gemini 3.1 Pro}
\newcommand{\mGemFlash}{\mdot{geminiflashcolor}Gemini 3.8 Flash}
\newcommand{\mGPT}{\mdot{gptcolor}GPT-6 Astra}
\newcommand{\mClaude}{\mdot{claudecolor}Claude Opus 5.5}
\newcommand{\mQwenMoE}{\mdot{qwenmoecolor}Qwen3-VL-30B-A3B}
\newcommand{\mQwenDense}{\mdot{qwendensecolor}Qwen3-VL-32B}

\newtcolorbox{promptbox}[1][]{
  enhanced,
  breakable,
colback=kolabblue!3,
colframe=kolabblue!70!black,
colbacktitle=kolabblue!70!black,
coltitle=white,
fonttitle=\small\bfseries,
boxrule=0.6pt,
  arc=2mm,
  left=7pt, right=7pt, top=6pt, bottom=6pt,
  fontupper=\small,
  before upper={\raggedright},
  #1
}
\title{\dataset: Benchmarking Human-Agent Collaboration in Long-Horizon Workflows}

\author{Shirley Anugrah Hayati\textsuperscript{1}, Patricia Anugrah Setiani\textsuperscript{1}, Ruizi Wang\textsuperscript{1}, Zhouyi Li\textsuperscript{2},\\
\textbf{Khanh Chi Le\textsuperscript{1}, Young-Jun Lee\textsuperscript{1} \& Dongyeop Kang\textsuperscript{1}} \\
\textsuperscript{1}Department of Computer Science, \textsuperscript{2}Department of Economics\\
University of Minnesota\\
\texttt{\{hayat023, dongyeop\}@umn.edu} \\
}

\usepackage{enumitem}

\newcommand{\eg}{\textit{e.g.,}\xspace}

\iclrfinalcopy 
\begin{document}

\maketitle
\thispagestyle{plain} 
\pagestyle{plain} 

\begin{abstract}
When humans collaborate with AI agents on long-horizon tasks, what happens to the agents' contributions? For solving long-horizon tasks, humans place greater importance on ongoing interaction with AI agents (\eg revising plans) than on an agent's ability to complete the task in a single attempt. However, most agentic benchmarks have focused more on automation. Strong performance on such benchmarks does not imply effective real-world collaboration. Professional workflows offer a useful setting, as they often involve interdependent steps. To address this gap, we introduce \dataset, a new dataset of 115+ hours of screen recordings of 50 professionals from 40 occupations collaborating with agents on realistic, long-horizon problems. \dataset also contains participants' collaboration and output quality judgments. Our analysis reveals diverse collaboration patterns, highlighting the importance of when and how professionals coordinate with agents. A collaborative agent must do two things that automation benchmarks never test: anticipate whether experts will adopt, modify, or reject its contribution once it leaves the chat window to guide its next steps, and assess collaboration quality from the unfolding workflow without constant expert feedback. Our benchmark operationalizes both: (1) forecasting how professionals use agent contributions at diverse workflows, and (2) verifying their judgments of collaboration satisfaction and output quality. Findings show that frontier models, both agentic and non-agentic, overestimate collaboration with experts and struggle to distinguish between high- and low-quality collaboration. Longer workflow context does not consistently improve these predictions. As frontier agents cannot yet understand long-horizon collaboration with experts, we position \dataset as a valuable dataset for studying human–agent collaboration grounded in real professional workflows.\footnote{Code and data are available at \url{https://minnesotanlp.github.io/kolab/}}
\end{abstract}

\section{Introduction}

As AI agents become more capable of complex reasoning and knowledge work, recent research has examined whether they can complete such tasks autonomously in a single attempt, given a single instruction~\citep{patwardhan2026gdpval,vidgen2026apex,zhu2026workflow,sun2026agents,merrill2026terminal,xu2026theagentcompany,xie2024osworld}. However, real-world problem solving may not follow this autonomous setting: human goals can evolve, earlier decisions shape subsequent work, and errors may require intervention before they propagate \citep{tack2026llms}. In professional work, for example, workers formulate requirements, produce intermediate artifacts, revise decisions, and move between agents and domain-specific tools. Thus, strong autonomous task performance may not translate into effective human-agent collaboration \citep{zou2025survey,mozannar2025magentic,shen-etal-2026-scaling}. Despite the need for studying effective human-agent collaboration, existing datasets provide limited grounding in real collaborative workflows. They are often not publicly released, rely on predefined tasks constructed or sourced by researchers, or focus on a single professional domain such as software development \citep{wang2025ai,chen26icml,shao2026collabskill}. As a result, benchmarking long-horizon human-agent collaboration workflows remains underexplored. 

\begin{figure}
    \centering
    \includegraphics[width=\linewidth]{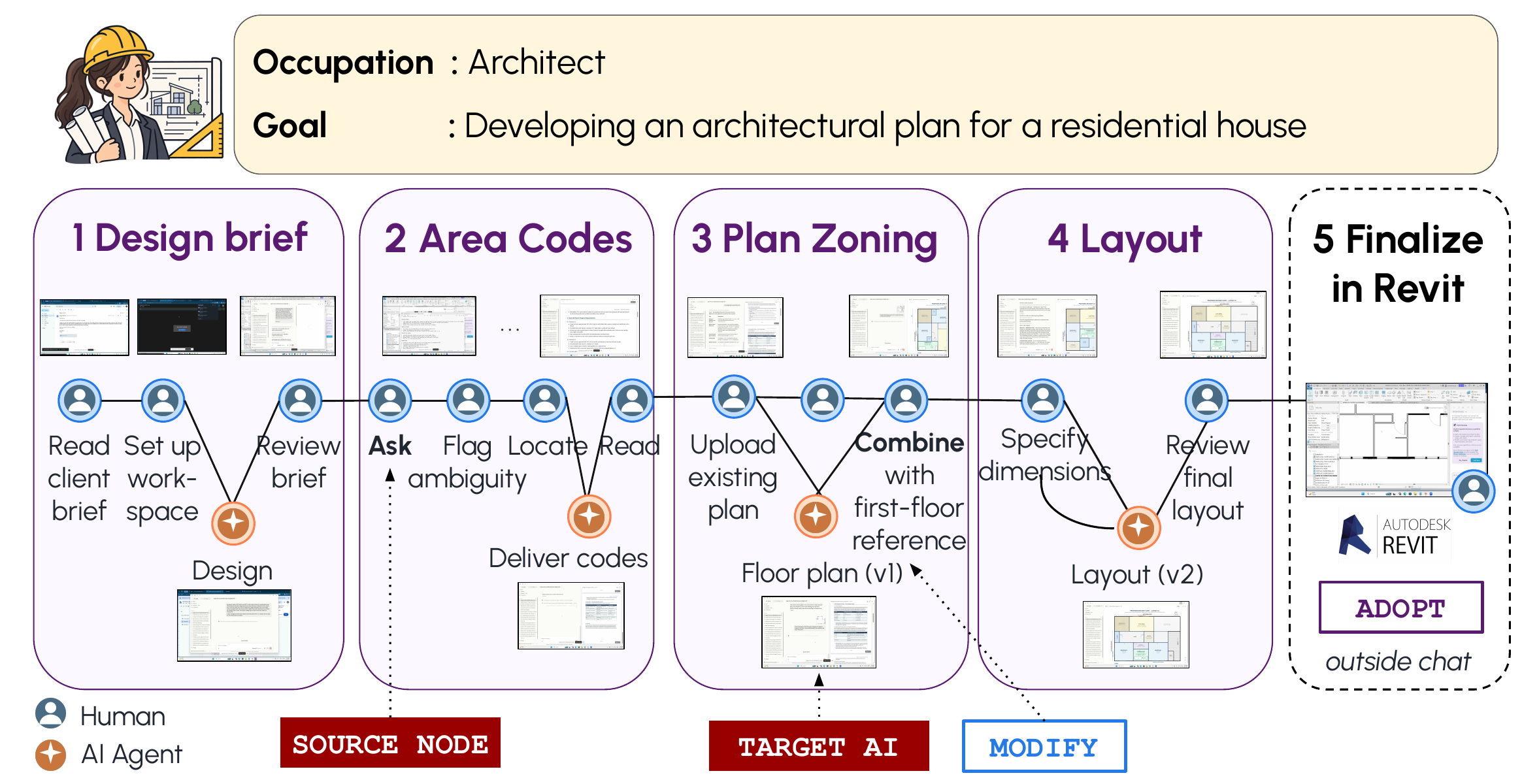}
\caption{An example human-agent collaborative workflow for an architect. Each step is a subgraph of \textcolor{humancolor}{human} and \textcolor{agentcolor}{agent} actions; annotations link a source node, the target AI output it triggers, and the human response to it (here, \modify and \adopt).}
\vspace{-3mm}
    \label{fig:main_figure}
    \vspace{-10pt}
\end{figure}

To address this gap, we introduce \dataset, a dataset of screen recordings capturing human-agent collaboration on long-horizon professional problems. We represent each workflow as a graph of semantically meaningful \textcolor{humancolor}{human} and \textcolor{agentcolor}{agent} actions and their dependencies (Figure~\ref{fig:main_figure}), allowing us to understand when and how collaboration occurs in the workflow. In this example, an architect completes a long-horizon task in five steps, and agent outputs can influence later steps. A task delegated in Step~2 (source node) produces information (target AI node) that is later modified and used in Step~3 (\modify). Human responses to agent outputs may also occur outside the chat interface. The architect ultimately adopts and executes the resulting plan in Revit, an action not directly observable by the agent.

Our analysis reveals that human-agent collaboration workflow varies in when and how professionals engage with AI agents. Burstier workflows are associated with longer periods of independent work before interacting with AI, suggesting that professionals may organize their collaboration around periods of independent work rather than continuous AI interaction. Workflows with more concurrent human-agent activities are associated with lower collaboration satisfaction, suggesting that parallel activities may introduce additional coordination costs.

Beyond characterizing collaboration, we ask whether frontier multimodal large language models (MLLMs) and agents can \emph{understand their own place} in a professional's workflow. We frame this as two capabilities that a deployed collaborative agent needs but that autonomous benchmarks never measure.

\textbf{Anticipating the fate of a contribution (RQ1).} In long-horizon work, an agent's output is rarely the end of the story: it is adopted, modified, or rejected, often outside the chat interface (\eg an architect incorporating an agent's plan in Revit). An agent that can \emph{recognize} what happened to its output after the fact (1a) could turn naturalistic screen recordings into feedback signals for training and for auditing deployed agents, without asking the expert to label anything. An agent that can \emph{forecast} this response at delivery time (1b)---before the human reacts---could act proactively: asking a clarifying question when rejection is likely, flagging the parts it expects to be edited, or escalating to the human when its contribution carries risk that would propagate to later steps~\citep{oh2026proactive}. Because such forecasts depend on how much of the collaboration the agent remembers, we vary the available history from the preceding steps (1c) to the entire workflow so far (1d), which directly informs how much context a long-running or computer-use agent must retain to model its collaborator.

\textbf{Judging the collaboration itself (RQ2).} Expert ratings of collaboration satisfaction (2a) and output quality (2b) are the ground truth for whether human-agent collaboration works, yet they are expensive to collect and unavailable at deployment. If an MLLM could recover these ratings from the workflow alone, it could serve as a process-level reward model for training collaborative agents, as an automatic monitor that detects friction or low-quality outcomes in deployed systems, and as a scalable evaluator for professional tasks where no reference answer exists. Crucially, such a verifier must separate genuinely successful interactions from dissatisfying ones, rather than echoing the predominantly positive ratings.

We observe that frontier multimodal agents still struggle to anticipate how professionals use agent outputs and to recover their judgments of collaboration quality. For example, Gemini 3.1 Pro and GPT-6 achieve only 39.1\% and 51.3\% F1 in forecasting adoption behavior, while their predictions of collaboration satisfaction correlate weakly with professional ratings (Spearman's $\rho=.08$ and $.38$). Providing longer workflow context or additional modalities does not consistently improve these predictions. Further analysis reveals systematic biases: models often favor certain adoption labels and overpredict high ratings, limiting their ability to distinguish successful from poor collaboration. These findings suggest that current models may rely on their own prediction tendencies rather than reliably understanding professionals' behaviors and judgments from the workflow. These results highlight an important challenge for developing collaborative agents that can anticipate human decisions and adapt to how their contributions are actually used. 

Our main contributions are as follows.
\begin{itemize} [leftmargin=2em, itemsep=3pt, parsep=0pt, topsep=1pt]
\setlength{\itemsep}{1pt}
\setlength{\parskip}{0pt}
\setlength{\parsep}{0pt}
\setlength{\topsep}{2pt}
\item We present \dataset, a new dataset of 115 hours of screen-recorded human-agent workflows from 50 professionals solving their own real-world problems, comprising 508 steps across 90+ softwares. Each workflow includes a final deliverable with step-level human ratings of collaboration experience and output quality.
\item We analyze human-agent collaboration dynamics using three metrics: collaboration burstiness, workflow topology, and AI retention rate.
\item We benchmark frontier MLLMs on their ability to understand human-agent collaborative workflows when solving professional problems. 
\item We uncover systematic MLLM failure modes, revealing prediction biases that vary by model and input modality.
\end{itemize}

\section{Related Work}
\label{related_work}
\begin{table}[t]
\centering
\small
\caption{Comparison of \dataset{} with long-horizon, multi-domain workflow benchmarks. ``H-AI Collab.'' indicates the presence of human-agent collaborative workflows; ``Software Diversity'' refers to tasks spanning multiple software applications; ``Modalities'' lists types of publicly released data.}
\setlength{\tabcolsep}{3pt}

\begin{tabular}{l c c c c l}
\toprule
\rowcolor{headerbg}
\textbf{Dataset} & \shortstack{\textbf{Human}\\\textbf{Workflow}} & \shortstack{\textbf{H-AI}\\\textbf{Collab.}} & \shortstack{\textbf{Software}\\\textbf{Diversity}} & \textbf{\#Domains} & \textbf{Modalities} \\
\midrule
Agents' Last Exam \scriptsize\citep{sun2026agents} & \xmark & \xmark & \cmark & 55 & Image, text, files \\
Workflow-GYM \scriptsize\citep{zhu2026workflow} & \cmark & \xmark & \cmark & 6 & Image \\
WeaveBench \scriptsize\citep{li2026weavebench} & \xmark & \xmark & \cmark & 8 & Image, text \\
TheAgentCompany \scriptsize\citep{xu2026theagentcompany} & \xmark & \xmark & \xmark & 6 & Text \\
GDPVal \scriptsize\citep{patwardhan2026gdpval} & \xmark & \xmark & \cmark & 44 & Text, files \\
DeskCraft \scriptsize\citep{wang2026deskcraft} & \xmark & \cmark & \cmark & 5 & Image \\
\midrule
\rowcolor{oursbg}
\textbf{\dataset{} (Ours)} & \cmark & \cmark & \cmark & \textbf{40} & \textbf{Video}, text, files\\
\bottomrule
\end{tabular}
\vspace{-10pt}
\label{tab:dataset-comparison}
\end{table}

\paragraph{Long-horizon agent and workflow benchmarks.}
Autonomous agent benchmarks cover desktop applications \citep{xie2024osworld,yuan2026osworld20benchmarkingcomputeruse}, enterprise software \citep{dai2026scuba}, hybrid GUI/CLI and scientific computer use \citep{li2026weavebench, jang2026mypcbench, wang2026deskcraft, zhu2026workflow,sun2026scienceboard}, professional deliverables \citep{patwardhan2026gdpval,sun2026agents,xu2026theagentcompany}, and long-running agent behavior \citep{maldaner2026sentinelbench}. Some include human baselines but do not release human action trajectories. Workflow-GYM \citep{zhu2026workflow} collects expert workflow trajectories but uses them as demonstrations to evaluate autonomous agents. Human-only process datasets such as ScholaWrite~\citep{le-etal-2026-scholawrite} and LawFlow~\citep{das2025lawflow} capture expert work in a single domain (scholarly writing and legal reasoning) without agent collaboration. Among the benchmarks in Table~\ref{tab:dataset-comparison}, \dataset uniquely combines multimodal trajectories of professionals collaborating with agents on self-defined problems with step-level human ratings.

\paragraph{Human-agent collaborative workflows.} Prior work argues that autonomous task performance does not ensure effective collaboration \citep{shen-etal-2026-scaling}, and that the role humans play relative to an AI system shapes the interaction \citep{wan2022user}; indeed, varying LLM access measurably changes how people work \citep{christenson2026effects}. Existing evaluations often rely on simulated users \citep{wu2026has,yao2024tau,vijayvargiya2026ambig,qian2025userbench} or human intervention in short shopping and web browsing tasks \citep{huq2025cowpilot,mozannar2025magentic}. Collaborative Gym \citep{shao2026collaborative} supports real or simulated users but covers only three predefined tasks. CollabSkill \citep{shao2026collabskill} assigns professionals problems from existing datasets; \citet{chen26icml} study interaction-level satisfaction with software-engineering agents; and \citet{wang2025ai} compare separate human and agent workflows. Our study setup is most similar to that of \citet{hayati2026orchestratinggenaiinterdisciplinaryresearch} who investigate human-agent collaboration in interdisciplinary research using only chat logs. Meanwhile, our work also captures the complete workflow, including how professionals act on agent contributions in their work that may happen outside the chat interface.

\section{\dataset: A Multimodal, Multidomain Dataset of Human-Agent Collaborative Workflows on Long-Horizon Problems}

\label{data_collection}
\dataset is built to understand human-agent collaborative workflows when humans solve long-horizon problems. 

\textbf{Problem criteria.}
To ensure that the collected workflows represent substantive professional problem solving rather than generic computer-use activities, we required each problem to satisfy four criteria:

\begin{itemize}[leftmargin=2em, itemsep=1pt, parsep=0pt, topsep=1pt]
    \item \textbf{Real-world.}
    The problem must originate from the professional's actual work. Professionals were not given tasks from existing workflow datasets or scenarios.
    
    \item \textbf{Domain-specific.}
    The problem must require the application of professional knowledge, skills, or judgment from the participant's field. Tasks that could be completed without domain expertise, such as routine personal activities or generic information seeking, were excluded.
    
    \item \textbf{Complex and long-horizon.}
    The problem must require a sequence of interdependent steps to achieve its final goal. Participants were asked to plan a workflow requiring at least one hour of work and consisting of at least five steps, with each step involving multiple computer actions or decisions.

    \item \textbf{Deliverable-oriented.}
    The problem must result in a concrete final deliverable. Each step must also produce an intermediate output that can be evaluated.
\end{itemize}

\begin{wrapfigure}[14]{r}{0.48\textwidth}
    \centering
    \vspace{-5mm}
\includegraphics[width=\linewidth]{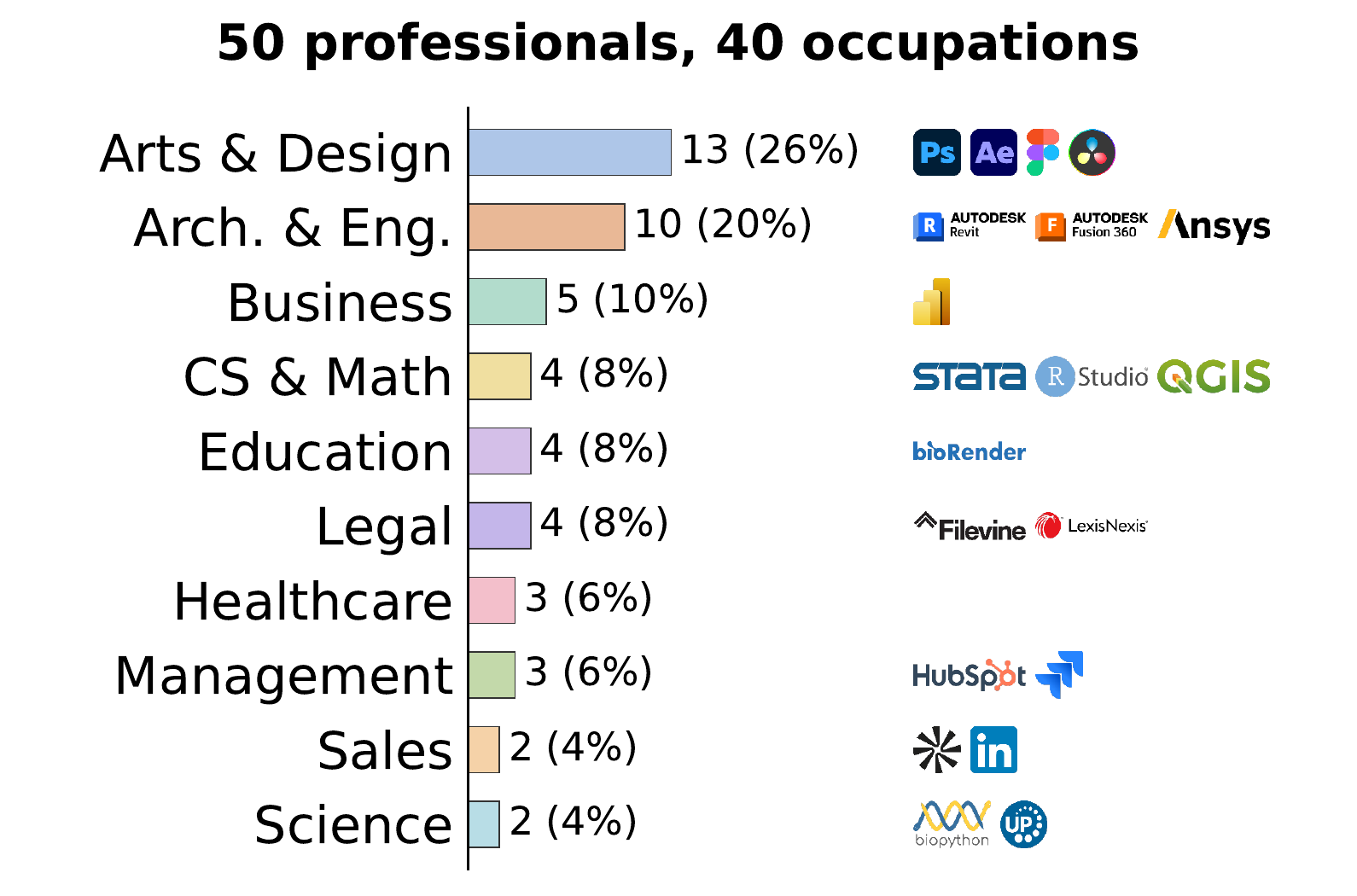}
\vspace{-3mm}
    \caption{\dataset domain distributions.}
    \label{fig:occupations}
    \vspace{-3mm}
\end{wrapfigure}
\textbf{Data collection.} We recruited 50 industry professionals via word of mouth, Prolific, and Upwork. Each participant received a compensation of \$100 with a bonus for extra time. Figure~\ref{fig:occupations} shows our participants' domains with details in Appendix~\ref{appendix:professional_background}. Professionals interacted with an agentic AI, Claude~\citep{anthropic2026claude}, for two 1-hour recording sessions.\footnote{They were free to use other AI agents when appropriate for their work. For example, an architect might use Claude for planning and reasoning while using ChatGPT to generate or render architectural images.} Each participant planned their workflow beforehand to ensure that their problem satisfied our criteria. Then they updated the workflow document, rated their collaboration experience and output quality, and filled out the post-interaction survey in the Appendix \ref{sec:post-interaction}.

\textbf{Dataset statistics.} \dataset covers 100 problem sets and human-agent collaborative workflows conducted by professionals from 10 sectors of 40 occupations (Figure~\ref{fig:occupations}).
The post-interaction survey also measured collaboration satisfaction and output quality on a 1 (low) to 5 (high) scale.\footnote{Details in Table~\ref{tab:rating-distribution} in the Appendix.} Collaboration satisfaction was rated 5 in 54.9\% of sessions and 4 in 29.7\%, with 15.3\% rated 3 or below. Similarly, output quality was rated 5 in 53.5\% of sessions and 4 in 31.0\%.


\textbf{Data release.} We publicly release \dataset with seven components: (1) screen-recording videos for all sessions; (2) the workflow documents written by the professionals, including goal, context, and step names/actions; (3) the human–agent chat logs, including agent tool calls; (4) optional input and output files of each workflow; (5) human ratings for collaboration experience and output quality, along with step-level output rubrics written by professionals; (6) the workflow graph per session, with step alignment, timestamps, and adoption labels; and (7) post-session survey responses.

\begin{table}[t]
\centering
\caption{Real examples of \adopt, \modify, and \reject responses from \dataset, drawn from a music producer workflow, architecture workflow, and a video-production workflow respectively.}
\label{tab:response-examples}
\resizebox{\linewidth}{!}{%
\begin{tabular}{l p{4.6cm} p{4.6cm} p{4.8cm} p{1.7cm}}
\toprule
\rowcolor{headerbg}
\textbf{Response} & \textbf{Source Node Action} & \textbf{Target AI Node Action} & \textbf{Current Node Action} & \textbf{Software} \\
\midrule
\multirow{2}{*}{\adoptc}
& Request Afrobeats/techno background music in ElevenLabs (67:33--68:30)
& ElevenLabs generates Afrobeats instrumental candidates (69:50--70:04)
& Capture the ElevenLabs track and import it into the DaVinci Resolve project (72:05--74:36)
& ElevenLabs, DaVinci Resolve \\ \\

\midrule
\multirow{2}{*}{\modifyc}
& Upload the existing shell plan and request a full room layout (24:10--25:28)
& Claude delivers a dimensioned floor-plan layout for the lobby/storage circulation (51:21--56:20)
& Rebuild walls/partitions in Revit, keeping the liked zoning but fixing lobby/storage circulation (58:40--72:59)
& Revit \\
\midrule
\multirow{2}{*}{\rejectc}
& Ask Kling AI to generate a Zanzibar schoolyard video clip (36:30--39:04)
& Kling AI generates the requested video clip (39:45--41:17)
& Judge the clip's audio unusable and abandon it, out of credits to iterate further (41:17--42:06)
& Kling AI \\
\bottomrule
\end{tabular}%
}
\end{table}

\begin{table}[t]
\centering
\caption{Summary statistics of \dataset for videos, workflow graphs, and human action annotations.}
\label{tab:dataset-statistics}
\small
\setlength{\tabcolsep}{5pt}
\begin{tabular}{@{}l r @{\hspace{16pt}} l r @{\hspace{16pt}} l r@{}}
\toprule
\multicolumn{2}{c}{\textcolor{kolabpink}{\textbf{Video}}} &
\multicolumn{2}{c}{\textcolor{kolabpink}{\textbf{Graph}}} &
\multicolumn{2}{c}{\textcolor{kolabpink}{\textbf{Annotation}} {\scriptsize(692 labels)}} \\
\cmidrule(r{6pt}){1-2} \cmidrule(lr{6pt}){3-4} \cmidrule(l){5-6}
Total duration & 115 hrs & Workflow steps & 508 & \adopt & 65.9\% (456) \\
Avg. duration/session & 69 mins & Avg. steps/session & 5.24 & \modify & 19.1\% (132) \\
 & & Avg. nodes/step & 8.68 & \reject & 15.0\% (104) \\
\bottomrule
\end{tabular}
\vspace{-10pt}
\end{table}




\section{Understanding Human-Agent Collaboration Dynamics}
\subsection{Data Processing and Annotation}
To make the workflow structured, each step in the workflow is represented as a temporal graph, where each node $v \in V$ corresponds to a human or agent action at a time interval $
\tau(v) = [t_s(v), t_e(v)] \subset \mathbb{R}_{\geq 0}$. The research team and professionals refined these graphs together. An example graph for one step is shown in Figure~\ref{fig:p4_graph} in the Appendix. A node can branch when activities occur at overlapping timestamps and loops happen when the same action is repeated. Formally, each workflow is represented as follows:
\begin{equation}
G = (V, E, a, w, \tau),
\end{equation}

where $V$ is a set of human (\hact) or AI (\aact) action nodes; $E \subseteq V \times V$ is the set of directed edges, with $(u, v) \in E$ indicating that action $v$ directly follows from action $u$. 

After graph construction, two human annotators labeled how humans incorporated agent outputs. A human node is identified as a \emph{source node} if its action triggered the agent's generation (\eg ``\textit{design a floor plan}''). An agent node is the \emph{target AI node} if it generates an output that is subsequently acted on by the participant. The human node that acts on this AI output is labeled as follows.
\begin{itemize}[leftmargin=2em, itemsep=1pt, parsep=0pt, topsep=2pt]
\item \adoptc: The agent output is observably incorporated into the workflow as-is, such as directly using an agent-generated design or adding a suggested paper to a literature review.

\item \modifyc: The agent output is incorporated with modifications, \eg the human revises agent-generated text before using it.

    \item \rejectc: The agent output is not incorporated after being reviewed, such as when the human discards it or clearly expresses a negative sentiment and proceeds with a different approach.
\end{itemize}

As seen from real examples in Table~\ref{tab:response-examples}, professionals evaluate agent contributions and act on softwares outside the AI interface. A summary of \dataset is provided in Table~\ref{tab:dataset-statistics}. In total, \dataset contains 100 workflow videos. Each video has corresponding human-curated temporal graphs annotated with AI adoption labels, yielding a total of 692 instances for benchmarking MLLMs.

\subsection{Collaboration Dynamics Metrics}
\textbf{Collaboration burstiness.}
Human-agent interactions are not uniformly distributed over time. Humans may work with an agent steadily or in short, intensive episodes separated by longer periods of independent work. To capture this temporal rhythm, following \citet{goh2008burstiness}, we estimate collaboration burstiness as
$\displaystyle B = \frac{\sigma_\tau - \mu_\tau}{\sigma_\tau + \mu_\tau} \in [-1,1]$,
where $\mu_\tau$ and $\sigma_\tau$ are the mean and standard deviation of the inter-event intervals. $B \to 1$ indicates bursty interactions, $B \to -1$ indicates regular interactions, and $B = 0$ corresponds to a Poisson-like process.

\textbf{Workflow topology.}
Motivated by graph-theoretic representations of complex systems \citep{bondy2008graph,boccaletti2006complex,newman2011structure}, we characterize workflow topology to capture the extent to which humans work concurrently while the agent generates an output. We operationalize this as the proportion of \emph{branch} nodes with out-degree $>1$ ($Br$) relative to the number of edges ($|E|$) as a workflow topology $W = \frac{Br}{|E|}$.

\textbf{AI retention rate.} To measure how agent outputs are retained in subsequent work \citep{sole2026large}, we define AI retention through adoption and modification rates:
\[
R_A = \frac{N_{\adopt}}{N_{\adopt}+N_{\modify}+N_{\reject}},
\qquad
R_M = \frac{N_{\modify}}{N_{\adopt}+N_{\modify}+N_{\reject}}.
\]
These capture the proportion of agent outputs incorporated directly or with modification.


\begin{figure}[t]
    \centering
    \begin{subfigure}[b]{0.32\linewidth}
        \centering
        \includegraphics[width=\linewidth]{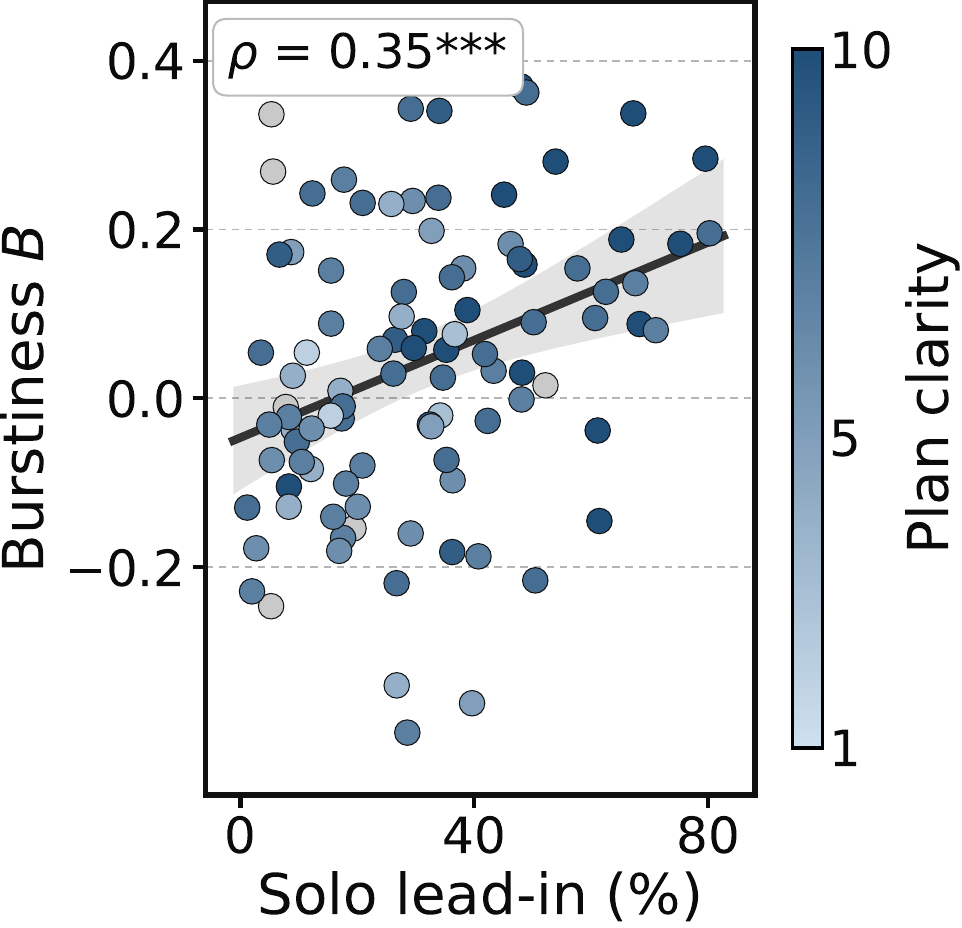}
        \caption{Collaboration burstiness}
        \label{fig:burstiness}
    \end{subfigure}
    \hfill
    \begin{subfigure}[b]{0.32\linewidth}
        \centering
        \includegraphics[width=\linewidth]{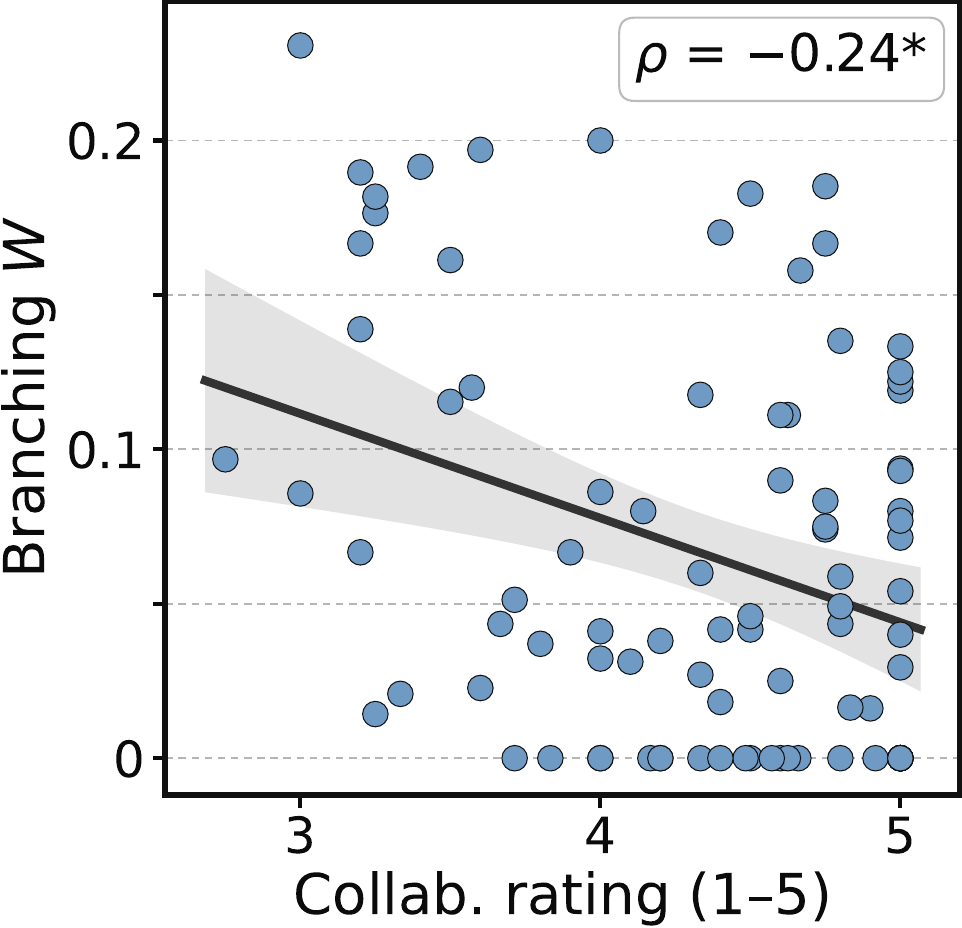}
        \caption{Workflow topology}
        \label{fig:branching}
    \end{subfigure}
    \hfill
    \begin{subfigure}[b]{0.32\linewidth}
        \centering
        \includegraphics[width=\linewidth]{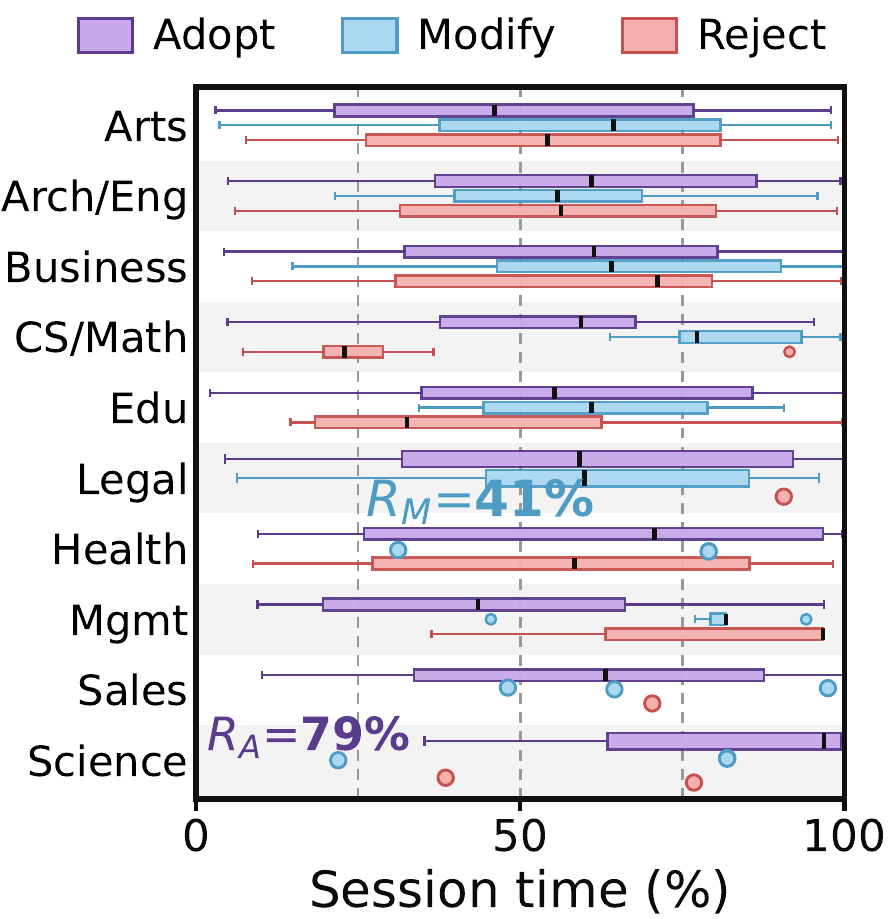}
        \caption{AI retention rate}
        \label{fig:adoption}
    \end{subfigure}
\caption{Collaboration dynamics in \dataset. (a) Burstiness vs.\ solo lead-in; (b) workflow branching vs.\ collaboration rating; (c) timing of \adopt, \modify, and \reject responses across sectors. *** refers to $p < 0.001$ and * is $p < 0.05$.}
\label{fig:three-metrics}
\vspace{-10pt}
\end{figure}

\subsection{Analysis}
Using these three metrics, we analyze how human-agent collaboration varies in its temporal rhythm ($B$), topology of concurrent work ($W$), and integration of agent contributions ($R$). Our analyses are intended to describe the collaborative patterns observed in \dataset, rather than as population-level comparisons across professional domains. Figure~\ref{fig:three-metrics} shows the results.

\textbf{\#1: Burstier workflows reflect more planning before interacting with the agent.}
Collaboration burstiness was relatively consistent within individuals across two sessions (Spearman $\rho=0.59$). We define solo lead-in as the proportion of a step completed before the first AI interaction, averaged over all steps in a session. This reflects the proportion of time a person spends working independently before first interacting with AI. For example, if the first AI interaction occurs 4 minutes into a 10-minute step, the solo lead-in is 40\%. As shown in Figure~\ref{fig:burstiness}, burstier workflows had longer solo lead-ins ($\rho=0.35$, $p<0.001$). Consistent with this pattern, participants who reported having a clearer plan before using AI in the post-interaction survey also had longer solo lead-ins ($\rho=0.53$, $p<0.001$).

\textbf{\#2: Workflows with branches are associated with lower perceived collaboration experience.}
Workflow branching ($W$) is negatively associated with collaboration ratings (Figure~\ref{fig:branching}) ($\rho=-0.24$, $p<0.05$). Although modest, this association suggests that concurrent human and agent activity may introduce coordination costs.

\textbf{\#3: AI retention patterns vary across professional domains.}
As shown in Figure~\ref{fig:adoption}, most domains predominantly adopt agent outputs as-is, with adoption rates ranging from 57\% in Legal to 79\% in Sciences and Computer \& Mathematical work. Legal workflows stand out for modification: 41\% of agent outputs were modified, about twice the rate of any other domain (Arts \& Sports, the next highest, is 22\%), consistent with lawyers’ reluctance to adopt AI outputs as-is \citep{schwarcz2026thinking}. This shows work style variation in how professional domains integrate agent contributions.

\section{Experiments and Results}
\label{experiments}
\subsection{Benchmark Tasks} Our benchmark tasks aim to answer the following research questions as summarized in Table \ref{tab:benchmark-tasks}.
\begin{itemize}[leftmargin=2em, itemsep=1pt, parsep=0pt, topsep=2pt]
\item \textcolor{kolabpink}{\textbf{RQ1:} (AI Adoption)} How well can frontier MLLMs forecast how humans use agent contributions in long-horizon professional workflows?

\item \textcolor{kolabpink}{\textbf{RQ2:} (Human Rating)} How well can frontier MLLMs predict human ratings of collaboration satisfaction and output quality given long-horizon human-agent collaboration workflows?
\end{itemize}

\begin{table}[t]
\centering
\caption{Benchmark tasks.
$T_x^{\mathrm{start}}$ and $T_x^{\mathrm{end}}$ denote the start and end timestamps of event or segment $x$, respectively; $x$ can refer to the $\mathrm{source}$ node, target $\mathrm{AI}$ node, human $\mathrm{adoption}$ node, or the full $\mathrm{workflow}$.}
\label{tab:benchmark-tasks}

\setlength{\tabcolsep}{4pt}
\renewcommand{\arraystretch}{1.08}
\begin{tabular}{l l c l}
\toprule
\textbf{Experiment Task} &
\shortstack{\textbf{Observed}\\\textbf{Workflow}} &
\shortstack{\textbf{Response}\\\textbf{Visible?}} &
\textbf{Prediction} \\
\midrule
\multicolumn{4}{l}{\textcolor{kolabpink}{\textbf{RQ1: AI adoption}}} \\
\textcolor{kolabpink}{\textbf{1a:}} Recognition
& $T_{\mathrm{source}}^{\mathrm{start}} \rightarrow T_{\mathrm{adoption}}^{\mathrm{end}}$
& \cmark
& \multirow{4}{*}{\shortstack[l]{\adoptc\\\modifyc\\\rejectc}} \\

\textcolor{kolabpink}{\textbf{1b:}} Forecasting
& $T_{\mathrm{source}}^{\mathrm{start}} \rightarrow T_{\mathrm{AI}}^{\mathrm{end}}$
& \xmark
& \\

\textcolor{kolabpink}{\textbf{1c:}} Step-$N$ forecasting
& $T_{\mathrm{step}}^{\mathrm{start}} \rightarrow T_{\mathrm{AI}}^{\mathrm{end}}$
& \xmark
& \\

\textcolor{kolabpink}{\textbf{1d:}} Workflow-level forecasting
& $T_{\mathrm{workflow}}^{\mathrm{start}} \rightarrow T_{\mathrm{AI}}^{\mathrm{end}}$
& \xmark
& \\

\midrule
\multicolumn{4}{l}{\textcolor{kolabpink}{\textbf{RQ2: Human rating}}} \\
\textcolor{kolabpink}{\textbf{2a:}} Collaboration satisfaction
& $T_{\mathrm{step}}^{\mathrm{start}} \rightarrow T_{\mathrm{step}}^{\mathrm{end}}$
& N/A
& \multirow{2}{*}{Rating 1--5} \\

\textcolor{kolabpink}{\textbf{2b:}} Output quality
& $T_{\mathrm{step}}^{\mathrm{start}} \rightarrow T_{\mathrm{step}}^{\mathrm{end}}$
& N/A
& \\

\bottomrule
\end{tabular}
\end{table}

\textcolor{kolabpink}{\textbf{RQ1} (AI Adoption).}  We evaluate whether MLLMs can predict whether humans \adopt, \modify, or \reject agent contributions under two settings—recognition (1a) and forecasting (1b) —which differ in whether the human response is observable or not (Table~\ref{tab:benchmark-tasks}). Each input begins at the source node, where the human delegates to the agent ($T_{\mathrm{source}}^{\mathrm{start}}$). Forecasting clips average 14.0 minutes ($SD=13.7$; max=73.0), while prediction clips average 8.8 minutes ($SD=11.7$; max=72.0) with more details in Table~\ref{tab:clip_lengths}.

We are also curious whether longer context improves or degrades model performance. Thus, for Experiment 1c (step-N forecasting), the model must forecast the AI adoption label for each target AI node in the last step, while providing the preceding 1, 2, or 3 steps as additional context. For the longest context setting, motivated by how human annotators often need to watch the entire video context to determine the adoption label, we introduce Experiment 1d (workflow-level forecasting). In this setting, we provide the entire workflow up to the current target AI node.




\textcolor{kolabpink}{\textbf{RQ2} (Human Rating).}
An evaluator should distinguish genuinely successful interactions from those in which professionals were dissatisfied or produced lower-quality outcomes, rather than simply reflecting the predominantly positive ratings. Thus, we investigate whether MLLMs can similarly recover professionals' ratings from multimodal, long-horizon human-agent collaboration workflows. Given the full workflow video for each step, we evaluate whether frontier MLLMs can predict the professional's 1-5 rating of collaboration satisfaction (\textbf{Experiment 2a}) and output quality (\textbf{Experiment 2b}).

\textbf{Experimental Setup} We evaluate four leading multimodal agentic models, Gemini 3.8 Flash, Gemini 3.1 Pro, GPT-6 Astra\footnote{$^{+}$GPT was evaluated on fewer instances and steps (279 instances for 1a and 349 for 1b instead of 692 instances, 276 steps for 2a and 2b) due to cost limitation.}, and Claude Code Opus 5.5, and two open-weight video-capable MLLMs: Qwen3-VL-30B-A3B-Instruct (MoE) and Qwen3-VL-32B-Instruct (dense)~\citep{bai2025qwen3}. We compare against majority-class (\adopt) and mode-rating baselines for RQ1 and RQ2, respectively. Our primary input is workflow video, which captures actions inside and outside the AI interface. We additionally evaluate all seven combinations of video, graph, and chat to assess the contribution of each modality.

\subsection{Results and Discussion}
\begin{figure}[t]
\centering
\includegraphics[width=0.48\linewidth]{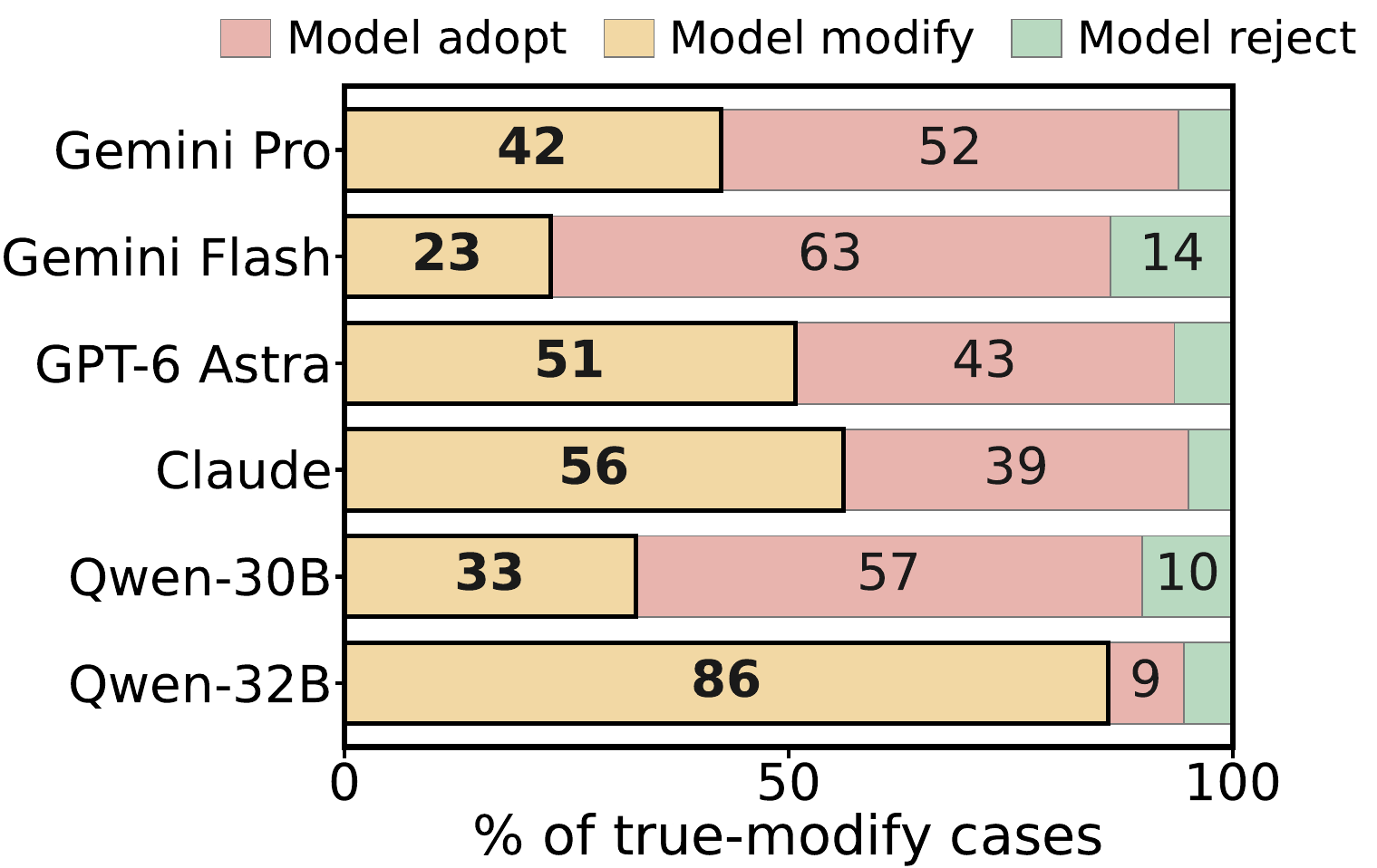} \hfill
\includegraphics[width=0.48\linewidth]{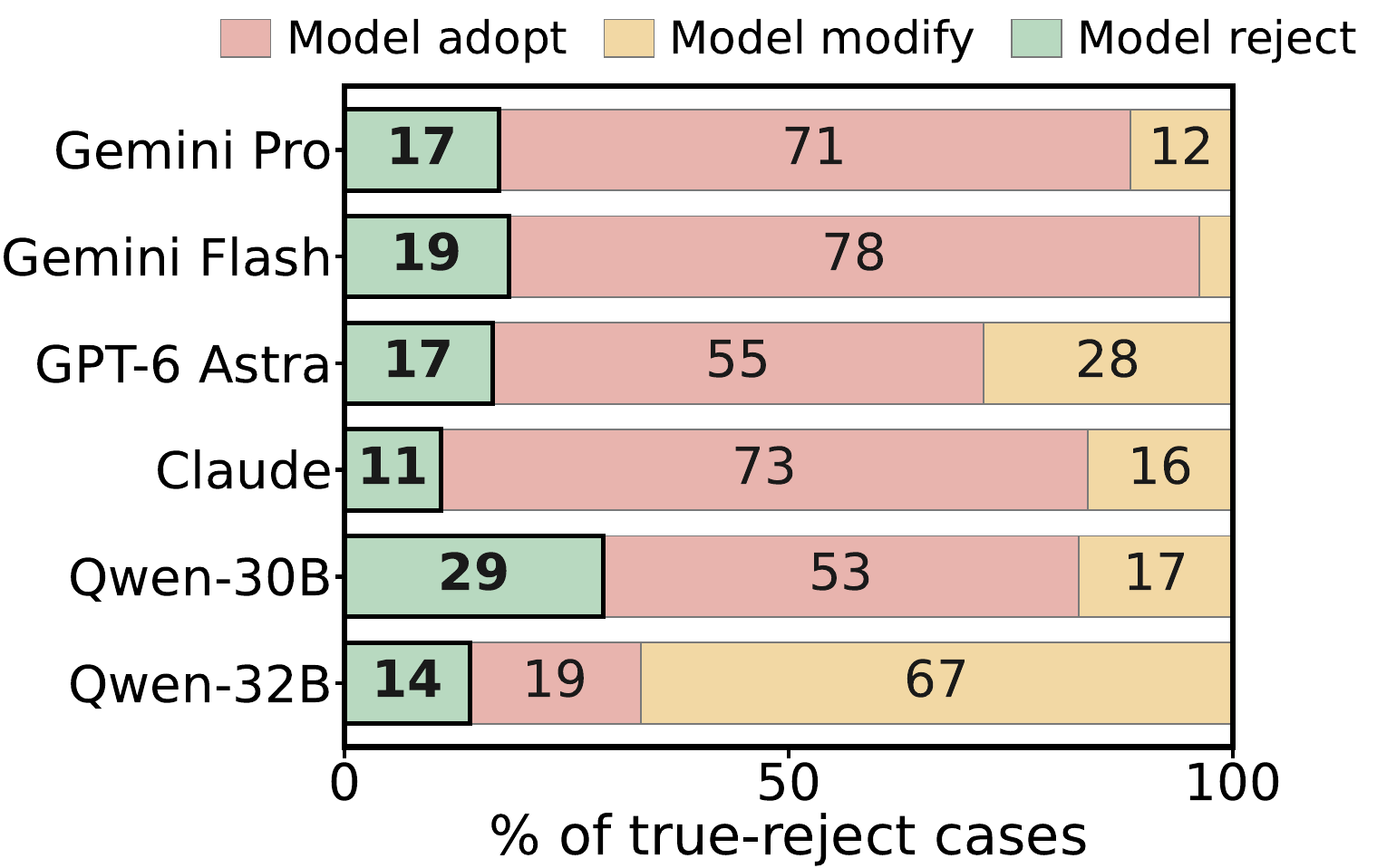}
\caption{Distributions of model-generated labels for \modify and \reject cases in AI adoption forecasting.}
\label{fig:failure_cases}
\vspace{-3mm}
\end{figure}

\textbf{Performance on forecasting AI adoption.}
Table~\ref{tab:response_prediction} shows each MLLM's performance with video only as input. Claude performs the best for both recognizing and forecasting human adoption behavior, yet even the upper-bound only reaches an F1-score of 0.676 aside from GPT-6, which was evaluated on fewer instances than other models. Full details of model comparison with the common instance sets are shown in the Table \ref{tab:exp1a_common_set} and Table \ref{tab:exp1b_common_set} in the Appendix. Since the default AI agent used in this study is Claude, we hypothesize that it has a strong advantage in understanding interactions within its own interface. Even so, Claude's F1 for forecasting human adoption is only 0.492, dominated by \adopt. It has substantially lower performance on \modify and \reject. Among the remaining models, Gemini 3.1 Pro performs the best when the human's actual response is observable, with an F1 score of 0.617. Gemini models performs particularly poorly at forecasting \reject, even worse than the open-weight Qwen models. Gemini 3.1 Pro is very biased toward generating \adopt. 52\% of true \modify and 71\% of true \reject are predicted as \adopt by Gemini 3.1 Pro (Figure~\ref{fig:failure_cases}).

Why should an AI model care to understand human collaboration with other AI models? Humans orchestrate various software tools in our dataset, and their workflows are not necessarily limited to a single AI model. Even within our study, professionals used other AI systems, such as Gemini, ChatGPT, and Grok, in addition to Claude. This makes it increasingly important for agents to understand human collaboration with other models, rather than only interactions involving themselves.

\begin{table}[t]
\centering
\caption{Benchmarking performance for AI adoption recognition (Experiment 1a) and for forecasting (Experiment 1b), for video only, mean{\scriptsize$\pm$}std over 3 runs per model. The best result in each column is \colorbox{bestbg}{\textbf{shaded}} and the second best result is \textbf{bolded}.}
\label{tab:response_prediction}
\resizebox{\linewidth}{!}{%
\setlength{\tabcolsep}{3.5pt}\renewcommand{\arraystretch}{1.15}\footnotesize
\begin{tabular}{l c @{\hspace{6pt}} ccc @{\hspace{6pt}} ccc}
\toprule
& \multicolumn{1}{c}{\textcolor{kolabpink}{\textbf{1a: Recognition}}} &
\multicolumn{3}{c}{\textcolor{kolabpink}{\textbf{1b: Forecasting}}} &
\multicolumn{3}{c}{\textcolor{kolabpink}{\textbf{1b: Per-label F1}}} \\
\cmidrule(lr){2-2} \cmidrule(lr){3-5} \cmidrule(l){6-8}
\textbf{Model} &
\textbf{F1} &
\textbf{Prec.} & \textbf{Rec.} & \textbf{F1} &
\adoptc & \modifyc & \rejectc \\
\midrule
\textcolor{baselinefg}{Majority-class} & \textcolor{baselinefg}{0.265} & \textcolor{baselinefg}{0.220} & \textcolor{baselinefg}{0.333} & \textcolor{baselinefg}{0.265} & \textcolor{baselinefg}{0.794} & \textcolor{baselinefg}{0.000} & \textcolor{baselinefg}{0.000} \\
\midrule
\mGemPro & 0.617\sd{0.007} & 0.461\sd{0.026} & 0.402\sd{0.011} & 0.391\sd{0.013} & 0.710\sd{0.012} & 0.333\sd{0.018} & 0.129\sd{0.031} \\
\mGemFlash & 0.587\sd{0.010} & 0.437\sd{0.016} & 0.383\sd{0.008} & 0.383\sd{0.009} & \textbf{0.758}\sd{0.009} & 0.218\sd{0.027} & 0.172\sd{0.013} \\
\mGPT$^{+}$ & \best{0.713} & \textbf{0.539}\sd{0.003} & \best{0.524}\sd{0.013} & \best{0.513}\sd{0.003} & 0.721\sd{0.023} & \best{0.582}\sd{0.098} & 0.235\sd{0.070} \\
\mClaude & \textbf{0.676} & \best{0.610} & \textbf{0.483} & \textbf{0.492} & \best{0.790} & \textbf{0.434} & 0.250 \\
\midrule
\mQwenMoE & 0.511\sd{0.019} & 0.389\sd{0.010} & 0.404\sd{0.011} & 0.364\sd{0.007} & 0.501\sd{0.004} & 0.305\sd{0.019} & \best{0.285}\sd{0.012} \\
\mQwenDense & 0.337\sd{0.007} & 0.436\sd{0.012} & 0.400\sd{0.012} & 0.235\sd{0.010} & 0.092\sd{0.006} & 0.337\sd{0.007} & \textbf{0.274\sd{0.022}} \\
\bottomrule
\end{tabular}%
}
\end{table}

\begin{figure}[t]
    \centering
    \includegraphics[width=\linewidth]{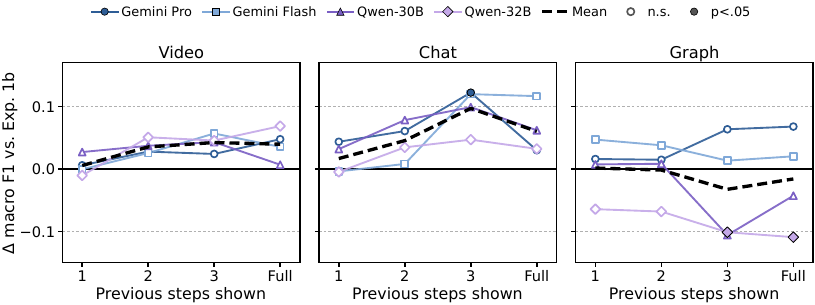}
    \caption{Change in macro-F1 as forecasting context grows. X-axis shows how many N previous steps are included as a context. Full refers to the start of the whole workflow.}
    \label{fig:context_help}
\end{figure}

\textbf{Does longer context help models anticipate human adoption behavior?}
We investigate whether providing more workflow context improves MLLMs' performance on forecasting human's behavior for AI adoption. Using 114 instances in which both the source and target AI nodes appear in the last step of each session, we extend the context to include up to three preceding steps (Experiment 1c) or the entire workflow up to the target AI node (Experiment 1d). Longer context seems to provide limited improvements, with macro-F1 changes below $0.07$ for most video and graph settings even though none of the improvements with video are statistically significant (Figure \ref{fig:context_help}). Chat logs benefit the most from additional context, with Gemini 3.1 Pro improving by $0.12$ F1 when given three preceding steps ($p<0.01$). However, this improvement does not persist when the entire workflow is provided. Longer context can also hurt performance, as Qwen-32B drops from $0.43$ to $0.32$ F1 with the full graph ($p<0.01$). These findings suggest that limited context alone may not explain models' difficulty in anticipating human adoption behavior as providing the entire workflow does not consistently improve adoption forecasting.

\begin{table}[t]
\centering
\caption{Rating prediction for Experiments 2a and 2b, mean{\scriptsize$\pm$}std over 3 runs. The best result in each column is \colorbox{bestbg}{\textbf{shaded}}; the second best result is \textbf{bolded}.}
\label{tab:collab_outcome}
\small
\setlength{\tabcolsep}{3.5pt}
\begin{tabular}{l ccc @{\hspace{8pt}} ccc}
\toprule
& \multicolumn{3}{c}{\textcolor{kolabpink}{\textbf{Exp. 2a: Collaboration}}} &
\multicolumn{3}{c}{\textcolor{kolabpink}{\textbf{Exp. 2b: Output Quality}}} \\
\cmidrule(lr){2-4} \cmidrule(l){5-7}
\textbf{Model} &
\textbf{MAE$\downarrow$} & \textbf{Within-1$\uparrow$} & \textbf{Spearman $\rho\uparrow$} &
\textbf{MAE$\downarrow$} & \textbf{Within-1$\uparrow$} & \textbf{Spearman $\rho\uparrow$} \\
\midrule
\textcolor{baselinefg}{Mode baseline} & \textcolor{baselinefg}{0.682} & \textcolor{baselinefg}{0.854} & \na & \textcolor{baselinefg}{0.735} & \textcolor{baselinefg}{0.831} & \na \\
\midrule
\mGemPro & \best{0.685}\sd{0.006} & \textbf{0.850}\sd{0.003} & 0.078\sd{0.057} & \best{0.764}\sd{0.016} & \textbf{0.821}\sd{0.006} & 0.046\sd{0.039} \\
\mGemFlash & \textbf{0.800}\sd{0.029} & 0.819\sd{0.014} & 0.011\sd{0.003} & \textbf{0.792}\sd{0.006} & 0.811\sd{0.002} & 0.042\sd{0.012} \\
\mGPT$^{+}$ & 0.945\sd{0.070} & 0.841\sd{0.032} & \best{0.379}\sd{0.130} & 0.912\sd{0.084} & 0.805\sd{0.030} & \best{0.352}\sd{0.101} \\
\mClaude$^{\dagger}$ & 0.858 & \best{0.865} & \textbf{0.136} & 0.817 & \best{0.849} & \textbf{0.208} \\
\midrule
\mQwenMoE & 0.898\sd{0.066} & 0.814\sd{0.033} & 0.102\sd{0.075} & 1.281\sd{0.011} & 0.641\sd{0.010} & 0.048\sd{0.034} \\
\mQwenDense & 1.194\sd{0.018} & 0.634\sd{0.004} & 0.130\sd{0.028} & 1.223\sd{0.013} & 0.633\sd{0.019} & 0.153\sd{0.005} \\
\bottomrule
\end{tabular}
\vspace{-3mm}
\end{table}

\textbf{Performance on predicting collaboration satisfaction and output quality.}
We evaluate alignment with human ratings using mean absolute error (MAE), within-1 accuracy, and Spearman's $\rho$. The mode baseline outperforms every MLLM in MAE of 0.682 for collaboration satisfaction and 0.735 for output quality, reflecting the strong skew of human ratings toward 5. Gemini 3.1 Pro achieves the lowest MAE of 0.685 for collaboration satisfaction and 0.764 for output quality, but even the best-performing model fails to outperform the baseline. Moreover, rank agreement with human judgments remains weak. GPT-6 achieves the highest Spearman correlation for both collaboration satisfaction ($\rho={0.379}$) and output quality ($\rho={0.352}$), while several models exhibit near-zero correlations. Thus, when disregarding differences in absolute rating scale, current frontier multimodal agents and LLMs still struggle to recover how professionals comparatively judge their collaboration and resulting outputs.

\begin{figure}[!t]
    \centering
    \begin{subfigure}[b]{0.5\linewidth}
        \centering
        \adjincludegraphics[width=\linewidth,]{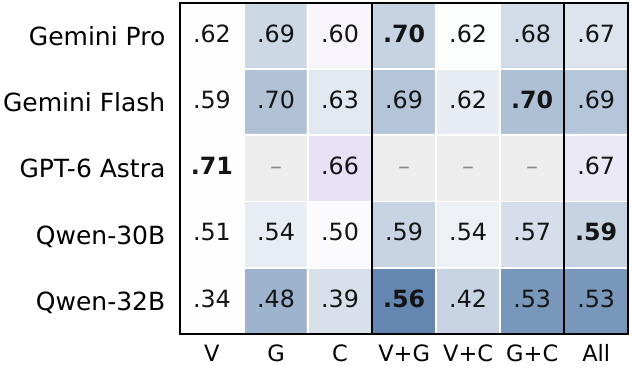}
        \caption{1a: Recognition (F1)}
        \label{fig:modality_variants_a}
    \end{subfigure}\hfill
    \begin{subfigure}[b]{0.5\linewidth}
        \centering
        \adjincludegraphics[width=\linewidth,]{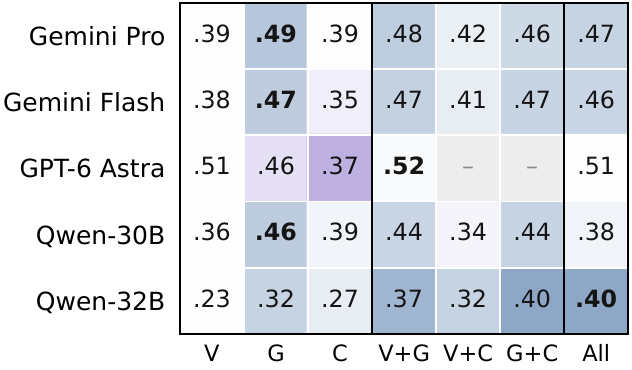}
        \caption{1b: Forecasting (F1)}
        \label{fig:modality_variants_b}
    \end{subfigure}

    \vspace{4pt}
    \begin{subfigure}[b]{0.5\linewidth}
        \centering
        \adjincludegraphics[width=\linewidth,]{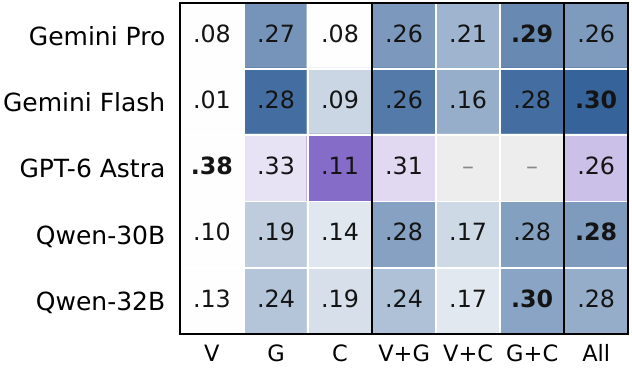}
        \caption{2a: Collaboration satisfaction ($\rho$)}
        \label{fig:modality_variants_c}
    \end{subfigure}\hfill
    \begin{subfigure}[b]{0.5\linewidth}
        \centering
        \adjincludegraphics[width=\linewidth,]{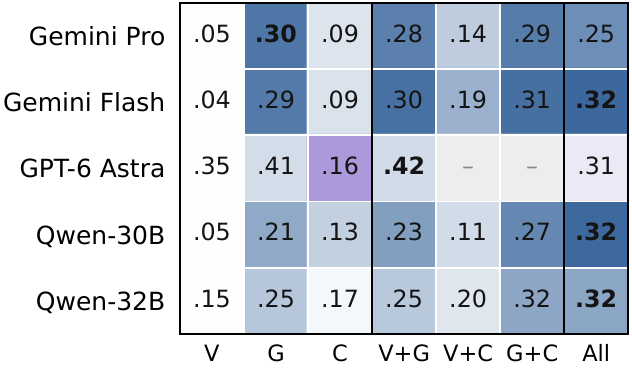}
        \caption{2b: Output quality ($\rho$)}
        \label{fig:modality_variants_d}
    \end{subfigure}
    \caption{Full results for seven input-modality combinations across the four benchmark tasks. V stands for video, G: graph, and C: chat log; ``+'' is a combination of modalities, ``All'' refers to using video, graph, and chat as the input. The best score in each row is in bold.}
    \label{fig:modality_variants}
    \vspace{-10pt}
\end{figure}

\begin{figure}[!t]
    \centering
    \includegraphics[width=\linewidth]{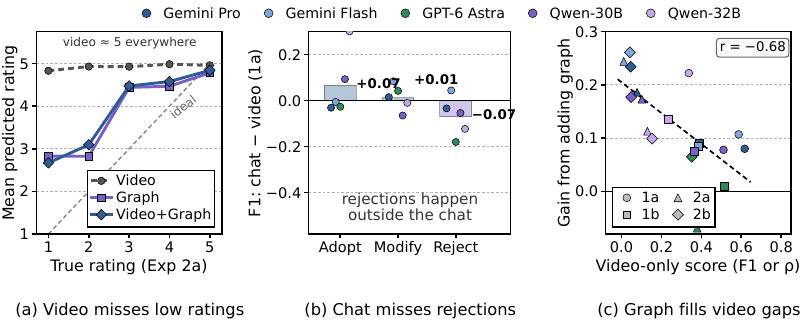}
    \vspace{-15pt}
    \caption{How each modality helps.}
    \label{fig:modality_findings}
    \vspace{-20pt}
\end{figure}

\subsection{Modality Analysis} 
We examine how video, graph, and chat, individually and in combination, affect models' understanding of human-agent collaboration (Figures~\ref{fig:modality_variants} and~\ref{fig:modality_findings}). No modality consistently performs best across models or experiment tasks. Video alone yields the highest recognition performance with GPT-6 (F1$=0.71$, Experiment 1a). Graphs, either alone or combined with video or chat, tend to yield the best forecasting performance, with GPT-6 achieving an F1 score of 0.52 in Experiment 1b. However, chat alone particularly hurts recognition of \reject cases: relative to video, F1 changes by $+0.07$ for \adopt, $+0.01$ for \modify, and $-0.07$ for \reject (Figure~\ref{fig:modality_findings}b), likely because chat does not capture how agent outputs are subsequently used outside the AI interface. Yet video is not consistently optimal. Human-curated workflow graphs, which capture high-level professional activities, particularly benefit the non-agentic Qwen models; for Qwen-32B, adding graphs to video improves forecasting F1 from $0.23$ to $0.37$.

For human-rating prediction, performance remains limited regardless of modality (Figure~\ref{fig:modality_variants}c--d). Correlations peak at only $\rho=0.38$ for collaboration satisfaction (GPT-6, video only) and $\rho=0.42$ for output quality (GPT-6, video+graph). Graphs are particularly informative for output quality: video alone yields near-zero correlations for Gemini 3.8 Flash and Qwen-30B ($\rho=0.04$--$0.05$), while adding graphs raises them to $\rho=0.23$--$0.30$. Overall, providing additional modalities does not resolve MLLMs' difficulty in approximating professionals' judgments of collaboration experience and output quality.

\begin{figure}[t]
\centering
\includegraphics[width=0.48\linewidth]{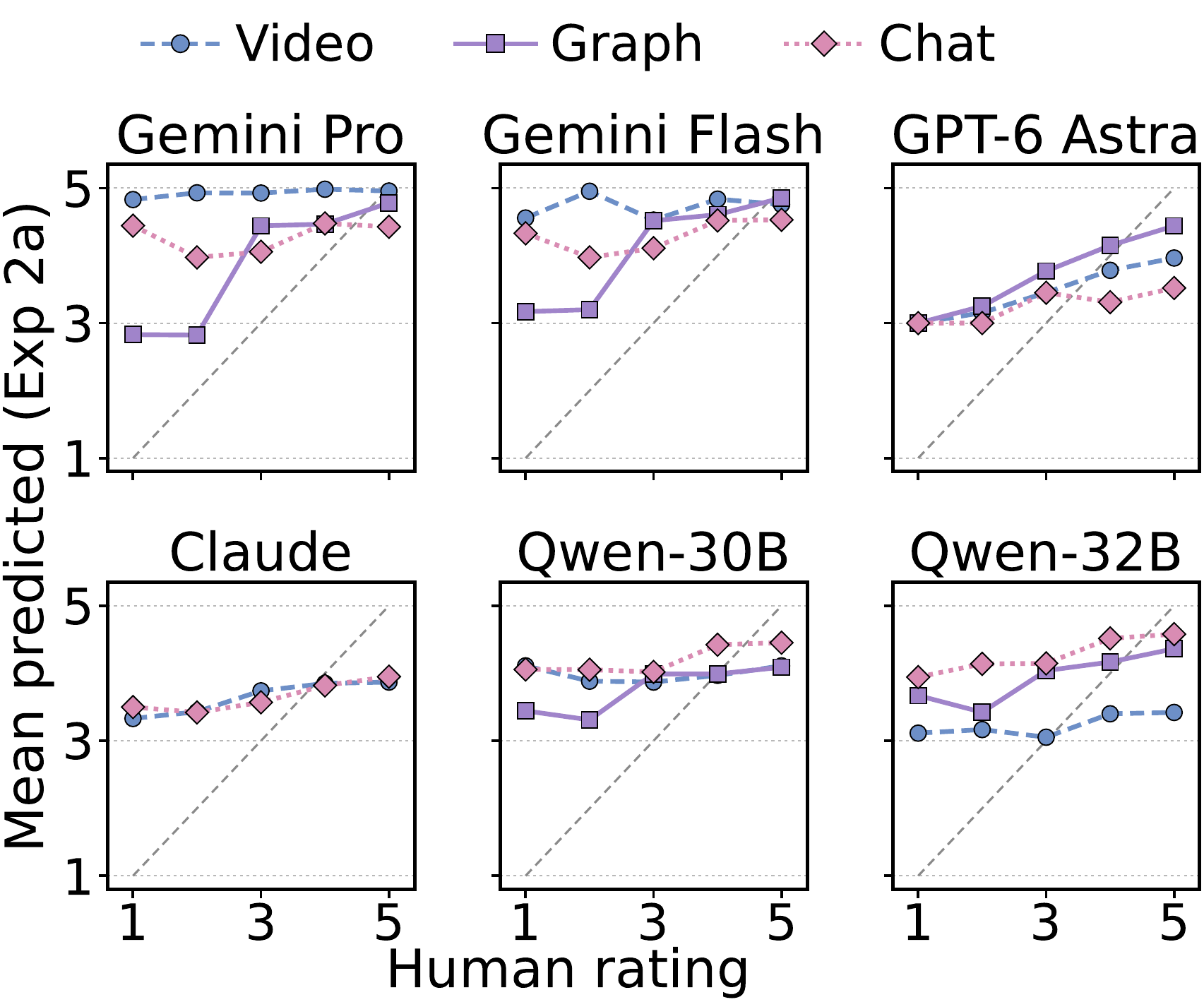}\hfill
\includegraphics[width=0.48\linewidth]{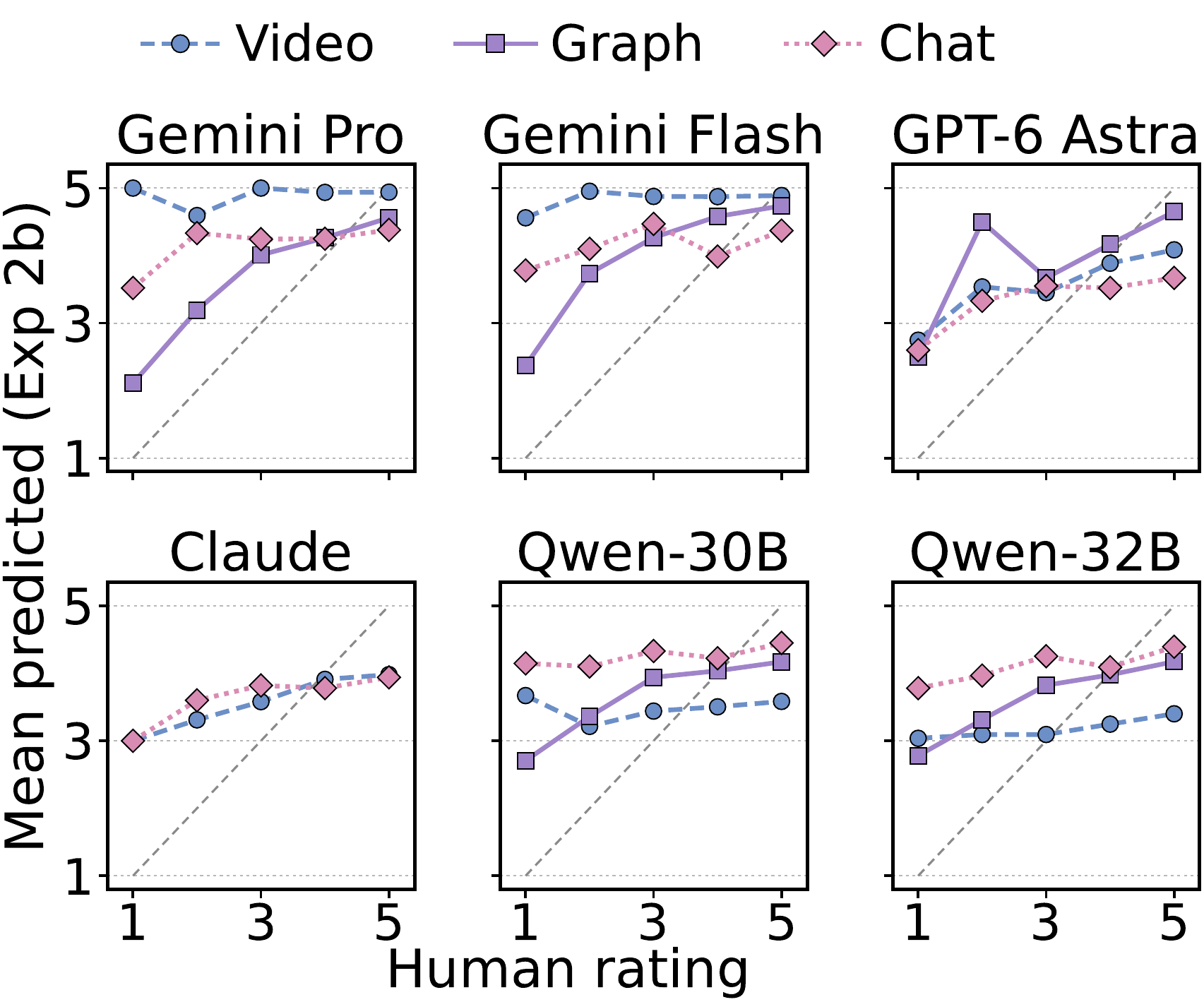}
\caption{Mean predicted rating vs. human rating for collaboration (Exp 2a) and output quality (Exp 2b).}
\label{fig:bias_for_exp2a_2b}
\vspace{-5.5mm}
\end{figure}

\subsection{Failure Cases}
For forecasting whether humans adopt agent contributions, models often strongly prefer certain labels although the true labels are \modify or \reject (Figure~\ref{fig:failure_cases}). Gemini 3.1 Pro and Flash overpredict \adopt. With video inputs, Gemini 3.8 Flash forecasts \adopt for 83.8\% of examples compared with a 65.9\% ground-truth rate, and Qwen3-32B instead favors \modify, predicting it in 87.3\% of cases compared with 19.1\% in the ground truth. These biases largely persist with graph and chat inputs as well (e.g., Flash generates \adopt for 82.3\% and 74.0\%, and Qwen3-32B outputs \modify for 67.7\% and 62.0\%). Models therefore exhibit distinct forecasting biases, which vary with both the model and input modality.

For human-rating prediction (Figure~\ref{fig:bias_for_exp2a_2b}), some models predict similar ratings all the time. With video, Gemini 3.1 Pro predicts 4.83 for a true rating of 1 and 4.95 for a 5 (Experiment 2a); in Experiment 2b, it predicts 5.00 and 4.94. Graph input improves this separation, with gaps between true ratings of 1 and 5 reaching 1.95 and 2.45 for Experiments 2a and 2b. Claude and Qwen variants also show a similar pattern. Claude's predictions range only from 3.3 to 3.9 across the entire 1–5 rating scale with video input in both experiments, while the ratings of Qwen variants range from 3.04 to 4.15. However, GPT-6 already separates low and high ratings with video and remains the most stable across modalities.

These systematic biases raise a more fundamental question: to what extent are MLLMs actually grounding their predictions in the observed workflow? If models understand the workflow context, their predictions should vary with the human behaviors and outcomes they observe. Some models repeatedly produce similar labels or ratings despite different inputs and ground-truths. It suggests that their predictions may be driven by model-specific priors instead of the workflow itself.

\section{Conclusion and Future Work}
We have introduced \dataset as a new dataset to benchmark frontier multimodal agentic models on their capabilities to support human-agent collaboration better on long-horizon workflows. Our results show that even frontier MLLMs, including GPT-6, struggle to predict and understand human–AI collaboration. As a future work, we will explore ways to improve these capabilities by constructing training datasets or developing sandbox environments where models can practice collaborating with simulated users before real-world deployment.


\subsection*{Ethics statement}
Understanding how humans collaborate with AI agents is important for developing AI systems that better support human needs and decisions. However, evaluating human-agent collaboration also raises ethical concerns. First, models that inaccurately predict whether humans will adopt, modify, or reject AI contributions may misrepresent human intentions and preferences. Our findings highlight the limitations of current MLLMs in understanding professional human-agent collaborative workflows. Relying on these models as proxies for human judgments could lead to misleading assessments, particularly in professional domains where decisions require specialized expertise and contextual understanding. Second, collecting and releasing real-world professional workflows raises privacy concerns, as screen recordings and interaction logs may contain sensitive information. To mitigate these risks, we remove all personally identifiable information (PII) in our data. Our work aims to support the development of human-centered collaborative agents while emphasizing that model predictions should not replace human judgments or undermine human agency. This study was approved by our institution's IRB. All participant data are anonymized and securely stored. 

\subsection*{Reproducibility statement}
Code and data are released in our project webpage \url{https://minnesotanlp.github.io/kolab/}. Details about data collection details are explained in \ref{appendix2:collection_details} in the Appendix. Data processing and annotation details are explained in \ref{appendix:data_processing}. All model prompts and model settings are elaborated in the Appendix. 


\subsubsection*{Acknowledgments}
We are grateful to Jingcheng Liang, Risako Owan, and Su Yeon Han for helping with data cleaning and annotation. This research was supported in part by the Interdisciplinary Doctoral Fellowship from the University of Minnesota for the first author. 

\bibliography{iclr2027_conference}
\bibliographystyle{iclr2027_conference}

\appendix
\section{Appendix}
\subsection{Professionals Background and Software Usage}
\label{appendix:professional_background}
\begin{figure}
    \centering
    \includegraphics[width=0.9\linewidth]{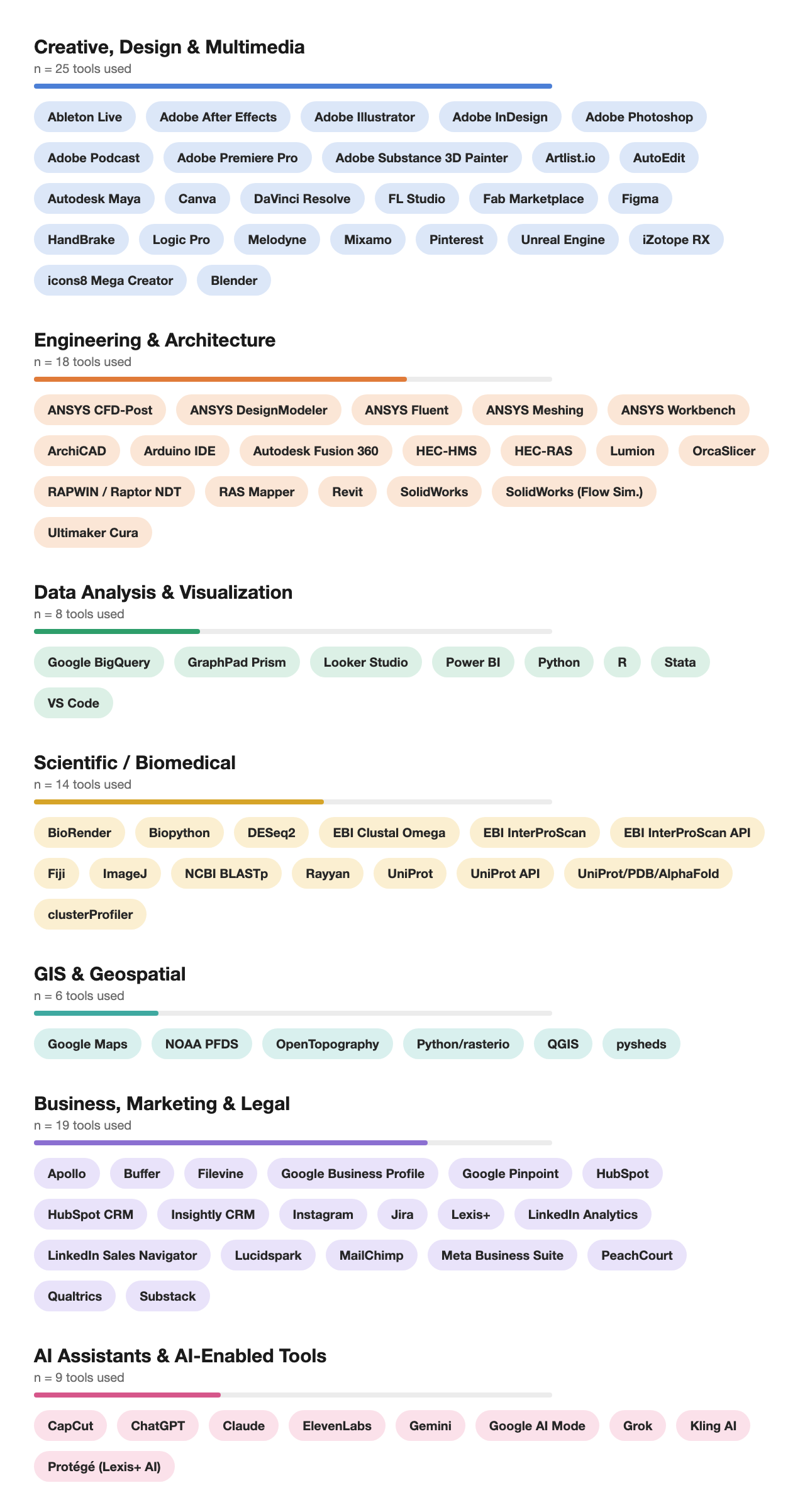}
\caption{Software used by participants in their day-to-day work, grouped by category.}
\label{fig:software_use}
\end{figure}

The 50 study participants span 10 O*NET sectors and 40 distinct occupations, ranging from film editors, graphic designers, and audio engineers to aerospace engineers, GIS specialists, attorneys, and bioinformaticians. We prioritize non-computer science professions, which remain comparatively under-resourced with real-world task data relative to software engineering \citep{patwardhan2026gdpval, shao2025future, sun2026agents, miserendino25icml}.
Across these roles, participants reported using 90 distinct profession-specific non-AI softwares (excluding Google or Microsoft office products) and 9 AI tools in their day-to-day work, as shown in Figure~\ref{fig:software_use}. This mix of shared productivity tools and domain-specific software illustrates the breadth of professional practice that the benchmark's workflows are grounded in.

\begin{table}[htbp]
\centering
\caption{Summary statistics of the 50 professionals.}
\label{tab:participant-summary}
\small
\begin{minipage}[t]{0.48\linewidth}
\centering
\begin{tabular}[t]{@{}llr@{}}
\toprule
\rowcolor{headerbg}
\textbf{Category} & \textbf{Item} & \textbf{Count} \\
\midrule
\multirow{3}{*}{Coverage}
& Total participants & 50 \\
& O*NET sectors & 10 \\
& Occupations & 40 \\
\midrule
\multirow{4}{*}{Education}
  & PhD & 10 \\
& Master's & 20 \\
& Bachelor's & 15 \\
& Other & 5 \\
\midrule
\multirow{5}{*}{Age range}
& 18--25 & 8 \\
& 26--35 & 20 \\
& 36--45 & 13 \\
& 46--55 & 8 \\
& 56--65 & 1 \\
\midrule
\multirow{5}{*}{\shortstack[l]{Work\\experience}}
& \textless\ 1 year & 7 \\
& 1--3 years & 19 \\
& 4--6 years & 7 \\
& 6--10 years & 5 \\
& \textgreater\ 10 years & 12 \\
\midrule
\multirow{4}{*}{\shortstack[l]{Self-rated\\expertise}}
& Basic (2/5) & 2 \\
& Intermediate (3/5) & 11 \\
& Advanced (4/5) & 18 \\
& Expert (5/5) & 19 \\
\midrule
\multirow{3}{*}{\shortstack[l]{GenAI\\familiarity}}
& Low (1--4/10) & 1 \\
& Medium (5--7/10) & 12 \\
& High (8--10/10) & 37 \\
\bottomrule
\end{tabular}
\end{minipage}\hfill
\begin{minipage}[t]{0.48\linewidth}
\centering
\begin{tabular}[t]{@{}llr@{}}
\toprule
\rowcolor{headerbg}
\textbf{Category} & \textbf{Country} & \textbf{Count} \\
\midrule
\multirow{21}{*}{\shortstack[l]{Geography\\(21 countries)}}
& USA & 20 \\
  & Colombia                     & 3  \\
  & India                        & 3  \\
  & Mexico                       & 3  \\
  & Germany                      & 3  \\
  & Nigeria                      & 2  \\
  & UK                           & 2  \\
  & Argentina                    & 1  \\
  & Ecuador                      & 1  \\
  & Egypt                        & 1  \\
  & El Salvador                  & 1  \\
  & Finland                      & 1  \\
  & Kenya                        & 1  \\
  & Montenegro                   & 1  \\
  & Morocco                      & 1  \\
  & Netherlands                  & 1  \\
  & Pakistan                     & 1  \\
  & South Africa                 & 1  \\
  & Sri Lanka                    & 1  \\
  & Turkey                       & 1  \\
  & Yemen & 1 \\
\bottomrule
\end{tabular}
\end{minipage}
\end{table}

\begin{figure}
    \centering
    \includegraphics[width=\linewidth]{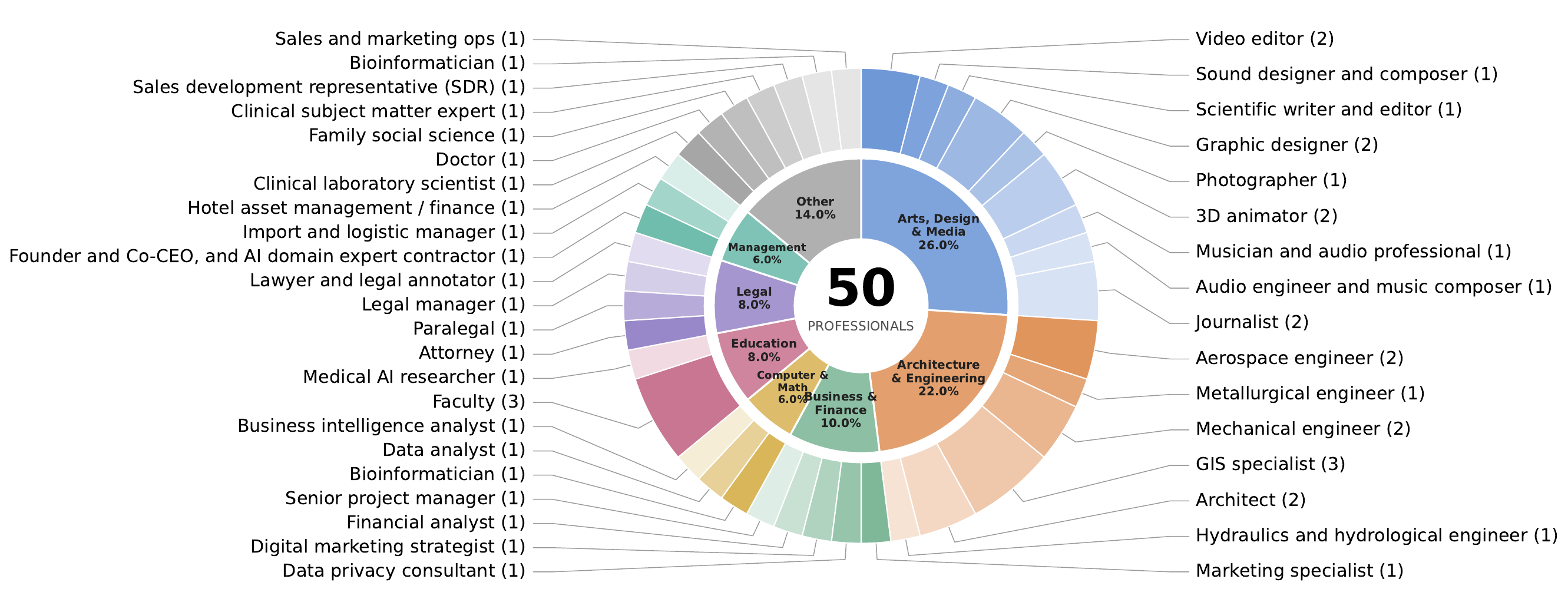}
\caption{Occupation distributions of \dataset. ``Other'' refers to three O*NET sectors: Life \& Social Science, Sales \& Related, and Healthcare.}
\label{fig:occupations_full}
\end{figure}

\subsection{Data Collection Details}
\label{appendix2:collection_details}
We collected \dataset through two sessions per participant, in which we recorded the participant's computer screen. Each session lasted approximately one hour or longer and centered on a problem chosen by the participant. After data processing and due to time constraints, some problems contained fewer than five steps; however, each interaction session still lasted approximately one hour. To capture different patterns of human-agent collaboration, we used two protocols. In Session 1, participants were encouraged to interact with the agent at least five times while completing their workflow. In Session 2, participants were encouraged to interact with the agent at each step of the workflow. The two sessions did not have to address the same problem. However, participants could use the second session to continue or extend the work from the first session. For example, a participant might use the first session to develop a branding logo and use the second session to create a marketing advertisement based on the resulting brand identity. Participants were required to interact with agentic AI, Claude, using our credentials during each session and were free to use other AI agents when appropriate for their work. For example, an architect might use Claude for planning and reasoning while using ChatGPT to generate or render architectural images. At the time of data collection, most participants used the default Claude model (Sonnet 5.5), while a few used Opus 5. Each participant planned their workflow before the recording sessions to ensure that their problem satisfied our task criteria. After the recording sessions, they updated the workflow document and completed surveys about their collaboration experience and output quality. During each recording session, participants were encouraged to think aloud; we used the transcriptions to construct the workflow graphs and to guide the annotation of response labels.





\begin{figure}
    \centering
\includegraphics[width=0.62\linewidth]{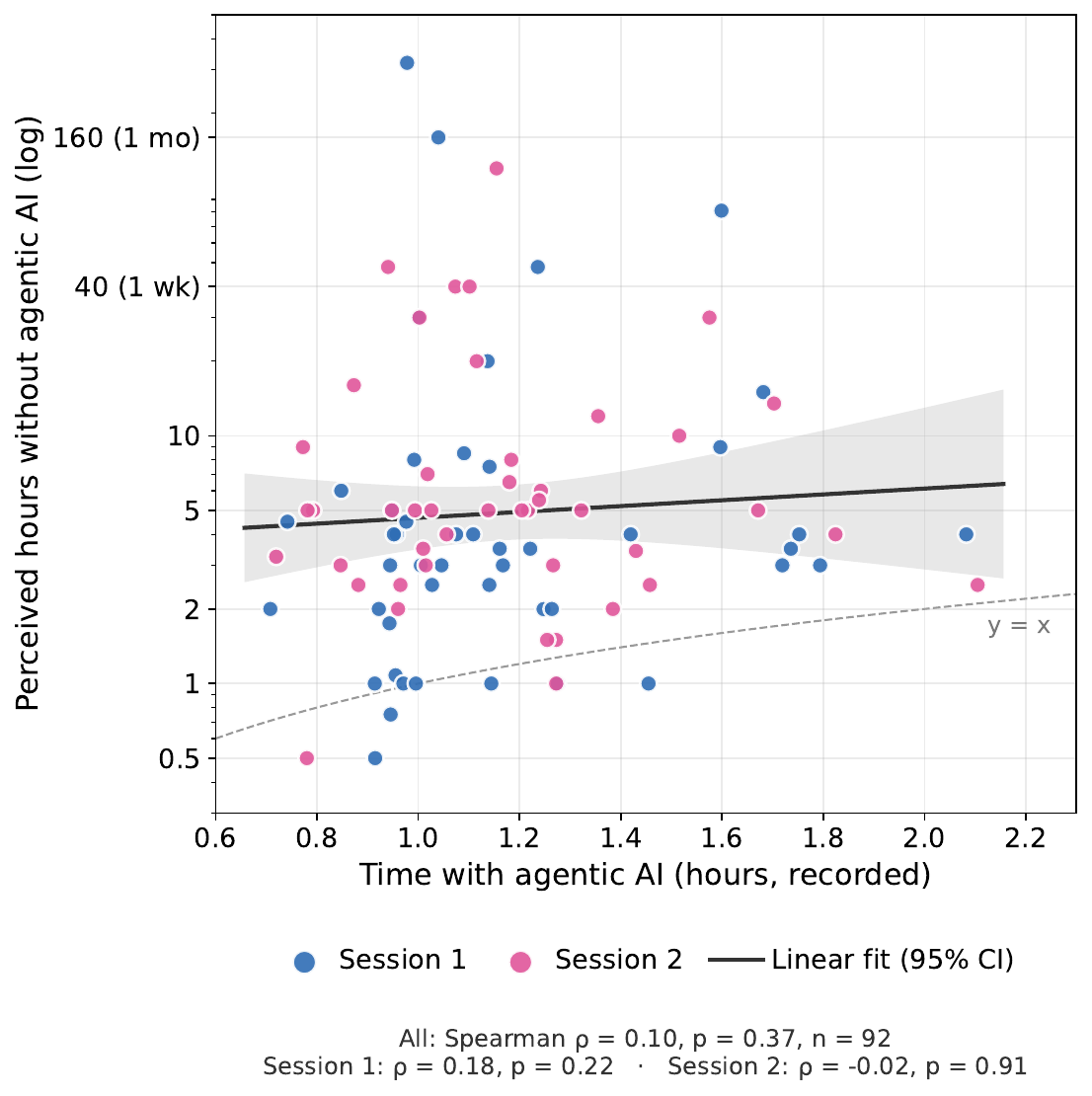}
\caption{Time spent in each recording session working with agentic AI versus the participant's estimated time to solve the same problem without agentic AI.}
\label{fig:time_comparison}
\end{figure}

\begin{figure}
    \centering
    \includegraphics[width=0.8\linewidth]{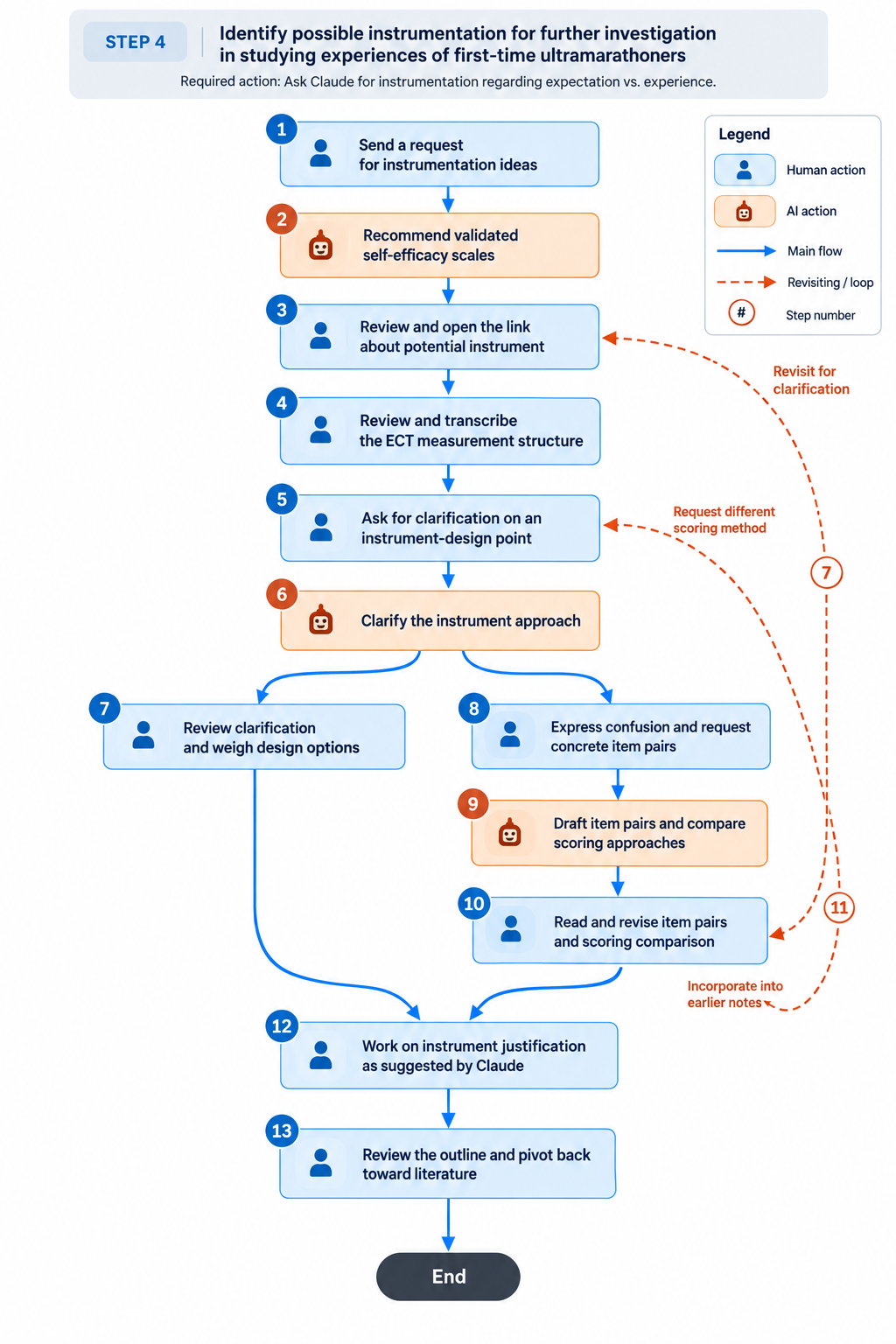}
\caption{Example workflow graph for one step of a participant's session.}
\label{fig:p4_graph}
\end{figure}

\subsection{Graph as a Collaboration Workflow Representation}
Actions and timestamps from workflow video are aligned with steps documented by each professional. We then represented each workflow as a directed graph, where each node corresponds to a human or agent action. The research team and professionals jointly reviewed and refined these graphs. Graphs can capture non-linear workflows; for instance, a node branches when activities occur in parallel and loops happen when the same action is repeated.

\label{graph_representation}
Formally, each session is represented as a directed temporal graph
\begin{equation}
G_{temporal}= (V, E, a, w, \tau),
\end{equation}

where $V$ is the set of nodes, each corresponding to a discrete human or AI action, and $E \subseteq V \times V$ is the set of directed edges, with $(u, v) \in E$ indicating that action $v$ directly follows from action $u$ within the workflow. Each node $v \in V$ is annotated by three functions:
an actor label 
\begin{equation}
a: V \to \{\hact, \aact\},
\end{equation}
indicating whether the action was performed by the human or the agent; a
workflow-step label
\begin{equation}
w: V \to \{1, \dots, K\},
\end{equation}

indicating which of the $K$ steps of the task the action belongs to; and a
time interval
\begin{equation}
\tau(v) = [t_s(v), t_e(v)] \subset \mathbb{R}_{\geq 0}.
\end{equation}
recording when the action began and ended.

After graph construction, two human annotators annotated how each AI output was incorporated into the workflow. For every observed user response to an AI output, annotators identified the \emph{prompt node} that triggered generation, the \emph{AI-content node} being acted on, and a \emph{response label}: \adopt (incorporated without modification), \modify (incorporated with changes), or \reject (explicitly rejected or not incorporated). This is an annotation layer on top of $G_{temporal}$: each node $n \in V$ representing a user response carries a set of response triples as follows.
\begin{equation}
U(n) = \{(p_i, c_i, \ell_i)\}_{i=1}^{m_n},
\end{equation}
where $p_i \in V \cup \{\emptyset\}$ is the prompt node (absent if the user acts on existing AI content without a new prompt), $c_i \in V$ is the AI-content node ($a(c_i) = \aact$), and $\ell_i \in \{\text{adopt}, \text{modify}, \text{reject}\}$ is the response label.

\newpage
The following is an example workflow document that an architect filled out to plan their workflow before each recording session. After the session, they revised the content and filled out the assessment questions about collaboration experience and output quality ratings. Other step details are omitted due to space. 

\begin{promptbox}
\textbf{Goal:} Developing an architectural plan for Residential House

\textbf{Context:} I have chosen this project because it requires a set of expertise to layout an architectural plan. Conditions including design brief, site and weather analysis, building codes, area and zoning standards to evaluate and design according to that as an expert.
\newline \newline
\textbf{Did you achieve your goal after \textasciitilde 1-hour interaction with Claude?} Yes
\newline
\textbf{Why or why not?} Claude was able to produce layouts
\newline \newline
\textbf{Step 1 name:} Summary of design brief, a set of requirements to start the project
\newline \newline
\textbf{Step action:} Prepare the design brief and define scope of project.
\newline
\textbf{Input:} client brief shared with me
\newline
\textbf{Output:} Revise details when design brief in comprehensive format is generated.
\newline
\textbf{Softwares:} Claude
\newline \newline
\textbf{Rubric:}
\begin{enumerate}
\item Each design improvement includes a clear architectural reason
\item Each revision idea includes a rationale.
\item The revised layout remains consistent with the client brief. 
\item Suggestions are consistent with the requirements.
\end{enumerate}

\textbf{Output Quality and Collaboration Assessment}
\begin{enumerate}
\item Rate the output quality from 1--5 (1 not satisfy, 5 satisfy). \textbf{4}
\item Reason for your rating? \textit{``It created a set of pointers in design brief from the narrative, and curated a long brief in detail. Added extra things also that were not questioned.''}
\item Which ones of the rubrics above are satisfied in this output? \textbf{2, 3, and 4}
\end{enumerate}
\textbf{Human-Agent Collaboration}
\begin{enumerate}
\item Did the AI Agent successfully complete the task? \textbf{Yes}
\item Why? \textit{It created a set of pointers in design brief from the narrative, and curated a long brief in detail. Added extra things also that were not questioned.}
\item Collaboration satisfaction --- write a number (1--5) to the left of each statement, where 1 = not satisfied and 5 = satisfied:
\begin{itemize}
\item \textbf{5} --- The agent understood the task, context, and instructions provided.
\item \textbf{4} --- The agent responded appropriately to requests and adapted when given feedback or corrections.
\item \textbf{4} --- The agent reduced effort, time, or cognitive load during the task.
\item \textbf{4} --- The agent proactively suggested useful ideas, alternatives, or next steps when appropriate.
\item \textbf{5} --- The agent produced accurate, relevant, and trustworthy outputs that required minimal correction.
\item \textbf{2} --- The interaction felt smooth, with the agent contributing at the right moments without creating unnecessary work.
\item \textbf{3} --- I was able to guide, redirect, and override the agent's actions when desired.
\end{itemize}
\item Rate the overall collaboration experience from 1--5 (1 not satisfy, 5 satisfy). \textbf{4}
\end{enumerate}

\vspace{4pt}
\textbf{Step 2:} area codes to use in project

\textbf{Step 3:} plan zoning and divisions of spaces

\textbf{Step 4:} Arrangement of spaces and Finalizing layout

\textbf{Step 5:} Architectural floor Plan finalization
\end{promptbox}

\newpage
Another example of workflow document by a different participant (sound designer and composer) is shown below. In this case, the agentic AI failed to help and actually slowed down the process. 

\begin{promptbox}
\textbf{Goal:} Sound Design a promotional video for a real estate sales pitch

\textbf{Context:} This goal will allow me to evaluate how AI can enhance my work as a sound designer. The plan is to work on a promotional sales pitch for apartments in cozumel, which was recorded by a production company, but they lack sound for many of the camera shots they use (I.E: Shots of the beach, of the town and its localities, etc.) This lack of sound makes the project seem empty, therefore, a sound designer is needed to enhance the immersion of what is shown by adding the ambience noise, sound of actions (Walking, placing, etc) while ensuring audio files don't overlap and work in symphony.
\newline \newline
\textbf{Did you achieve your goal after \textasciitilde 1-hour interaction with Claude?} No
\newline \newline
\textbf{Why or why not?} \textit{``Although we did advance on the video, It was not finished in the 1 hour mark. I felt like I had to keep up with Claude at the start, rather than being benefited by it while working. This gradually improved when I used it for smaller activities like finding specific SFX or a good usage of effect plugins, but when I initially tried to use it as an analysis assistant, it took almost 20 minutes just analyzing the video, followed by 10 more minutes of analyzing just a fragment of it and providing ideas, which had slowed me down significantly, as I was already done with most of the other things I had not delegated to Claude.''}
\newline \newline
\textbf{Step 1 name:} Analyze the video and understand key moments from it
\newline \newline
\textbf{Step action:} Understand key moments where sound design can exponentially increase quality and immersion for the video, uploading the video to Claude for it to create this in depth analysis.
\textbf{Input:} MP4 video with dialogue and visual insights to the required SFX
\textbf{Output:} A pdf or DOCX document with key moments, time stamps and descriptions of the required sound effects.
\textbf{Softwares:} FL Studio, Artlist, Claude Max
\newline \newline
\textbf{Rubric:}
\begin{enumerate}
\item A score of 5 means that Claude was able to analyze the video and describe with detail what it is, what its selling, what key visuals and moments it has and how to interpret it as a client.
\item A score of 5 means the analysis included accurate, verifiable timestamps for each key moment, ready to use directly without needing to re-watch the video.
\end{enumerate}

\textbf{Output Quality and Collaboration Assessment}
\begin{enumerate}
\item Rate the output quality from 1--5 (1 not satisfy, 5 satisfy). \textbf{1}
\item Reason for your rating? \textit{``Claude was not able to fully analyze the video, due to a runout of time.''}
\item Which ones of the rubrics above are satisfied in this output? \textbf{N/A}
\end{enumerate}
\textbf{Human-Agent Collaboration}
\begin{enumerate}
\item Did the AI Agent successfully complete the task? \textbf{No.}
\item Why? \textit{``Claude mentioned a timeout when trying to analyze the video.''}
\item Collaboration satisfaction --- write a number (1--5) to the left of each statement, where 1 = not satisfied and 5 = satisfied:
\begin{itemize}
\item \textbf{3} --- The agent understood the task, context, and instructions provided.
\item \textbf{1} --- The agent responded appropriately to requests and adapted when given feedback or corrections.
\item \textbf{1} --- The agent reduced effort, time, or cognitive load during the task.
\item \textbf{1} --- The agent proactively suggested useful ideas, alternatives, or next steps when appropriate.
\item \textbf{1} --- The agent produced accurate, relevant, and trustworthy outputs that required minimal correction.
\item \textbf{1} --- The interaction felt smooth, with the agent contributing at the right moments without creating unnecessary work.
\item \textbf{2} --- I was able to guide, redirect, and override the agent's actions when desired.
\end{itemize}
\item Rate the overall collaboration experience from 1--5 (1 not satisfy, 5 satisfy). \textbf{1}
\end{enumerate}

\vspace{4pt}
\textbf{Step 2:} Develop the sound design concept

\textbf{Step 3:} Find and select appropriate sound effects

\textbf{Step 4:} Generate a sound effect from scratch

\textbf{Step 5:} Create a step by step guide to improve the realism of a sound effect using effects (EQ, Reverb, etc.)
\end{promptbox}

\begin{table}[H]
\centering
\caption{Distribution of collaboration satisfaction and output quality ratings, by session.}
\label{tab:rating-distribution}
\begin{tabular}{lrrr}
\toprule
Rating & Session 1 & Session 2 & Total \\
\midrule
\multicolumn{4}{l}{\textbf{Collaboration Satisfaction}} \\ \midrule
5 & 117 (53.7\%) & 122 (56.2\%) & 239 (54.9\%) \\
4 & 61 (28.0\%)  & 68 (31.3\%)  & 129 (29.7\%) \\
3 & 28 (12.8\%)  & 16 (7.4\%)   & 44 (10.1\%)  \\
2 & 8 (3.7\%)    & 7 (3.2\%)    & 15 (3.4\%)   \\
1 & 4 (1.8\%)    & 4 (1.8\%)    & 8 (1.8\%)    \\
\midrule
\multicolumn{4}{l}{\textbf{Output Quality}} \\ \midrule
5 & 119 (49.8\%) & 133 (57.3\%) & 252 (53.5\%) \\
4 & 81 (33.9\%)  & 65 (28.0\%)  & 146 (31.0\%) \\
3 & 27 (11.3\%)  & 20 (8.6\%)   & 47 (10.0\%)  \\
2 & 6 (2.5\%)    & 9 (3.9\%)    & 15 (3.2\%)   \\
1 & 6 (2.5\%)    & 5 (2.2\%)    & 11 (2.3\%)   \\
\bottomrule
\end{tabular}
\end{table}

Table \ref{tab:rating-distribution} shows the rating distribution by participants for collaboration satisfaction (1 low, 5 high) and output quality (1 low, 5 high) for each step. 

\begin{figure}
    \centering
    \includegraphics[width=0.9\linewidth, trim={0cm 0cm 0cm 3cm}, clip]{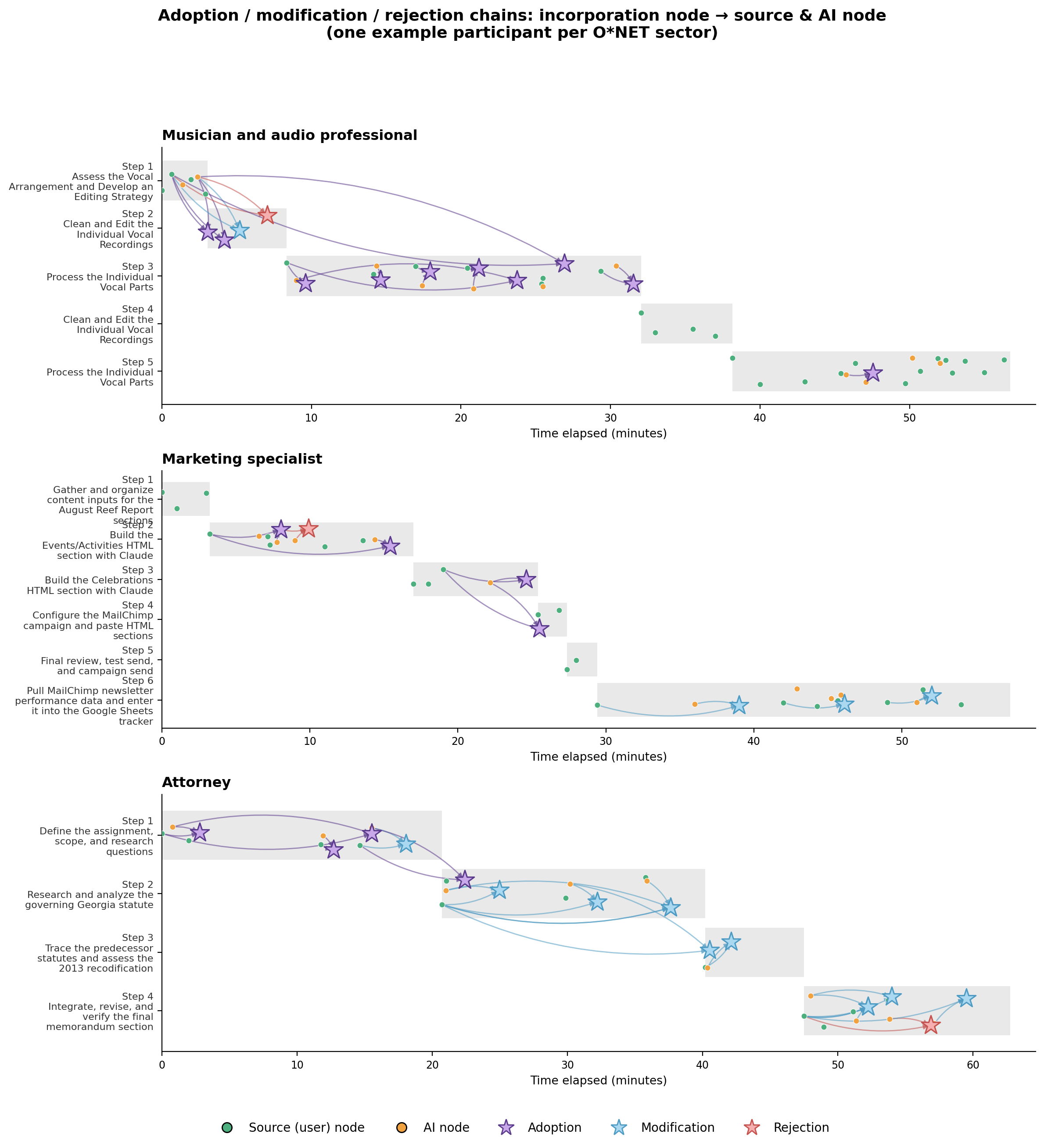}
\caption{How three professionals (musician, marketing specialist, and attorney) adopt, modify, and reject agent outputs.This is based on the real data point in \dataset.}
\label{fig:real_data}
\end{figure}

\subsection{Data Processing and Annotation Details}
\label{appendix:data_processing}
The actions from workflow video are aligned with steps documented by each professional. Claude was used to map video timestamps to the corresponding steps, after which professionals reviewed and corrected the alignments. We then represented each workflow as a directed graph, where each node corresponds to a human or AI action and is associated with a workflow step. Nodes with out-degree greater than one denote a branch into concurrent sub-paths, and nodes with in-degree greater than one denote a merge of previously concurrent
branches back into a single line of work. The authors and professionals jointly reviewed and refined these graphs. Graphs can include branches when activities occur in parallel and loops when the same action is repeated, allowing the representation to capture non-linear workflow structures as shown in Figure~\ref{fig:p4_graph}. After graph construction, two human annotators annotated how agent outputs were incorporated into the workflow. For each observed response to an agent output, annotators identified the source node, corresponding to the human action that initiated the request, and the agent node, corresponding to the resulting agent response. They then identified the subsequent human node in which the response was acted upon and labeled the user's response as \adopt, \modify, or \reject. \adopt indicates that the AI output was incorporated without modification; \modify indicates that it was incorporated with changes; and \reject indicates that the user explicitly rejected the output or did not incorporate it into the workflow. Because an AI output may be acted upon later in the workflow rather than immediately after it is produced, these annotations explicitly link each user response to the AI output it concerns. 

\begin{table}[t]
\centering
\caption{Real examples of \adopt, \modify, and \reject responses from \dataset, drawn from a music producer workflow, architecture workflow, and a video-production workflow respectively.}
\label{tab:response-examples-appendix}
\footnotesize
\resizebox{\linewidth}{!}{%
\begin{tabular}{l p{4.3cm} p{4.3cm} p{4.8cm} p{2.2cm}}
\toprule
\rowcolor{headerbg}
\textbf{Response} & \textbf{Source Node Action} & \textbf{Target AI Node Action} & \textbf{Current Node Action} & \textbf{Software} \\
\midrule
\multirow{2}{*}{\adopt}
& Request Afrobeats/techno background music in ElevenLabs (67:33--68:30)
& ElevenLabs generates Afrobeats instrumental candidates (69:50--70:04)
& Capture the ElevenLabs track and import it into the DaVinci Resolve project (72:05--74:36)
& ElevenLabs, DaVinci Resolve \\ \\
& Ask Claude to add a faith-based closing line to the voiceover script (13:07--13:41)
& Claude adds the requested faith-based reference to the script (34:17--34:22)
& Generate the final voiceover in ElevenLabs and import it into DaVinci Resolve (34:22--36:30)
& Claude, ElevenLabs, DaVinci Resolve \\
\midrule
\multirow{2}{*}{\modify}
& Upload the existing shell plan and request a full room layout (24:10--25:28)
& Claude delivers a dimensioned floor-plan layout for the lobby/storage circulation (51:21--56:20)
& Rebuild walls/partitions in Revit, keeping the liked zoning but fixing lobby/storage circulation (58:40--72:59)
& Revit \\ \\
& Ask whether CapCut's AI Relight tool requires the desktop app (63:49--63:55)
& Claude explains AI Relight also works in the web version (64:01--64:04)
& Retry AI Relight in the browser; still blocked by the desktop-app requirement (64:04--65:17)
& CapCut \\
\midrule
\multirow{2}{*}{\reject}
& Ask Kling AI to generate a Zanzibar schoolyard video clip (36:30--39:04)
& Kling AI generates the requested video clip (39:45--41:17)
& Judge the clip's audio unusable and abandon it, out of credits to iterate further (41:17--42:06)
& Kling AI \\ \\
& Ask ChatGPT to generate the same video clip (40:09--40:28)
& ChatGPT explains it cannot generate video and offers a still image instead (40:28--41:03)
& Review ChatGPT's limitation and move to another tool (41:03--41:17)
& ChatGPT \\
\bottomrule
\end{tabular}%
}
\vspace{-10pt}
\end{table}

\subsubsection{Annotation Guidelines}
\label{app:user-response}

Each workflow graph records not only the sequence of user prompts and Claude
responses, but also how each Claude response was incorporated into the
user's workflow. This is captured by the \texttt{user\_response} field,
which is added only to the user node that represents the adopt, modify, or
reject decision.

\texttt{user\_response} takes the
following form:
\begin{verbatim}
"user_response": [["source_node_id", "claude_node_id", "label",
                    "source_node_time", "claude_node_time",
                    "user_response_time"]]
\end{verbatim}
This ensures that each adopt, modify, or reject annotation is linked to the
correct occurrence of the repeated node, rather than being ambiguous across
its several instances.

\paragraph{Field definitions.}
\begin{itemize}
    \item \textbf{\texttt{source\_node\_id}} --- The user node ID that
    initiated the Claude request for this step of the workflow. This node
    may originate from a previous step: all subsequent user responses within
    the same workflow continue to reference it as the source, even as the
    user issues additional requests. \texttt{source\_node\_id} is left empty
    when the user reuses a prior Claude response within a new step whose
    narrative is not coherently linked back to the node from that earlier
    step.
    \item \textbf{\texttt{claude\_node\_id}} --- The Claude node ID
    containing the response to the request.
    \item \textbf{\texttt{label}} --- One of \texttt{adopt}, \texttt{modify},
    or \texttt{reject}, indicating how the user incorporated the Claude
    response into the workflow. On rare occasions a single node carries more
    than one label; in that case, each label is recorded as a separate
    tuple with its own \texttt{source\_node\_id} and \texttt{claude\_node\_id}.
\end{itemize}

\paragraph{Label definitions.}
\begin{itemize}
    \item \adopt: The user incorporates Claude's response into
    the workflow without changes.
    \item \modify: The user incorporates Claude's response into
    the workflow after modifying it.
    \item \reject: The user does not incorporate Claude's response into the workflow. This label is applied only when the user explicitly states that they are rejecting the response.
\end{itemize}

Examples of how professionals adopt, modify, or reject agent outputs are shown in Figure~\ref{fig:real_data}. A request from a previous step can undergo multiple refinements before being adopted in a later step.

\subsection{Details for Constructing and Annotating Graphs from Videos}

\begin{figure*}[t]
    \centering
    \begin{subfigure}[t]{\textwidth}
        \centering
        \includegraphics[width=\linewidth]{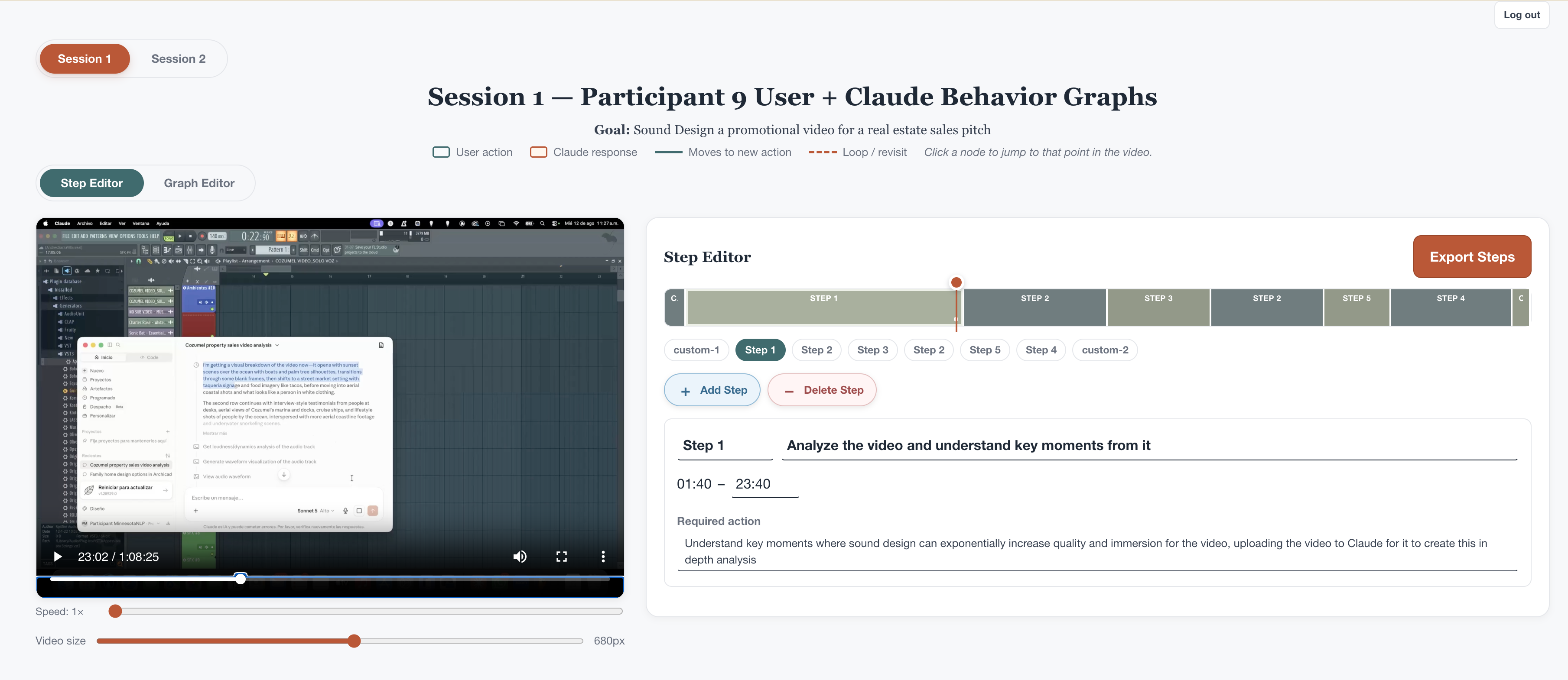}
        \caption{The interface for professionals to align steps and timestamps.}
        \label{fig:annotation-step}
    \end{subfigure}
    \hfill
    \begin{subfigure}[t]{\textwidth}
        \centering
        \includegraphics[width=\linewidth]{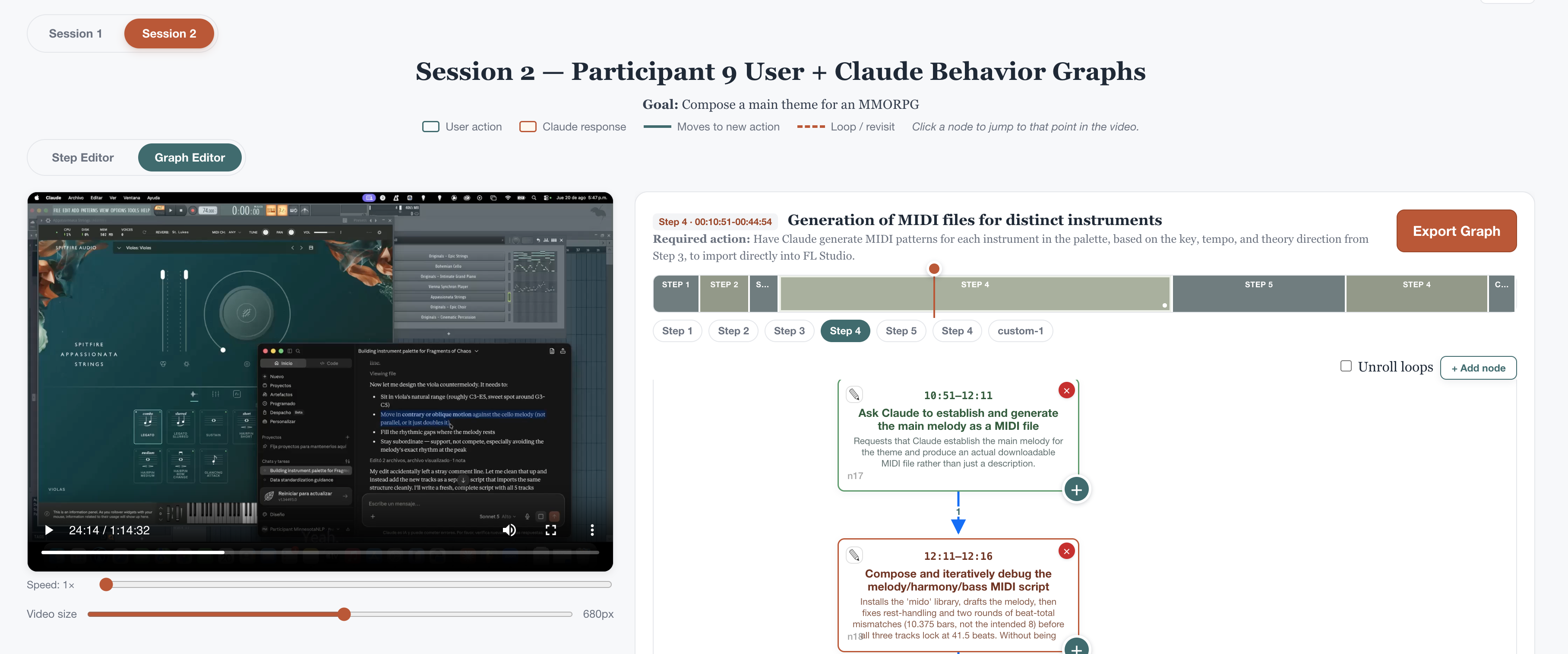}
        \caption{The interface for professionals to verify and revise workflow.}
        \label{fig:annotation-graph}
    \end{subfigure}
\caption{A webpage for professionals to align timestamps with steps and verify the generated graphs. Professionals first verify workflow steps and their corresponding video timestamps (top), and subsequently inspect the workflow graph for each step (bottom).}
\label{fig:annotation-interface}
\end{figure*}

Figure~\ref{fig:annotation-interface} shows our workflow annotation interface. Professionals first align their documented workflow steps with the corresponding video timestamps (Figure~\ref{fig:annotation-step}). Step names are taken from their workflow documents and initially mapped to the video by Claude Code Sonnet 5.5. After professionals correct the step boundaries, we extract a workflow graph for each step and present it for verification (Figure~\ref{fig:annotation-graph}), where professionals confirm whether the high-level workflow structure is consistent with their actual process. The prompt for generating the initial graph by Claude is shown below.

Four members of the research team then reconstruct and annotate the full workflow graphs, as the automatically extracted graphs frequently contain errors in action labels, edges, timestamps, and actor assignments. Annotators use the screen recordings, think-aloud transcripts, agent chat logs, and input/output files to ground these corrections. When an action or dependency remains ambiguous, the research team consults the corresponding professional for clarification.
\newpage
\begin{promptbox}[title=Workflow Graph Extraction]
You are given:

\begin{enumerate}
  \item A workflow video.
  \item A list of workflow steps with timestamps.
  \item The workflow goal.
  \item A transcript of the video (.srt).
  \item A workflow planning document (.docx).
  \item A chat log between human user and Claude
\end{enumerate}
\vspace{2em}

Your task is to extract a behavioral workflow graph, capturing goal-directed activities performed by the human user and AI throughout the workflow.
\newline\newline
\textbf{Behavioral nodes}
\newline\newline
A behavioral node represents what the actor is trying to accomplish, not the computer use actions.
\newline\newline
You must focus on the underlying work activity, intention, or objective and include relevant context, such as the object being worked on or the purpose of the activity.
\newline\newline
Examples:
\newline\newline
* Define objective\newline
* Plan approach\newline
* Seek information\newline
* Gather evidence\newline
* Analyze information\newline
* Compare alternatives\newline
* Refine requirements\newline
* Verify findings\newline
* Make decision\newline
* Implement solution\newline
* Validate outcome\newline
* Revise solution
\newline\newline
These are examples only, not a fixed taxonomy. Use a more specific label when the context supports it.
\newline\newline
For example, prefer:
\newline\newline
\textgreater{} ``Evaluate candidate methods for the analysis''
\newline\newline
over:
\newline\newline
\textgreater{} ``Read information''
\newline\newline
Do not create nodes for low-level actions such as clicking, typing, scrolling, switching tabs, opening files, copying text, or reading an AI response. Similarly, do not create nodes simply because the actor is using a particular software application.
\newline\newline
Only include a software interaction when it represents a meaningful work activity.
\newline\newline
A behavioral graph corresponds to the workflow step at one time-stamp range. For example, if the workflow is ``Step 2'' $\rightarrow$ ``Step 3'' $\rightarrow$ ``Step 2'', you must have 3 different behavioral graphs.
\newline\newline
Use the video and workflow context to infer the actor's behavior. Do not invent intentions that are unsupported by the evidence.
\newline\newline
\textbf{Actors}
\newline\newline
Include meaningful behavioral activities performed by both:
\newline\newline
* \textasciigrave user\textasciigrave\newline
* \textasciigrave ai brand\textasciigrave
\newline\newline
For AI, indicate whether it is \textasciigrave claude\textasciigrave, \textasciigrave grok\textasciigrave, or other types of AI. Most of AI here are Claude, but some users may use other AI too, so please check carefully.
\newline\newline
For AI behavior, describe the meaningful contribution to the user's workflow rather than individual tool calls.
\newline\newline
\textbf{Softwares}
\newline\newline
Infer the softwares being used in the screen.
\newline\newline
\textbf{Graph structure}
\newline\newline
Construct a directed graph:
\newline\newline
* Nodes represent behavioral activities.\newline
* Edges represent transitions between activities.\newline
* Include loops when the workflow revisits an earlier behavioral activity or repeatedly follows a behavioral cycle.\newline
* Use edge types such as \textasciigrave sequential\textasciigrave\ or \textasciigrave loop\textasciigrave\ when appropriate.
\newline\newline
\textbf{Output}
\newline\newline
Return only valid JSON following the format below:
\newline\newline
\begin{verbatim}
{
  "session_id": "...",
  "participant_id": "...",
  "title": "...",
  "goal": "...",
  "steps": [
   {
      "key": "seg0a",
      "badge": "Step ID (e.g., Step-1, Custom-1, and so on",
      "step_name": "Step name from steps.json",
      "step_action": "Step action from steps.json",
      "time_range": "Step time range from steps.json",
      "start_seconds": 0,
      "end_seconds": 600,
      "nodes": [
      {
          "id": "n1",
          "actor": "user",
          "label": "The behavioral node label",
          "detail": "A brief description of the behavioral activity.",
          "time": ["HH:MM:SS1-HH:MM:SS2", ...],
          "actor_type": "user",
          "softwares": "...",
          "source_steps": "..."
        },
        {
          "id": "n2",
          "actor": "claude or grok or gemini or chatgpt",
          "label": "The behavioral node label for an AI-authored 
          action",
          "detail": "A brief description of the behavioral activity.",
          "time": ["HH:MM:SS1-HH:MM:SS2", ...],
          "actor_type": "ai",
          "softwares": "...",
          "source_steps": "..."
        }
      ],
      "edges": [
        {
          "from": "node_id",
          "to": "node_id",
          "type": "sequential"
        }
      ]
    }
  ]
}
\end{verbatim}

Note on \textasciigrave actor\textasciigrave\ vs.\ \textasciigrave actor\_type\textasciigrave: \textasciigrave actor\textasciigrave\ is the specific actor performing the node --- \textasciigrave user\textasciigrave, or the AI's actual name (\textasciigrave claude\textasciigrave, \textasciigrave grok\textasciigrave, \textasciigrave chatgpt\textasciigrave, \textasciigrave gemini\textasciigrave, etc.). \textasciigrave actor\_type\textasciigrave\ is always just the coarse category --- \textasciigrave user\textasciigrave\ or \textasciigrave ai\textasciigrave\ --- never an AI brand name. Do not swap these: an AI node must have \textasciigrave actor\_type: "ai"\textasciigrave\ (with the brand name in \textasciigrave actor\textasciigrave), not \textasciigrave actor\_type: "claude"\textasciigrave.
\newline\newline
For each node:
\newline\newline
* \textasciigrave label\textasciigrave: the behavioral node label\newline
* \textasciigrave detail\textasciigrave: brief contextual description of the activity. Only include 2--3 short sentences!\newline
* \textasciigrave time\textasciigrave: [the start and ending timestamps of that behavior]\newline
* \textasciigrave source\_steps\textasciigrave: workflow step(s) supporting the node
\newline\newline
Before returning the result, verify that every node represents a meaningful goal-directed activity rather than a low-level interface action, and that the graph accurately reflects the workflow's behavioral sequence and iterations.
\newline\newline
\textbf{You must watch the video to correctly create the graph! Don't rely on the .srt transcript since it can be wrong! Save the video frames for later use.}
\newline
\newline
Double-check if there are some loops in the graph. If there are loops, make sure that it is presented in the JSON file correctly. For example, if the loop is node A $\rightarrow$ node B $\rightarrow$ node A, there should be two edges (from node A to node B and node B back to node A). A loop can also be A $\rightarrow$ B $\rightarrow$ C $\rightarrow$ A $\rightarrow$ D, so please check carefully.
\end{promptbox}

\subsubsection{How Humans Construct the Graph}
While an LLM was used to generate initial workflow graphs from the video, think-aloud transcript, workflow document, step mappings, and AI chat logs collected from professionals, the research team manually reviewed and substantially reconstructed the graphs, often effectively from scratch.

\subsection{Prompts for Evaluating MLLMs}

We run each experiment under seven input conditions, one for every non-empty combination of video, workflow graph, and chat. The combined conditions reuse the parts of the single-source prompts, so we show only the video-only, graph-only, and chat-only prompts.

\subsubsection{Experiment 1a: Adoption Behavior Prediction Prompt (with target timestamp)}
\label{app:prompt-setting1}

\begin{promptbox}[title=Experiment 1a: video-only]
You are given a video snippet of \{PROFESSION\_PHRASE\} working with an AI agent toward the following goal:
\newline\newline 
Goal: \{GOAL\}
\newline\newline
The workflow consists of multiple steps that contribute to achieving this goal. The AI may produce an output in one step, and the user may act on that output either within the same step or in a later step.
\newline\newline
AI Action Step: \{AI\_ACTION\_STEP\}\\
User Action Step: \{USER\_ACTION\_STEP\}
\newline\newline
Target AI Action Timestamp: \{START\_TIME\}--\{END\_TIME\}
\newline\newline
You must predict how the user will respond to the output of the target AI action:
\newline\newline
* ADOPT: Use the AI output as provided, without any modification.\\
* MODIFY: Use the AI output but modify it in any way, including minor edits, formatting changes, or other adjustments.\\
* REJECT/BYPASS: Do not use the AI output; reject, ignore, replace, or proceed without it.
\newline\newline
Use the workflow context available to make your prediction.
\newline\newline
Output exactly one label:\\
\textasciigrave ADOPT\textasciigrave, \textasciigrave MODIFY\textasciigrave, or \textasciigrave REJECT/BYPASS\textasciigrave
\end{promptbox}

\begin{promptbox}[title=Experiment 1a: graph-only]
You are given the workflow graph of \{PROFESSION\_PHRASE\} working with an AI agent toward the following goal (no video is provided for this task -- rely only on the workflow graph below):
\newline\newline
Goal: \{GOAL\}
\newline\newline
The workflow consists of multiple steps that contribute to achieving this goal. The AI may produce an output in one step, and the user may act on that output either within the same step or in a later step.
\newline\newline
AI Action Step: \{AI\_ACTION\_STEP\}\\
User Action Step: \{USER\_ACTION\_STEP\}
\newline\newline
Workflow graph for this portion of the workflow is defined as follows (each node is one action taken by the human or the agent; edges show the order they happened in):
\newline\newline
\texttt{[n0]} (\{actor\}) \{label\}: \{details\}\\
\texttt{[n1]} (\{actor\}) \{label\}: \{details\}\\
...
\newline\newline
A directed edge X $\rightarrow$ Y means that action Y follows from or depends on action X. The graph is directed and may contain branches.
\newline\newline
Edges (order of actions):\\
n0 -> n1\\
n1 -> n2\\
...
\newline\newline
You must predict how the user will respond to the output of the target AI action (the node above matching the ``AI Action Step'' described earlier):
\newline\newline
* ADOPT: Use the AI output as provided, without any modification.\newline
* MODIFY: Use the AI output but modify it in any way, including minor edits, formatting changes, or other adjustments.\newline
* REJECT/BYPASS: Do not use the AI output; reject, ignore, replace, or proceed without it.
\newline\newline
Use the workflow context available to make your prediction.
\newline\newline
Output exactly one label:\\
\textasciigrave ADOPT\textasciigrave, \textasciigrave MODIFY\textasciigrave, or \textasciigrave REJECT/BYPASS\textasciigrave
\end{promptbox}
\begin{promptbox}[title=Experiment 1a: chat-only]
You are given the chat of \{PROFESSION\_PHRASE\} working with an AI agent toward the following goal (no video is provided for this task -- rely only on the chat below):
\newline\newline
Goal: \{GOAL\}
\newline\newline
The workflow consists of multiple steps that contribute to achieving this goal. The AI may produce an output in one step, and the user may act on that output either within the same step or in a later step.
\newline\newline
AI Action Step: \{AI\_ACTION\_STEP\}\\
User Action Step: \{USER\_ACTION\_STEP\}
\newline\newline
Chat between the human and the AI agent during this portion of the workflow, in chronological order (each message from the human is wrapped in \textasciigrave$<$human$>$\textasciigrave\ tags, each message from the AI agent in \textasciigrave$<$ai$>$\textasciigrave\ tags):
\newline\newline
\texttt{<human>}\\
\{PROMPT\}\\
\texttt{</human>}
\newline\newline
\texttt{<ai>}\\
\{AI\_RESPONSE\}\\
\texttt{</ai>}
\newline\newline
\{...ADDITIONAL TURNS...\}
\newline\newline
You must predict how the user will respond to the output of the target AI action:
\newline\newline
* ADOPT: Use the AI output as provided, without any modification.\newline
* MODIFY: Use the AI output but modify it in any way, including minor edits, formatting changes, or other adjustments.\newline
* REJECT/BYPASS: Do not use the AI output; reject, ignore, replace, or proceed without it.
\newline\newline
Use the workflow context available to make your prediction.
\newline\newline
Output exactly one label:\\
\textasciigrave ADOPT\textasciigrave, \textasciigrave MODIFY\textasciigrave, or \textasciigrave REJECT/BYPASS\textasciigrave
\end{promptbox}

\subsubsection{Experiment 1b \& 1c \& 1d: Adoption Behavior Forecasting Prompt (clip trimmed to AI action)}
\label{app:prompt-setting3}

\begin{promptbox}[title=Experiment 1b \& 1c \& 1d: video-only]
You are given a video snippet of \{PROFESSION\_PHRASE\} working with an AI agent toward the following goal:
\newline\newline
Goal: \{GOAL\}
\newline\newline
The workflow consists of multiple steps that contribute to achieving this goal. The AI may produce an output in one step, and the user may act on that output either within the same step or in a later step.
\newline\newline
AI Action Step: \{AI\_ACTION\_STEP\}
\newline\newline
You must predict how the user will respond to the output of this AI action:
\newline\newline
* ADOPT: Use the AI output as provided, without any modification.\\
* MODIFY: Use the AI output but modify it in any way, including minor edits, formatting changes, or other adjustments.\\
* REJECT/BYPASS: Do not use the AI output; reject, ignore, replace, or proceed without it.
\newline\newline
Use the workflow context available to make your prediction.
\newline\newline
Output exactly one label:\\
\textasciigrave ADOPT\textasciigrave, \textasciigrave MODIFY\textasciigrave, or \textasciigrave REJECT/BYPASS\textasciigrave
\end{promptbox}

\begin{promptbox}[title=Experiment 1b \& 1c \& 1d: graph-only]
You are given the workflow graph of \{PROFESSION\_PHRASE\} working with an AI agent toward the following goal (no video is provided for this task -- rely only on the workflow graph below):
\newline\newline
Goal: \{GOAL\}
\newline\newline
The workflow consists of multiple steps that contribute to achieving this goal. The AI may produce an output in one step, and the user may act on that output either within the same step or in a later step.
\newline\newline
AI Action Step: \{AI\_ACTION\_STEP\}
\newline\newline
Workflow graph for this portion of the workflow is defined as follows (each node is one action taken by the human or the agent; edges show the order they happened in):
\newline\newline
\texttt{[n0]} (\{actor\}) \{label\}: \{details\}\\
\texttt{[n1]} (\{actor\}) \{label\}: \{details\}\\
...
\newline\newline
A directed edge X $\rightarrow$ Y means that action Y follows from or depends on action X. The graph is directed and may contain branches.
\newline\newline
Edges (order of actions):\\
n0 -> n1\\
n1 -> n2\\
...
\newline\newline
You must predict how the user will respond to the output of the target AI action (the node above matching the ``AI Action Step'' described earlier):
\newline\newline
* ADOPT: Use the AI output as provided, without any modification.\newline
* MODIFY: Use the AI output but modify it in any way, including minor edits, formatting changes, or other adjustments.\newline
* REJECT/BYPASS: Do not use the AI output; reject, ignore, replace, or proceed without it.
\newline\newline
Use the workflow context available to make your prediction.
\newline\newline
Output exactly one label:\\
\textasciigrave ADOPT\textasciigrave, \textasciigrave MODIFY\textasciigrave, or \textasciigrave REJECT/BYPASS\textasciigrave
\end{promptbox}
\begin{promptbox}[title=Experiment 1b \& 1c \& 1d: chat-only]
You are given the chat of \{PROFESSION\_PHRASE\} working with an AI agent toward the following goal (no video is provided for this task -- rely only on the chat below):
\newline\newline
Goal: \{GOAL\}
\newline\newline
The workflow consists of multiple steps that contribute to achieving this goal. The AI may produce an output in one step, and the user may act on that output either within the same step or in a later step.
\newline\newline
AI Action Step: \{AI\_ACTION\_STEP\}
\newline\newline
Chat between the human and the AI agent during this portion of the workflow, in chronological order (each message from the human is wrapped in \textasciigrave$<$human$>$\textasciigrave\ tags, each message from the AI agent in \textasciigrave$<$ai$>$\textasciigrave\ tags):
\newline\newline
\texttt{<human>}\\
\{PROMPT\}\\
\texttt{</human>}
\newline\newline
\texttt{<ai>}\\
\{AI\_RESPONSE\}\\
\texttt{</ai>}
\newline\newline
\{...ADDITIONAL TURNS...\}
\newline\newline
You must predict how the user will respond to the output of the target AI action:
\newline\newline
* ADOPT: Use the AI output as provided, without any modification.\newline
* MODIFY: Use the AI output but modify it in any way, including minor edits, formatting changes, or other adjustments.\newline
* REJECT/BYPASS: Do not use the AI output; reject, ignore, replace, or proceed without it.
\newline\newline
Use the workflow context available to make your prediction.
\newline\newline
Output exactly one label:\\
\textasciigrave ADOPT\textasciigrave, \textasciigrave MODIFY\textasciigrave, or \textasciigrave REJECT/BYPASS\textasciigrave
\end{promptbox}

\subsubsection{Experiment 2a: Collaboration Rating Prediction Prompt}
\label{app:prompt-exp2a}

\begin{promptbox}[title=Experiment 2a: video-only]
You are given a video snippet of \{PROFESSION\_PHRASE\} working with an AI agent toward the following goal:
\newline\newline
Goal: \{GOAL\}
\newline\newline
The video shows a step of a multi-step workflow for accomplishing the goal:
\newline\newline
Step: \{STEP\_NAME\}
\newline\newline
Watch how the human and the AI agent interact throughout this step. Examine whether the agent understood the task, context, and instructions; whether it adapted appropriately when given feedback or corrections; whether it reduced the human's effort, time, or cognitive load; whether it proactively suggested useful ideas or next steps; whether its output was accurate, relevant, and trustworthy with minimal correction needed; whether the interaction felt smooth, with the agent contributing at the right moments without creating unnecessary work; and whether the human was able to guide, redirect, or override the agent's actions when desired.
\newline\newline
Based on this, rate the overall collaboration experience for this step on a scale from 1 to 5, where 1 means that the user is not satisfied with the collaboration experience with the AI agent and 5 means that the user is satisfied with the collaboration experience with the AI agent.
\newline\newline
Output exactly one integer, with no other text:\\
1, 2, 3, 4, or 5
\end{promptbox}

\begin{promptbox}[title=Experiment 2a: graph-only]
You are given a video snippet of \{PROFESSION\_PHRASE\} working with an AI agent toward the following goal:
\newline\newline
Goal: \{GOAL\}
\newline\newline
The video shows a step of a multi-step workflow for accomplishing the goal:
\newline\newline
Step: \{STEP\_NAME\}
\newline\newline
The workflow consists of multiple steps that contribute to achieving this goal. The AI may produce an output in one step, and the user may act on that output either within the same step or in a later step.
\newline\newline
Workflow graph for this portion of the workflow is defined as follows (each node is one action taken by the human or the agent; edges show the order they happened in):
\newline\newline
\texttt{[n0]} (\{actor\}) \{label\}: \{details\}\\
\texttt{[n1]} (\{actor\}) \{label\}: \{details\}\\
...
\newline\newline
A directed edge X $\rightarrow$ Y means that action Y follows from or depends on action X. The graph is directed and may contain branches.
\newline\newline
Edges (order of actions):\\
n0 -> n1\\
n1 -> n2\\
...
\newline\newline
Read the graph above. Examine whether the agent understood the task, context, and instructions; whether it adapted appropriately when given feedback or corrections; whether it reduced the human's effort, time, or cognitive load; whether it proactively suggested useful ideas or next steps; whether its output was accurate, relevant, and trustworthy with minimal correction needed; whether the interaction felt smooth, with the agent contributing at the right moments without creating unnecessary work; and whether the human was able to guide, redirect, or override the agent's actions when desired.
\newline\newline
Based on this, rate the overall collaboration experience for this step on a scale from 1 to 5, where 1 means that the user is not satisfied with the collaboration experience with the AI agent and 5 means that the user is satisfied with the collaboration experience with the AI agent.
\newline\newline
Output exactly one integer, with no other text:\\
1, 2, 3, 4, or 5
\end{promptbox}

\begin{promptbox}[title=Experiment 2a: chat-only]
You are given a video snippet of \{PROFESSION\_PHRASE\} working with an AI agent toward the following goal:
\newline\newline
Goal: \{GOAL\}
\newline\newline
The video shows a step of a multi-step workflow for accomplishing the goal:
\newline\newline
Step: \{STEP\_NAME\}
\newline\newline
Chat between the human and the AI agent during this portion of the workflow, in chronological order (each message from the human is wrapped in \textasciigrave$<$human$>$\textasciigrave\ tags, each message from the AI agent in \textasciigrave$<$ai$>$\textasciigrave\ tags):
\newline\newline
\texttt{<human>}\\
\{PROMPT\}\\
\texttt{</human>}
\newline\newline
\texttt{<ai>}\\
\{AI\_RESPONSE\}\\
\texttt{</ai>}
\newline\newline
\{...ADDITIONAL TURNS...\}
\newline\newline
Read the chat above. Examine whether the agent understood the task, context, and instructions; whether it adapted appropriately when given feedback or corrections; whether it reduced the human's effort, time, or cognitive load; whether it proactively suggested useful ideas or next steps; whether its output was accurate, relevant, and trustworthy with minimal correction needed; whether the interaction felt smooth, with the agent contributing at the right moments without creating unnecessary work; and whether the human was able to guide, redirect, or override the agent's actions when desired.
\newline\newline
Based on this, rate the overall collaboration experience for this step on a scale from 1 to 5, where 1 means that the user is not satisfied with the collaboration experience with the AI agent and 5 means that the user is satisfied with the collaboration experience with the AI agent.
\newline\newline
Output exactly one integer, with no other text:\\
1, 2, 3, 4, or 5
\end{promptbox}

\subsubsection{Experiment 2b: Output Quality Rating Prediction Prompt}
\label{app:prompt-exp2b}

\begin{promptbox}[title=Experiment 2b: video-only]
You are given a video snippet of \{PROFESSION\_PHRASE\} working with an AI agent toward the following goal:
\newline\newline
Goal: \{GOAL\}
\newline\newline
The video shows a step of a multi-step workflow for accomplishing the goal:
\newline\newline
Step: \{STEP\_NAME\}
\newline\newline
Watch the output the AI agent produced and how it was used during this step. Examine whether the output was accurate, relevant, and directly usable for the task; whether it required significant correction, rework, or was discarded; and whether it met the specific requirements of this step.
\newline\newline
Based on this, rate the output quality for this step on a scale from 1 to 5, where 1 means that the user is not satisfied with the output quality and 5 means that the user is satisfied with the output quality.
\newline\newline
Output exactly one integer, with no other text:\\
1, 2, 3, 4, or 5
\end{promptbox}
\begin{promptbox}[title=Experiment 2b: graph-only]
You are given a video snippet of \{PROFESSION\_PHRASE\} working with an AI agent toward the following goal:
\newline\newline
Goal: \{GOAL\}
\newline\newline
The video shows a step of a multi-step workflow for accomplishing the goal:
\newline\newline
Step: \{STEP\_NAME\}
\newline\newline
Workflow graph for this portion of the workflow is defined as follows (each node is one action taken by the human or the agent; edges show the order they happened in):
\newline\newline
\texttt{[n0]} (\{actor\}) \{label\}: \{details\}\\
\texttt{[n1]} (\{actor\}) \{label\}: \{details\}\\
...
\newline\newline
A directed edge X $\rightarrow$ Y means that action Y follows from or depends on action X. The graph is directed and may contain branches.
\newline\newline
Edges (order of actions):\\
n0 -> n1\\
n1 -> n2\\
...
\newline\newline
Read the graph above, focusing on the output the AI agent produced and how it was used during this step. Examine whether the output was accurate, relevant, and directly usable for the task; whether it required significant correction, rework, or was discarded; and whether it met the specific requirements of this step.
\newline\newline
Based on this, rate the output quality for this step on a scale from 1 to 5, where 1 means that the user is not satisfied with the output quality and 5 means that the user is satisfied with the output quality.
\newline\newline
Output exactly one integer, with no other text:\\
1, 2, 3, 4, or 5
\end{promptbox}
\begin{promptbox}[title=Experiment 2b: chat-only]
You are given a video snippet of \{PROFESSION\_PHRASE\} working with an AI agent toward the following goal:
\newline\newline
Goal: \{GOAL\}
\newline\newline
The video shows a step of a multi-step workflow for accomplishing the goal:
\newline\newline
Step: \{STEP\_NAME\}
\newline\newline
Chat between the human and the AI agent during this portion of the workflow, in chronological order (each message from the human is wrapped in \textasciigrave$<$human$>$\textasciigrave\ tags, each message from the AI agent in \textasciigrave$<$ai$>$\textasciigrave\ tags):
\newline\newline
\texttt{<human>}\\
\{PROMPT\}\\
\texttt{</human>}
\newline\newline
\texttt{<ai>}\\
\{AI\_RESPONSE\}\\
\texttt{</ai>}
\newline\newline
\{...ADDITIONAL TURNS...\}
\newline\newline
Read the chat above, focusing on the output the AI agent produced and how it was used during this step. Examine whether the output was accurate, relevant, and directly usable for the task; whether it required significant correction, rework, or was discarded; and whether it met the specific requirements of this step.
\newline\newline
Based on this, rate the output quality for this step on a scale from 1 to 5, where 1 means that the user is not satisfied with the output quality and 5 means that the user is satisfied with the output quality.
\newline\newline
Output exactly one integer, with no other text:\\
1, 2, 3, 4, or 5
\end{promptbox}

\subsection{Model Selection}
For proprietary models, we evaluate leading video-capable multimodal models from major providers: \textbf{Gemini 3.1 Pro} and \textbf{Gemini 3.8 Flash} (Google), \textbf{GPT-6 Astra} (OpenAI), and \textbf{Claude Opus 5.5}. Gemini natively accepts video as an input. GPT-6 Astra has no native video modality, so we sample evenly-spaced frames from each clip as its input; this also lets GPT-6 Astra serve as a frame-sampled point of comparison against Gemini's genuinely video-native processing. We additionally evaluate Claude despite it likewise lacking native video input (also requiring frame sampling), because Claude is the specific agent our study participants interacted with during their recorded sessions, so evaluating it tests whether the very model embedded in the human-AI collaboration can itself judge or predict the outcome of that collaboration.

For open-weight models, we evaluate two representative video-capable Qwen3-VL variants \citep{bai2025qwen3}: \textbf{Qwen3-VL-30B-A3B-Instruct}, a mixture-of-experts model (30B total parameters, 3B active per token) included to test whether MoE-style efficiency scaling preserves video-temporal understanding, and \textbf{Qwen3-VL-32B-Instruct}, a dense model of comparable total scale that instead invests its full parameter budget in deeper visual-language fusion at every token. Comparing the two isolates the effect of the MoE-vs-dense architectural choice at a fixed parameter scale.

\subsection{Sample Size and Selection}
\textbf{Claude evaluation subsets} Claude was queried manually through its web interface, so for Experiments 1a, 2a and 2b we evaluated it on fixed subsets; Experiment 1b uses all instances. For 1a, we sampled 25 instances per label (N = 75), because the labels are imbalanced (13\% reject), and a random sample would leave too few rejects for per-class F1. For 2a, we kept every step rated 1–4 and sampled 30 of the steps rated 5 (N = 78), since 5s make up 60\% of the data. 2b uses the same steps, except six with no output-quality rating (N = 72). All subsets were drawn with a fixed seed, spread across participants

\subsection{Experiment Results}

\begin{table}[t]
\centering
\begin{tabular}{lrrrr}
\toprule
\textbf{Experiment} & \textbf{\# Clips} & \textbf{Mean $\pm$ std} & \textbf{Min} & \textbf{Max} \\
\midrule
Experiment 1a & 692 & 14.0{\scriptsize$\pm$13.7} & 0.8 & 73.0 \\
Experiment 1b & 691 & 8.8{\scriptsize$\pm$11.7} & 0.2 & 72.0 \\
Experiment 1c & 638 & 37.9{\scriptsize$\pm$22.7} & 1.5 & 117.1 \\
Experiment 1d: last1 & 114 & 10.1{\scriptsize$\pm$9.6} & 0.4 & 53.6 \\
Experiment 1d: last2 & 114 & 20.7{\scriptsize$\pm$11.7} & 3.6 & 72.7 \\
Experiment 1d: last3 & 114 & 32.3{\scriptsize$\pm$13.8} & 7.0 & 87.8 \\
Experiment 1d: full & 114 & 58.6{\scriptsize$\pm$17.3} & 26.1 & 117.1 \\
Experiment 2a & 350 & 14.9{\scriptsize$\pm$11.0} & 0.8 & 66.5 \\
Experiment 2b & 373 & 14.9{\scriptsize$\pm$11.0} & 0.8 & 66.5 \\
\bottomrule
\end{tabular}
\caption{Length (in minutes) of the video clips given to the models in each experiment.}
\label{tab:clip_lengths}
\end{table}

Results for Experiment 1a (recognizing the user's response to the AI's action) are shown in
Table~\ref{1a_results_full_table}, and results for Experiment 1b (forecasting the user's
response) are shown in Table~\ref{1b_results_full_table}.
Results for Experiments 2a and 2b are given in
Table~\ref{2a_results_full_table} and Table~\ref{2b_results_full_table}, respectively.


\begin{table}[t]
\centering
\resizebox{\linewidth}{!}{%
\setlength{\tabcolsep}{4pt}
\begin{tabular}{llccc|ccc}
\toprule
& & \multicolumn{3}{c}{\textbf{Macro}} & \multicolumn{3}{c}{\textbf{Per-class F1}} \\
\cmidrule(lr){3-5} \cmidrule(lr){6-8}
\textbf{Model} & \textbf{Setting} &
\textbf{P} & \textbf{R} & \textbf{F1} &
\textbf{Adopt} & \textbf{Modify} & \textbf{Reject} \\
\midrule
\multirow{3}{*}{GPT-6-Astra} & video-only & \textbf{0.742} & 0.725 & \textbf{0.713} & 0.804 & 0.602 & \textbf{0.732} \\
 & chat-only & 0.648 & 0.673 & 0.658 & 0.777 & \textbf{0.644} & 0.552 \\
 & video-graph-chat & 0.699 & \textbf{0.742} & 0.673 & \textbf{0.812} & 0.480 & 0.727 \\
\midrule
\multirow{3}{*}{Claude} & video-only & \textbf{0.736} & \textbf{0.648} & \textbf{0.676} & \textbf{0.843} & \textbf{0.556} & 0.630 \\
 & chat-only & 0.685 & 0.584 & 0.618 & 0.831 & 0.449 & 0.574 \\
 & video-graph-chat & 0.688 & 0.641 & 0.659 & 0.835 & 0.505 & \textbf{0.638} \\
\midrule
\multirow{7}{*}{Gemini-3.1-Pro-Preview} & video-only & 0.616{\scriptsize$\pm$0.005} & 0.622{\scriptsize$\pm$0.009} & 0.617{\scriptsize$\pm$0.007} & 0.795{\scriptsize$\pm$0.005} & 0.479{\scriptsize$\pm$0.014} & 0.577{\scriptsize$\pm$0.008} \\
 & graph-only & 0.683{\scriptsize$\pm$0.003} & 0.702{\scriptsize$\pm$0.005} & 0.689{\scriptsize$\pm$0.004} & 0.823{\scriptsize$\pm$0.002} & 0.533{\scriptsize$\pm$0.004} & 0.711{\scriptsize$\pm$0.007} \\
 & chat-only & 0.589{\scriptsize$\pm$0.004} & 0.617{\scriptsize$\pm$0.006} & 0.600{\scriptsize$\pm$0.005} & 0.764{\scriptsize$\pm$0.005} & 0.494{\scriptsize$\pm$0.018} & 0.543{\scriptsize$\pm$0.007} \\
 & video-graph & 0.690{\scriptsize$\pm$0.010} & \textbf{0.712}{\scriptsize$\pm$0.008} & \textbf{0.697}{\scriptsize$\pm$0.009} & \textbf{0.830}{\scriptsize$\pm$0.007} & \textbf{0.557}{\scriptsize$\pm$0.011} & 0.703{\scriptsize$\pm$0.017} \\
 & video-chat & 0.618{\scriptsize$\pm$0.009} & 0.633{\scriptsize$\pm$0.003} & 0.621{\scriptsize$\pm$0.006} & 0.784{\scriptsize$\pm$0.008} & 0.493{\scriptsize$\pm$0.007} & 0.587{\scriptsize$\pm$0.014} \\
 & graph-chat & \textbf{0.691}{\scriptsize$\pm$0.009} & 0.693{\scriptsize$\pm$0.009} & 0.682{\scriptsize$\pm$0.007} & 0.804{\scriptsize$\pm$0.005} & 0.525{\scriptsize$\pm$0.016} & \textbf{0.716}{\scriptsize$\pm$0.006} \\
 & video-graph-chat & 0.675{\scriptsize$\pm$0.007} & 0.673{\scriptsize$\pm$0.006} & 0.665{\scriptsize$\pm$0.005} & 0.804{\scriptsize$\pm$0.006} & 0.518{\scriptsize$\pm$0.006} & 0.671{\scriptsize$\pm$0.008} \\
\midrule
\multirow{7}{*}{Gemini-3.8-Flash} & video-only & 0.588{\scriptsize$\pm$0.009} & 0.588{\scriptsize$\pm$0.009} & 0.587{\scriptsize$\pm$0.010} & 0.809{\scriptsize$\pm$0.002} & 0.425{\scriptsize$\pm$0.020} & 0.527{\scriptsize$\pm$0.008} \\
 & graph-only & 0.699{\scriptsize$\pm$0.011} & 0.700{\scriptsize$\pm$0.007} & 0.699{\scriptsize$\pm$0.009} & \textbf{0.849}{\scriptsize$\pm$0.003} & \textbf{0.566}{\scriptsize$\pm$0.006} & 0.684{\scriptsize$\pm$0.018} \\
 & chat-only & 0.625{\scriptsize$\pm$0.012} & 0.632{\scriptsize$\pm$0.010} & 0.628{\scriptsize$\pm$0.011} & 0.803{\scriptsize$\pm$0.007} & 0.509{\scriptsize$\pm$0.025} & 0.571{\scriptsize$\pm$0.010} \\
 & video-graph & 0.690{\scriptsize$\pm$0.006} & 0.700{\scriptsize$\pm$0.006} & 0.694{\scriptsize$\pm$0.006} & \textbf{0.849}{\scriptsize$\pm$0.004} & 0.545{\scriptsize$\pm$0.013} & 0.689{\scriptsize$\pm$0.002} \\
 & video-chat & 0.612{\scriptsize$\pm$0.004} & 0.629{\scriptsize$\pm$0.007} & 0.620{\scriptsize$\pm$0.005} & 0.799{\scriptsize$\pm$0.006} & 0.479{\scriptsize$\pm$0.016} & 0.580{\scriptsize$\pm$0.006} \\
 & graph-chat & \textbf{0.700}{\scriptsize$\pm$0.010} & \textbf{0.706}{\scriptsize$\pm$0.012} & \textbf{0.702}{\scriptsize$\pm$0.012} & 0.842{\scriptsize$\pm$0.005} & 0.560{\scriptsize$\pm$0.020} & \textbf{0.702}{\scriptsize$\pm$0.013} \\
 & video-graph-chat & 0.689{\scriptsize$\pm$0.014} & 0.696{\scriptsize$\pm$0.015} & 0.691{\scriptsize$\pm$0.015} & 0.834{\scriptsize$\pm$0.008} & 0.565{\scriptsize$\pm$0.024} & 0.675{\scriptsize$\pm$0.020} \\
\midrule
\multirow{7}{*}{Qwen3-VL-30B-A3B-Instruct} & video-only & 0.500{\scriptsize$\pm$0.017} & 0.557{\scriptsize$\pm$0.023} & 0.511{\scriptsize$\pm$0.019} & 0.680{\scriptsize$\pm$0.009} & 0.371{\scriptsize$\pm$0.035} & 0.481{\scriptsize$\pm$0.025} \\
 & graph-only & 0.549{\scriptsize$\pm$0.001} & 0.605{\scriptsize$\pm$0.001} & 0.542{\scriptsize$\pm$0.005} & 0.723{\scriptsize$\pm$0.012} & 0.409{\scriptsize$\pm$0.013} & 0.494{\scriptsize$\pm$0.013} \\
 & chat-only & 0.540{\scriptsize$\pm$0.009} & 0.518{\scriptsize$\pm$0.002} & 0.502{\scriptsize$\pm$0.002} & 0.773{\scriptsize$\pm$0.001} & 0.306{\scriptsize$\pm$0.004} & 0.427{\scriptsize$\pm$0.002} \\
 & video-graph & 0.593{\scriptsize$\pm$0.011} & \textbf{0.611}{\scriptsize$\pm$0.010} & 0.589{\scriptsize$\pm$0.010} & 0.791{\scriptsize$\pm$0.005} & \textbf{0.426}{\scriptsize$\pm$0.018} & 0.550{\scriptsize$\pm$0.013} \\
 & video-chat & 0.583{\scriptsize$\pm$0.026} & 0.541{\scriptsize$\pm$0.018} & 0.536{\scriptsize$\pm$0.019} & 0.788{\scriptsize$\pm$0.007} & 0.327{\scriptsize$\pm$0.033} & 0.494{\scriptsize$\pm$0.018} \\
 & graph-chat & 0.586{\scriptsize$\pm$0.009} & 0.607{\scriptsize$\pm$0.011} & 0.570{\scriptsize$\pm$0.012} & 0.785{\scriptsize$\pm$0.006} & 0.382{\scriptsize$\pm$0.009} & 0.544{\scriptsize$\pm$0.024} \\
 & video-graph-chat & \textbf{0.611}{\scriptsize$\pm$0.007} & \textbf{0.611}{\scriptsize$\pm$0.001} & \textbf{0.591}{\scriptsize$\pm$0.005} & \textbf{0.806}{\scriptsize$\pm$0.004} & 0.401{\scriptsize$\pm$0.012} & \textbf{0.568}{\scriptsize$\pm$0.002} \\
\midrule
\multirow{7}{*}{Qwen3-VL-32B-Instruct} & video-only & 0.524{\scriptsize$\pm$0.009} & 0.529{\scriptsize$\pm$0.010} & 0.337{\scriptsize$\pm$0.007} & 0.123{\scriptsize$\pm$0.007} & 0.361{\scriptsize$\pm$0.004} & 0.526{\scriptsize$\pm$0.014} \\
 & graph-only & 0.589{\scriptsize$\pm$0.003} & 0.558{\scriptsize$\pm$0.001} & 0.476{\scriptsize$\pm$0.002} & 0.430{\scriptsize$\pm$0.004} & 0.365{\scriptsize$\pm$0.001} & 0.634{\scriptsize$\pm$0.005} \\
 & chat-only & 0.492{\scriptsize$\pm$0.005} & 0.476{\scriptsize$\pm$0.006} & 0.393{\scriptsize$\pm$0.001} & 0.424{\scriptsize$\pm$0.009} & 0.351{\scriptsize$\pm$0.007} & 0.403{\scriptsize$\pm$0.003} \\
 & video-graph & \textbf{0.597}{\scriptsize$\pm$0.004} & \textbf{0.615}{\scriptsize$\pm$0.002} & \textbf{0.559}{\scriptsize$\pm$0.004} & 0.621{\scriptsize$\pm$0.007} & \textbf{0.414}{\scriptsize$\pm$0.002} & \textbf{0.643}{\scriptsize$\pm$0.004} \\
 & video-chat & 0.481{\scriptsize$\pm$0.005} & 0.492{\scriptsize$\pm$0.006} & 0.416{\scriptsize$\pm$0.004} & 0.462{\scriptsize$\pm$0.005} & 0.364{\scriptsize$\pm$0.006} & 0.423{\scriptsize$\pm$0.012} \\
 & graph-chat & 0.564{\scriptsize$\pm$0.001} & 0.585{\scriptsize$\pm$0.001} & 0.528{\scriptsize$\pm$0.002} & 0.609{\scriptsize$\pm$0.003} & 0.405{\scriptsize$\pm$0.005} & 0.570{\scriptsize$\pm$0.001} \\
 & video-graph-chat & 0.533{\scriptsize$\pm$0.011} & 0.570{\scriptsize$\pm$0.009} & 0.529{\scriptsize$\pm$0.010} & \textbf{0.664}{\scriptsize$\pm$0.007} & 0.388{\scriptsize$\pm$0.007} & 0.537{\scriptsize$\pm$0.017} \\
\bottomrule
\end{tabular}%
}
\caption{Result for Experiment 1a. Mean$\pm$std over runs where a model has several; bold marks the best score within each model.}
\label{1a_results_full_table}
\end{table}


\begin{table}[t]
\centering
\resizebox{\linewidth}{!}{%
\setlength{\tabcolsep}{4pt}
\begin{tabular}{llccc|ccc}
\toprule
& & \multicolumn{3}{c}{\textbf{Macro}} & \multicolumn{3}{c}{\textbf{Per-class F1}} \\
\cmidrule(lr){3-5} \cmidrule(lr){6-8}
\textbf{Model} & \textbf{Setting} &
\textbf{P} & \textbf{R} & \textbf{F1} &
\textbf{Adopt} & \textbf{Modify} & \textbf{Reject} \\
\midrule
\multirow{3}{*}{GPT-6-Astra} & video-only & 0.539{\scriptsize$\pm$0.003} & \textbf{0.524}{\scriptsize$\pm$0.013} & \textbf{0.513}{\scriptsize$\pm$0.003} & \textbf{0.721}{\scriptsize$\pm$0.023} & 0.582{\scriptsize$\pm$0.098} & 0.235{\scriptsize$\pm$0.070} \\
 & chat-only & 0.365 & 0.371 & 0.368 & 0.695 & 0.408 & 0.000 \\
 & video-graph-chat & \textbf{0.546} & 0.512 & 0.512 & 0.696 & \textbf{0.592} & \textbf{0.250} \\
\midrule
\multirow{3}{*}{Claude} & video-only & 0.610 & 0.483 & 0.492 & 0.790 & 0.434 & 0.250 \\
 & chat-only & 0.428 & 0.414 & 0.408 & 0.770 & 0.378 & 0.076 \\
 & video-graph-chat & \textbf{0.623} & \textbf{0.494} & \textbf{0.514} & \textbf{0.809} & \textbf{0.461} & \textbf{0.272} \\
\midrule
\multirow{7}{*}{Gemini-3.1-Pro-Preview} & video-only & 0.461{\scriptsize$\pm$0.026} & 0.402{\scriptsize$\pm$0.011} & 0.391{\scriptsize$\pm$0.013} & 0.710{\scriptsize$\pm$0.012} & 0.333{\scriptsize$\pm$0.018} & 0.129{\scriptsize$\pm$0.031} \\
 & graph-only & \textbf{0.523}{\scriptsize$\pm$0.017} & \textbf{0.478}{\scriptsize$\pm$0.016} & \textbf{0.492}{\scriptsize$\pm$0.017} & 0.775{\scriptsize$\pm$0.006} & 0.390{\scriptsize$\pm$0.037} & \textbf{0.311}{\scriptsize$\pm$0.015} \\
 & chat-only & 0.399{\scriptsize$\pm$0.011} & 0.387{\scriptsize$\pm$0.016} & 0.390{\scriptsize$\pm$0.014} & 0.694{\scriptsize$\pm$0.007} & 0.313{\scriptsize$\pm$0.010} & 0.162{\scriptsize$\pm$0.027} \\
 & video-graph & 0.522{\scriptsize$\pm$0.008} & 0.469{\scriptsize$\pm$0.003} & 0.482{\scriptsize$\pm$0.004} & \textbf{0.781}{\scriptsize$\pm$0.005} & \textbf{0.404}{\scriptsize$\pm$0.013} & 0.261{\scriptsize$\pm$0.022} \\
 & video-chat & 0.431{\scriptsize$\pm$0.007} & 0.415{\scriptsize$\pm$0.004} & 0.420{\scriptsize$\pm$0.005} & 0.743{\scriptsize$\pm$0.003} & 0.320{\scriptsize$\pm$0.028} & 0.195{\scriptsize$\pm$0.018} \\
 & graph-chat & 0.488{\scriptsize$\pm$0.031} & 0.446{\scriptsize$\pm$0.023} & 0.459{\scriptsize$\pm$0.026} & 0.760{\scriptsize$\pm$0.012} & 0.357{\scriptsize$\pm$0.037} & 0.259{\scriptsize$\pm$0.033} \\
 & video-graph-chat & 0.509{\scriptsize$\pm$0.025} & 0.458{\scriptsize$\pm$0.014} & 0.472{\scriptsize$\pm$0.017} & 0.773{\scriptsize$\pm$0.009} & 0.374{\scriptsize$\pm$0.025} & 0.269{\scriptsize$\pm$0.018} \\
\midrule
\multirow{7}{*}{Gemini-3.8-Flash} & video-only & 0.437{\scriptsize$\pm$0.016} & 0.383{\scriptsize$\pm$0.008} & 0.383{\scriptsize$\pm$0.009} & 0.758{\scriptsize$\pm$0.009} & 0.218{\scriptsize$\pm$0.027} & 0.172{\scriptsize$\pm$0.013} \\
 & graph-only & \textbf{0.557}{\scriptsize$\pm$0.013} & \textbf{0.458}{\scriptsize$\pm$0.011} & \textbf{0.475}{\scriptsize$\pm$0.014} & \textbf{0.797}{\scriptsize$\pm$0.005} & 0.333{\scriptsize$\pm$0.031} & 0.293{\scriptsize$\pm$0.008} \\
 & chat-only & 0.387{\scriptsize$\pm$0.027} & 0.357{\scriptsize$\pm$0.020} & 0.351{\scriptsize$\pm$0.023} & 0.725{\scriptsize$\pm$0.008} & 0.163{\scriptsize$\pm$0.040} & 0.164{\scriptsize$\pm$0.027} \\
 & video-graph & 0.526{\scriptsize$\pm$0.020} & 0.454{\scriptsize$\pm$0.007} & 0.467{\scriptsize$\pm$0.010} & 0.779{\scriptsize$\pm$0.006} & 0.319{\scriptsize$\pm$0.013} & \textbf{0.302}{\scriptsize$\pm$0.020} \\
 & video-chat & 0.465{\scriptsize$\pm$0.030} & 0.408{\scriptsize$\pm$0.018} & 0.414{\scriptsize$\pm$0.021} & 0.767{\scriptsize$\pm$0.009} & 0.263{\scriptsize$\pm$0.021} & 0.212{\scriptsize$\pm$0.033} \\
 & graph-chat & 0.532{\scriptsize$\pm$0.022} & 0.450{\scriptsize$\pm$0.016} & 0.465{\scriptsize$\pm$0.018} & 0.786{\scriptsize$\pm$0.009} & \textbf{0.337}{\scriptsize$\pm$0.020} & 0.272{\scriptsize$\pm$0.029} \\
 & video-graph-chat & 0.522{\scriptsize$\pm$0.012} & 0.445{\scriptsize$\pm$0.001} & 0.458{\scriptsize$\pm$0.001} & 0.783{\scriptsize$\pm$0.005} & 0.319{\scriptsize$\pm$0.007} & 0.273{\scriptsize$\pm$0.002} \\
\midrule
\multirow{7}{*}{Qwen3-VL-30B-A3B-Instruct} & video-only & 0.389{\scriptsize$\pm$0.010} & 0.404{\scriptsize$\pm$0.011} & 0.364{\scriptsize$\pm$0.007} & 0.501{\scriptsize$\pm$0.004} & \textbf{0.305}{\scriptsize$\pm$0.019} & 0.285{\scriptsize$\pm$0.012} \\
 & graph-only & 0.478{\scriptsize$\pm$0.014} & \textbf{0.448}{\scriptsize$\pm$0.014} & \textbf{0.458}{\scriptsize$\pm$0.014} & 0.748{\scriptsize$\pm$0.006} & 0.246{\scriptsize$\pm$0.022} & \textbf{0.381}{\scriptsize$\pm$0.019} \\
 & chat-only & 0.474{\scriptsize$\pm$0.016} & 0.387{\scriptsize$\pm$0.007} & 0.385{\scriptsize$\pm$0.008} & 0.746{\scriptsize$\pm$0.002} & 0.203{\scriptsize$\pm$0.010} & 0.205{\scriptsize$\pm$0.016} \\
 & video-graph & \textbf{0.520}{\scriptsize$\pm$0.017} & 0.426{\scriptsize$\pm$0.008} & 0.439{\scriptsize$\pm$0.011} & 0.772{\scriptsize$\pm$0.006} & 0.252{\scriptsize$\pm$0.026} & 0.295{\scriptsize$\pm$0.006} \\
 & video-chat & 0.396{\scriptsize$\pm$0.038} & 0.358{\scriptsize$\pm$0.013} & 0.343{\scriptsize$\pm$0.018} & 0.769{\scriptsize$\pm$0.002} & 0.105{\scriptsize$\pm$0.016} & 0.155{\scriptsize$\pm$0.036} \\
 & graph-chat & 0.507{\scriptsize$\pm$0.019} & 0.439{\scriptsize$\pm$0.008} & 0.444{\scriptsize$\pm$0.011} & 0.781{\scriptsize$\pm$0.006} & 0.193{\scriptsize$\pm$0.019} & 0.358{\scriptsize$\pm$0.013} \\
 & video-graph-chat & 0.447{\scriptsize$\pm$0.007} & 0.389{\scriptsize$\pm$0.003} & 0.383{\scriptsize$\pm$0.004} & \textbf{0.782}{\scriptsize$\pm$0.003} & 0.142{\scriptsize$\pm$0.008} & 0.227{\scriptsize$\pm$0.006} \\
\midrule
\multirow{7}{*}{Qwen3-VL-32B-Instruct} & video-only & 0.436{\scriptsize$\pm$0.012} & 0.400{\scriptsize$\pm$0.012} & 0.235{\scriptsize$\pm$0.010} & 0.092{\scriptsize$\pm$0.006} & 0.337{\scriptsize$\pm$0.007} & \textbf{0.274}{\scriptsize$\pm$0.022} \\
 & graph-only & 0.494{\scriptsize$\pm$0.012} & 0.395{\scriptsize$\pm$0.002} & 0.316{\scriptsize$\pm$0.007} & 0.454{\scriptsize$\pm$0.002} & 0.335{\scriptsize$\pm$0.004} & 0.158{\scriptsize$\pm$0.023} \\
 & chat-only & 0.339{\scriptsize$\pm$0.003} & 0.344{\scriptsize$\pm$0.005} & 0.266{\scriptsize$\pm$0.004} & 0.371{\scriptsize$\pm$0.003} & 0.314{\scriptsize$\pm$0.007} & 0.115{\scriptsize$\pm$0.012} \\
 & video-graph & \textbf{0.524}{\scriptsize$\pm$0.038} & 0.428{\scriptsize$\pm$0.015} & 0.370{\scriptsize$\pm$0.014} & 0.523{\scriptsize$\pm$0.012} & 0.358{\scriptsize$\pm$0.010} & 0.229{\scriptsize$\pm$0.039} \\
 & video-chat & 0.358{\scriptsize$\pm$0.002} & 0.375{\scriptsize$\pm$0.005} & 0.315{\scriptsize$\pm$0.002} & 0.480{\scriptsize$\pm$0.014} & 0.336{\scriptsize$\pm$0.012} & 0.129{\scriptsize$\pm$0.021} \\
 & graph-chat & 0.445{\scriptsize$\pm$0.009} & \textbf{0.440}{\scriptsize$\pm$0.004} & 0.395{\scriptsize$\pm$0.006} & 0.616{\scriptsize$\pm$0.008} & \textbf{0.380}{\scriptsize$\pm$0.003} & 0.188{\scriptsize$\pm$0.010} \\
 & video-graph-chat & 0.428{\scriptsize$\pm$0.009} & 0.425{\scriptsize$\pm$0.004} & \textbf{0.396}{\scriptsize$\pm$0.005} & \textbf{0.629}{\scriptsize$\pm$0.005} & 0.356{\scriptsize$\pm$0.005} & 0.204{\scriptsize$\pm$0.013} \\
\bottomrule
\end{tabular}%
}
\caption{Result for Experiment 1b. Mean$\pm$std over runs where a model has several; bold marks the best score within each model.}
\label{1b_results_full_table}
\end{table}

\begin{table}[t]
\centering
\resizebox{\linewidth}{!}{%
\setlength{\tabcolsep}{4pt}
\begin{tabular}{llcccc|cccc}
\toprule
& & \multicolumn{4}{c}{\textbf{Overall}} & \multicolumn{4}{c}{\textbf{MAE by true rating}} \\
\cmidrule(lr){3-6} \cmidrule(lr){7-10}
\textbf{Model} & \textbf{Setting} &
\textbf{MAE}$\downarrow$ & \textbf{Exact acc.} & \textbf{Within-1} & \textbf{Spearman $\rho$} &
\textbf{Low 1--2} & \textbf{Moderate 3} & \textbf{Mod-high 4} & \textbf{High 5} \\
\midrule
\multirow{3}{*}{GPT-6-Astra} & video-only & 0.945{\scriptsize$\pm$0.070} & 0.244{\scriptsize$\pm$0.031} & 0.841{\scriptsize$\pm$0.032} & \textbf{0.379}{\scriptsize$\pm$0.130} & 1.775{\scriptsize$\pm$0.391} & 0.684{\scriptsize$\pm$0.031} & 0.577{\scriptsize$\pm$0.150} & 0.988{\scriptsize$\pm$0.062} \\
 & chat-only & 1.246 & 0.183 & 0.611 & 0.107 & 1.714 & 0.636 & 0.759 & 1.480 \\
 & video-graph-chat & \textbf{0.740} & \textbf{0.365} & \textbf{0.887} & 0.263 & 1.429 & 0.917 & 0.373 & 0.802 \\
\midrule
\multirow{3}{*}{Claude} & video-only & 0.858 & 0.266 & 0.865 & 0.136 & 1.675 & 0.742 & 0.275 & 1.131 \\
 & chat-only & 0.857 & 0.276 & 0.867 & 0.244 & 1.778 & 0.633 & 0.346 & 1.048 \\
 & video-graph-chat & \textbf{0.692} & \textbf{0.407} & \textbf{0.895} & \textbf{0.299} & 1.353 & 0.933 & 0.371 & 0.735 \\
\midrule
\multirow{7}{*}{Gemini-3.1-Pro-Preview} & video-only & 0.685{\scriptsize$\pm$0.006} & 0.527{\scriptsize$\pm$0.003} & 0.850{\scriptsize$\pm$0.003} & 0.078{\scriptsize$\pm$0.057} & 3.175{\scriptsize$\pm$0.050} & 1.968{\scriptsize$\pm$0.032} & 0.961{\scriptsize$\pm$0.009} & 0.048{\scriptsize$\pm$0.006} \\
 & graph-only & 0.632{\scriptsize$\pm$0.012} & 0.509{\scriptsize$\pm$0.003} & \textbf{0.872}{\scriptsize$\pm$0.003} & 0.273{\scriptsize$\pm$0.016} & 1.563{\scriptsize$\pm$0.027} & 1.699{\scriptsize$\pm$0.049} & 0.855{\scriptsize$\pm$0.005} & 0.221{\scriptsize$\pm$0.020} \\
 & chat-only & 0.992{\scriptsize$\pm$0.018} & 0.471{\scriptsize$\pm$0.002} & 0.752{\scriptsize$\pm$0.003} & 0.078{\scriptsize$\pm$0.027} & 2.611{\scriptsize$\pm$0.056} & 1.856{\scriptsize$\pm$0.038} & 1.131{\scriptsize$\pm$0.032} & 0.576{\scriptsize$\pm$0.030} \\
 & video-graph & 0.635{\scriptsize$\pm$0.018} & 0.525{\scriptsize$\pm$0.003} & 0.862{\scriptsize$\pm$0.010} & 0.264{\scriptsize$\pm$0.020} & 1.642{\scriptsize$\pm$0.076} & 1.731{\scriptsize$\pm$0.049} & 0.952{\scriptsize$\pm$0.009} & 0.161{\scriptsize$\pm$0.021} \\
 & video-chat & 0.768{\scriptsize$\pm$0.029} & 0.530{\scriptsize$\pm$0.011} & 0.813{\scriptsize$\pm$0.005} & 0.215{\scriptsize$\pm$0.024} & 2.627{\scriptsize$\pm$0.034} & 1.811{\scriptsize$\pm$0.038} & 1.051{\scriptsize$\pm$0.048} & 0.232{\scriptsize$\pm$0.057} \\
 & graph-chat & 0.631{\scriptsize$\pm$0.002} & \textbf{0.559}{\scriptsize$\pm$0.003} & 0.851{\scriptsize$\pm$0.002} & \textbf{0.295}{\scriptsize$\pm$0.008} & 1.852{\scriptsize$\pm$0.116} & 1.778{\scriptsize$\pm$0.019} & 0.984{\scriptsize$\pm$0.033} & 0.100{\scriptsize$\pm$0.007} \\
 & video-graph-chat & \textbf{0.609}{\scriptsize$\pm$0.019} & 0.550{\scriptsize$\pm$0.015} & 0.864{\scriptsize$\pm$0.005} & 0.261{\scriptsize$\pm$0.048} & 1.784{\scriptsize$\pm$0.238} & 1.822{\scriptsize$\pm$0.019} & 0.962{\scriptsize$\pm$0.015} & 0.079{\scriptsize$\pm$0.020} \\
\midrule
\multirow{7}{*}{Gemini-3.8-Flash} & video-only & 0.800{\scriptsize$\pm$0.029} & 0.503{\scriptsize$\pm$0.007} & 0.819{\scriptsize$\pm$0.014} & 0.011{\scriptsize$\pm$0.003} & 3.108{\scriptsize$\pm$0.289} & 1.914{\scriptsize$\pm$0.019} & 1.015{\scriptsize$\pm$0.045} & 0.245{\scriptsize$\pm$0.026} \\
 & graph-only & 0.607{\scriptsize$\pm$0.003} & 0.521{\scriptsize$\pm$0.004} & \textbf{0.882}{\scriptsize$\pm$0.002} & 0.276{\scriptsize$\pm$0.010} & 1.738{\scriptsize$\pm$0.048} & 1.688{\scriptsize$\pm$0.019} & 0.870{\scriptsize$\pm$0.009} & 0.149{\scriptsize$\pm$0.008} \\
 & chat-only & 0.892{\scriptsize$\pm$0.018} & 0.480{\scriptsize$\pm$0.004} & 0.777{\scriptsize$\pm$0.004} & 0.085{\scriptsize$\pm$0.011} & 2.537{\scriptsize$\pm$0.032} & 1.644{\scriptsize$\pm$0.038} & 1.076{\scriptsize$\pm$0.019} & 0.471{\scriptsize$\pm$0.030} \\
 & video-graph & 0.611{\scriptsize$\pm$0.007} & 0.525{\scriptsize$\pm$0.007} & 0.876{\scriptsize$\pm$0.003} & 0.255{\scriptsize$\pm$0.015} & 1.825{\scriptsize$\pm$0.087} & 1.849{\scriptsize$\pm$0.049} & 0.925{\scriptsize$\pm$0.018} & 0.094{\scriptsize$\pm$0.008} \\
 & video-chat & 0.731{\scriptsize$\pm$0.045} & 0.543{\scriptsize$\pm$0.024} & 0.826{\scriptsize$\pm$0.007} & 0.159{\scriptsize$\pm$0.040} & 2.843{\scriptsize$\pm$0.180} & 1.811{\scriptsize$\pm$0.019} & 0.975{\scriptsize$\pm$0.032} & 0.180{\scriptsize$\pm$0.053} \\
 & graph-chat & 0.607{\scriptsize$\pm$0.007} & 0.537{\scriptsize$\pm$0.002} & 0.878{\scriptsize$\pm$0.003} & 0.276{\scriptsize$\pm$0.008} & 1.926{\scriptsize$\pm$0.032} & 1.700{\scriptsize$\pm$0.033} & 0.924{\scriptsize$\pm$0.032} & 0.094{\scriptsize$\pm$0.004} \\
 & video-graph-chat & \textbf{0.589}{\scriptsize$\pm$0.011} & \textbf{0.559}{\scriptsize$\pm$0.008} & 0.868{\scriptsize$\pm$0.002} & \textbf{0.299}{\scriptsize$\pm$0.027} & 1.980{\scriptsize$\pm$0.068} & 1.789{\scriptsize$\pm$0.051} & 0.924{\scriptsize$\pm$0.019} & 0.048{\scriptsize$\pm$0.007} \\
\midrule
\multirow{7}{*}{Qwen3-VL-30B-A3B-Instruct} & video-only & 0.898{\scriptsize$\pm$0.066} & 0.330{\scriptsize$\pm$0.020} & 0.814{\scriptsize$\pm$0.033} & 0.102{\scriptsize$\pm$0.075} & 2.225{\scriptsize$\pm$0.100} & 1.043{\scriptsize$\pm$0.093} & 0.635{\scriptsize$\pm$0.086} & 0.887{\scriptsize$\pm$0.083} \\
 & graph-only & 0.770{\scriptsize$\pm$0.005} & 0.295{\scriptsize$\pm$0.006} & \textbf{0.930}{\scriptsize$\pm$0.002} & 0.194{\scriptsize$\pm$0.018} & 1.770{\scriptsize$\pm$0.055} & 0.989{\scriptsize$\pm$0.037} & 0.278{\scriptsize$\pm$0.019} & 0.910{\scriptsize$\pm$0.003} \\
 & chat-only & 0.827{\scriptsize$\pm$0.031} & 0.438{\scriptsize$\pm$0.030} & 0.844{\scriptsize$\pm$0.004} & 0.145{\scriptsize$\pm$0.047} & 2.500{\scriptsize$\pm$0.056} & 1.467{\scriptsize$\pm$0.033} & 0.764{\scriptsize$\pm$0.074} & 0.549{\scriptsize$\pm$0.032} \\
 & video-graph & \textbf{0.622}{\scriptsize$\pm$0.017} & 0.482{\scriptsize$\pm$0.010} & 0.900{\scriptsize$\pm$0.006} & 0.275{\scriptsize$\pm$0.046} & 1.992{\scriptsize$\pm$0.029} & 1.387{\scriptsize$\pm$0.112} & 0.609{\scriptsize$\pm$0.019} & 0.358{\scriptsize$\pm$0.009} \\
 & video-chat & 0.721{\scriptsize$\pm$0.019} & 0.519{\scriptsize$\pm$0.014} & 0.835{\scriptsize$\pm$0.007} & 0.170{\scriptsize$\pm$0.029} & 2.765{\scriptsize$\pm$0.156} & 1.778{\scriptsize$\pm$0.051} & 0.908{\scriptsize$\pm$0.046} & 0.211{\scriptsize$\pm$0.031} \\
 & graph-chat & 0.684{\scriptsize$\pm$0.011} & 0.431{\scriptsize$\pm$0.009} & 0.902{\scriptsize$\pm$0.007} & 0.277{\scriptsize$\pm$0.025} & 1.907{\scriptsize$\pm$0.032} & 1.167{\scriptsize$\pm$0.033} & 0.562{\scriptsize$\pm$0.015} & 0.516{\scriptsize$\pm$0.008} \\
 & video-graph-chat & 0.634{\scriptsize$\pm$0.016} & \textbf{0.520}{\scriptsize$\pm$0.011} & 0.875{\scriptsize$\pm$0.007} & \textbf{0.285}{\scriptsize$\pm$0.033} & 2.059{\scriptsize$\pm$0.059} & 1.544{\scriptsize$\pm$0.084} & 0.811{\scriptsize$\pm$0.053} & 0.222{\scriptsize$\pm$0.000} \\
\midrule
\multirow{7}{*}{Qwen3-VL-32B-Instruct} & video-only & 1.194{\scriptsize$\pm$0.018} & 0.193{\scriptsize$\pm$0.016} & 0.634{\scriptsize$\pm$0.004} & 0.130{\scriptsize$\pm$0.028} & 1.425{\scriptsize$\pm$0.087} & 0.333{\scriptsize$\pm$0.145} & 0.739{\scriptsize$\pm$0.033} & 1.578{\scriptsize$\pm$0.005} \\
 & graph-only & 0.692{\scriptsize$\pm$0.013} & 0.415{\scriptsize$\pm$0.012} & 0.891{\scriptsize$\pm$0.003} & 0.236{\scriptsize$\pm$0.018} & 1.754{\scriptsize$\pm$0.027} & 1.043{\scriptsize$\pm$0.019} & 0.479{\scriptsize$\pm$0.024} & 0.640{\scriptsize$\pm$0.009} \\
 & chat-only & 0.803{\scriptsize$\pm$0.005} & 0.476{\scriptsize$\pm$0.003} & 0.830{\scriptsize$\pm$0.005} & 0.187{\scriptsize$\pm$0.014} & 2.519{\scriptsize$\pm$0.032} & 1.544{\scriptsize$\pm$0.038} & 0.899{\scriptsize$\pm$0.007} & 0.421{\scriptsize$\pm$0.004} \\
 & video-graph & 0.657{\scriptsize$\pm$0.012} & 0.426{\scriptsize$\pm$0.004} & \textbf{0.914}{\scriptsize$\pm$0.008} & 0.243{\scriptsize$\pm$0.013} & 1.825{\scriptsize$\pm$0.050} & 1.226{\scriptsize$\pm$0.056} & 0.493{\scriptsize$\pm$0.014} & 0.535{\scriptsize$\pm$0.025} \\
 & video-chat & 0.719{\scriptsize$\pm$0.004} & 0.496{\scriptsize$\pm$0.007} & 0.840{\scriptsize$\pm$0.004} & 0.172{\scriptsize$\pm$0.007} & 2.588{\scriptsize$\pm$0.000} & 1.567{\scriptsize$\pm$0.088} & 0.861{\scriptsize$\pm$0.037} & 0.289{\scriptsize$\pm$0.016} \\
 & graph-chat & 0.615{\scriptsize$\pm$0.006} & 0.524{\scriptsize$\pm$0.006} & 0.878{\scriptsize$\pm$0.003} & \textbf{0.297}{\scriptsize$\pm$0.012} & 2.259{\scriptsize$\pm$0.085} & 1.489{\scriptsize$\pm$0.038} & 0.777{\scriptsize$\pm$0.007} & 0.184{\scriptsize$\pm$0.013} \\
 & video-graph-chat & \textbf{0.604}{\scriptsize$\pm$0.012} & \textbf{0.532}{\scriptsize$\pm$0.008} & 0.876{\scriptsize$\pm$0.002} & 0.282{\scriptsize$\pm$0.028} & 2.196{\scriptsize$\pm$0.034} & 1.589{\scriptsize$\pm$0.019} & 0.781{\scriptsize$\pm$0.013} & 0.159{\scriptsize$\pm$0.011} \\
\bottomrule
\end{tabular}%
}
\caption{Result for Experiment 2a. Mean$\pm$std over runs where a model has several; bold marks the best score within each model. Band columns give the MAE within each true-rating band.}
\label{2a_results_full_table}
\end{table}

\begin{table}[t]
\centering
\resizebox{\linewidth}{!}{%
\setlength{\tabcolsep}{4pt}
\begin{tabular}{llcccc|cccc}
\toprule
& & \multicolumn{4}{c}{\textbf{Overall}} & \multicolumn{4}{c}{\textbf{MAE by true rating}} \\
\cmidrule(lr){3-6} \cmidrule(lr){7-10}
\textbf{Model} & \textbf{Setting} &
\textbf{MAE}$\downarrow$ & \textbf{Exact acc.} & \textbf{Within-1} & \textbf{Spearman $\rho$} &
\textbf{Low 1--2} & \textbf{Moderate 3} & \textbf{Mod-high 4} & \textbf{High 5} \\
\midrule
\multirow{3}{*}{GPT-6-Astra} & video-only & 0.912{\scriptsize$\pm$0.084} & 0.332{\scriptsize$\pm$0.067} & 0.805{\scriptsize$\pm$0.030} & \textbf{0.352}{\scriptsize$\pm$0.101} & 1.853{\scriptsize$\pm$0.255} & 0.711{\scriptsize$\pm$0.077} & 0.710{\scriptsize$\pm$0.157} & 0.855{\scriptsize$\pm$0.105} \\
 & chat-only & 1.134 & 0.221 & 0.713 & 0.163 & 1.500 & 0.727 & 0.788 & 1.320 \\
 & video-graph-chat & \textbf{0.765} & \textbf{0.438} & \textbf{0.854} & 0.311 & 1.500 & 0.750 & 0.500 & 0.741 \\
\midrule
\multirow{3}{*}{Claude} & video-only & 0.817 & 0.344 & 0.849 & 0.208 & 1.568 & 0.625 & 0.405 & 1.018 \\
 & chat-only & 0.973 & 0.267 & 0.809 & 0.137 & 1.789 & 1.059 & 0.610 & 1.052 \\
 & video-graph-chat & \textbf{0.741} & \textbf{0.362} & \textbf{0.917} & \textbf{0.322} & 1.556 & 0.765 & 0.280 & 0.897 \\
\midrule
\multirow{7}{*}{Gemini-3.1-Pro-Preview} & video-only & 0.764{\scriptsize$\pm$0.016} & \textbf{0.508}{\scriptsize$\pm$0.002} & 0.821{\scriptsize$\pm$0.006} & 0.046{\scriptsize$\pm$0.039} & 3.225{\scriptsize$\pm$0.050} & 2.000{\scriptsize$\pm$0.000} & 1.023{\scriptsize$\pm$0.018} & 0.062{\scriptsize$\pm$0.017} \\
 & graph-only & 0.750{\scriptsize$\pm$0.022} & 0.471{\scriptsize$\pm$0.008} & 0.836{\scriptsize$\pm$0.005} & \textbf{0.299}{\scriptsize$\pm$0.021} & 1.428{\scriptsize$\pm$0.050} & 1.458{\scriptsize$\pm$0.014} & 0.891{\scriptsize$\pm$0.030} & 0.440{\scriptsize$\pm$0.030} \\
 & chat-only & 1.069{\scriptsize$\pm$0.029} & 0.436{\scriptsize$\pm$0.008} & 0.729{\scriptsize$\pm$0.013} & 0.093{\scriptsize$\pm$0.021} & 2.561{\scriptsize$\pm$0.110} & 1.676{\scriptsize$\pm$0.051} & 1.248{\scriptsize$\pm$0.063} & 0.621{\scriptsize$\pm$0.066} \\
 & video-graph & 0.763{\scriptsize$\pm$0.005} & 0.478{\scriptsize$\pm$0.004} & 0.825{\scriptsize$\pm$0.004} & 0.280{\scriptsize$\pm$0.013} & 1.514{\scriptsize$\pm$0.109} & 1.558{\scriptsize$\pm$0.029} & 0.974{\scriptsize$\pm$0.017} & 0.384{\scriptsize$\pm$0.033} \\
 & video-chat & 0.895{\scriptsize$\pm$0.032} & 0.485{\scriptsize$\pm$0.004} & 0.776{\scriptsize$\pm$0.013} & 0.139{\scriptsize$\pm$0.038} & 3.123{\scriptsize$\pm$0.030} & 1.765{\scriptsize$\pm$0.051} & 1.142{\scriptsize$\pm$0.025} & 0.246{\scriptsize$\pm$0.056} \\
 & graph-chat & \textbf{0.734}{\scriptsize$\pm$0.031} & 0.485{\scriptsize$\pm$0.008} & \textbf{0.845}{\scriptsize$\pm$0.014} & 0.287{\scriptsize$\pm$0.037} & 1.789{\scriptsize$\pm$0.053} & 1.598{\scriptsize$\pm$0.090} & 0.951{\scriptsize$\pm$0.024} & 0.261{\scriptsize$\pm$0.035} \\
 & video-graph-chat & 0.759{\scriptsize$\pm$0.010} & 0.479{\scriptsize$\pm$0.009} & 0.834{\scriptsize$\pm$0.004} & 0.253{\scriptsize$\pm$0.010} & 1.842{\scriptsize$\pm$0.053} & 1.706{\scriptsize$\pm$0.029} & 0.976{\scriptsize$\pm$0.032} & 0.262{\scriptsize$\pm$0.018} \\
\midrule
\multirow{7}{*}{Gemini-3.8-Flash} & video-only & 0.792{\scriptsize$\pm$0.006} & 0.503{\scriptsize$\pm$0.004} & 0.811{\scriptsize$\pm$0.002} & 0.042{\scriptsize$\pm$0.012} & 3.167{\scriptsize$\pm$0.050} & 1.975{\scriptsize$\pm$0.025} & 1.049{\scriptsize$\pm$0.030} & 0.114{\scriptsize$\pm$0.036} \\
 & graph-only & 0.652{\scriptsize$\pm$0.016} & 0.513{\scriptsize$\pm$0.009} & \textbf{0.870}{\scriptsize$\pm$0.007} & 0.286{\scriptsize$\pm$0.028} & 1.659{\scriptsize$\pm$0.066} & 1.517{\scriptsize$\pm$0.038} & 0.810{\scriptsize$\pm$0.023} & 0.260{\scriptsize$\pm$0.005} \\
 & chat-only & 1.113{\scriptsize$\pm$0.004} & 0.431{\scriptsize$\pm$0.004} & 0.697{\scriptsize$\pm$0.004} & 0.094{\scriptsize$\pm$0.008} & 2.596{\scriptsize$\pm$0.132} & 1.804{\scriptsize$\pm$0.045} & 1.313{\scriptsize$\pm$0.019} & 0.633{\scriptsize$\pm$0.025} \\
 & video-graph & \textbf{0.643}{\scriptsize$\pm$0.003} & 0.524{\scriptsize$\pm$0.004} & 0.869{\scriptsize$\pm$0.003} & 0.303{\scriptsize$\pm$0.011} & 1.877{\scriptsize$\pm$0.025} & 1.667{\scriptsize$\pm$0.038} & 0.891{\scriptsize$\pm$0.013} & 0.138{\scriptsize$\pm$0.009} \\
 & video-chat & 0.777{\scriptsize$\pm$0.020} & 0.509{\scriptsize$\pm$0.006} & 0.811{\scriptsize$\pm$0.009} & 0.187{\scriptsize$\pm$0.014} & 2.842{\scriptsize$\pm$0.091} & 1.853{\scriptsize$\pm$0.029} & 1.012{\scriptsize$\pm$0.053} & 0.108{\scriptsize$\pm$0.008} \\
 & graph-chat & 0.669{\scriptsize$\pm$0.015} & 0.513{\scriptsize$\pm$0.013} & 0.864{\scriptsize$\pm$0.002} & 0.306{\scriptsize$\pm$0.017} & 1.807{\scriptsize$\pm$0.030} & 1.618{\scriptsize$\pm$0.051} & 0.919{\scriptsize$\pm$0.025} & 0.146{\scriptsize$\pm$0.020} \\
 & video-graph-chat & 0.658{\scriptsize$\pm$0.011} & \textbf{0.525}{\scriptsize$\pm$0.002} & 0.862{\scriptsize$\pm$0.006} & \textbf{0.320}{\scriptsize$\pm$0.022} & 1.772{\scriptsize$\pm$0.132} & 1.647{\scriptsize$\pm$0.029} & 0.907{\scriptsize$\pm$0.025} & 0.129{\scriptsize$\pm$0.022} \\
\midrule
\multirow{7}{*}{Qwen3-VL-30B-A3B-Instruct} & video-only & 1.281{\scriptsize$\pm$0.011} & 0.278{\scriptsize$\pm$0.019} & 0.641{\scriptsize$\pm$0.010} & 0.048{\scriptsize$\pm$0.034} & 1.978{\scriptsize$\pm$0.190} & 1.125{\scriptsize$\pm$0.025} & 0.971{\scriptsize$\pm$0.033} & 1.415{\scriptsize$\pm$0.019} \\
 & graph-only & 0.767{\scriptsize$\pm$0.006} & 0.384{\scriptsize$\pm$0.007} & \textbf{0.876}{\scriptsize$\pm$0.010} & 0.206{\scriptsize$\pm$0.028} & 1.558{\scriptsize$\pm$0.066} & 0.975{\scriptsize$\pm$0.050} & 0.437{\scriptsize$\pm$0.013} & 0.829{\scriptsize$\pm$0.008} \\
 & chat-only & 0.926{\scriptsize$\pm$0.019} & 0.439{\scriptsize$\pm$0.017} & 0.783{\scriptsize$\pm$0.013} & 0.133{\scriptsize$\pm$0.008} & 2.702{\scriptsize$\pm$0.061} & 1.529{\scriptsize$\pm$0.078} & 0.923{\scriptsize$\pm$0.060} & 0.546{\scriptsize$\pm$0.026} \\
 & video-graph & 0.731{\scriptsize$\pm$0.008} & 0.445{\scriptsize$\pm$0.014} & 0.868{\scriptsize$\pm$0.004} & 0.225{\scriptsize$\pm$0.032} & 1.529{\scriptsize$\pm$0.109} & 1.300{\scriptsize$\pm$0.066} & 0.652{\scriptsize$\pm$0.057} & 0.566{\scriptsize$\pm$0.021} \\
 & video-chat & 0.884{\scriptsize$\pm$0.040} & 0.473{\scriptsize$\pm$0.019} & 0.781{\scriptsize$\pm$0.010} & 0.113{\scriptsize$\pm$0.066} & 2.877{\scriptsize$\pm$0.152} & 1.706{\scriptsize$\pm$0.078} & 1.008{\scriptsize$\pm$0.019} & 0.348{\scriptsize$\pm$0.072} \\
 & graph-chat & 0.711{\scriptsize$\pm$0.004} & 0.469{\scriptsize$\pm$0.006} & 0.866{\scriptsize$\pm$0.004} & 0.268{\scriptsize$\pm$0.011} & 1.789{\scriptsize$\pm$0.139} & 1.363{\scriptsize$\pm$0.045} & 0.772{\scriptsize$\pm$0.014} & 0.376{\scriptsize$\pm$0.007} \\
 & video-graph-chat & \textbf{0.699}{\scriptsize$\pm$0.023} & \textbf{0.499}{\scriptsize$\pm$0.008} & 0.858{\scriptsize$\pm$0.018} & \textbf{0.323}{\scriptsize$\pm$0.027} & 1.667{\scriptsize$\pm$0.132} & 1.588{\scriptsize$\pm$0.135} & 0.915{\scriptsize$\pm$0.032} & 0.232{\scriptsize$\pm$0.021} \\
\midrule
\multirow{7}{*}{Qwen3-VL-32B-Instruct} & video-only & 1.223{\scriptsize$\pm$0.013} & 0.202{\scriptsize$\pm$0.007} & 0.633{\scriptsize$\pm$0.019} & 0.153{\scriptsize$\pm$0.005} & 1.442{\scriptsize$\pm$0.066} & 0.408{\scriptsize$\pm$0.123} & 0.828{\scriptsize$\pm$0.046} & 1.601{\scriptsize$\pm$0.026} \\
 & graph-only & 0.775{\scriptsize$\pm$0.015} & 0.385{\scriptsize$\pm$0.012} & 0.849{\scriptsize$\pm$0.003} & 0.254{\scriptsize$\pm$0.009} & 1.558{\scriptsize$\pm$0.025} & 0.858{\scriptsize$\pm$0.014} & 0.509{\scriptsize$\pm$0.015} & 0.825{\scriptsize$\pm$0.016} \\
 & chat-only & 0.925{\scriptsize$\pm$0.016} & 0.413{\scriptsize$\pm$0.015} & 0.794{\scriptsize$\pm$0.000} & 0.165{\scriptsize$\pm$0.025} & 2.456{\scriptsize$\pm$0.030} & 1.373{\scriptsize$\pm$0.017} & 0.935{\scriptsize$\pm$0.031} & 0.607{\scriptsize$\pm$0.011} \\
 & video-graph & 0.732{\scriptsize$\pm$0.003} & 0.425{\scriptsize$\pm$0.004} & 0.858{\scriptsize$\pm$0.007} & 0.252{\scriptsize$\pm$0.011} & 1.674{\scriptsize$\pm$0.087} & 1.008{\scriptsize$\pm$0.052} & 0.520{\scriptsize$\pm$0.025} & 0.689{\scriptsize$\pm$0.040} \\
 & video-chat & 0.804{\scriptsize$\pm$0.028} & 0.436{\scriptsize$\pm$0.023} & 0.817{\scriptsize$\pm$0.008} & 0.199{\scriptsize$\pm$0.044} & 2.404{\scriptsize$\pm$0.030} & 1.284{\scriptsize$\pm$0.045} & 0.813{\scriptsize$\pm$0.007} & 0.470{\scriptsize$\pm$0.048} \\
 & graph-chat & 0.686{\scriptsize$\pm$0.013} & 0.452{\scriptsize$\pm$0.006} & \textbf{0.898}{\scriptsize$\pm$0.009} & 0.317{\scriptsize$\pm$0.029} & 1.982{\scriptsize$\pm$0.030} & 1.118{\scriptsize$\pm$0.088} & 0.703{\scriptsize$\pm$0.014} & 0.399{\scriptsize$\pm$0.008} \\
 & video-graph-chat & \textbf{0.661}{\scriptsize$\pm$0.010} & \textbf{0.478}{\scriptsize$\pm$0.008} & 0.893{\scriptsize$\pm$0.008} & \textbf{0.319}{\scriptsize$\pm$0.022} & 2.000{\scriptsize$\pm$0.000} & 1.216{\scriptsize$\pm$0.045} & 0.703{\scriptsize$\pm$0.007} & 0.324{\scriptsize$\pm$0.014} \\
\bottomrule
\end{tabular}%
}
\caption{Result for Experiment 2b. Mean$\pm$std over runs where a model has several; bold marks the best score within each model. Band columns give the MAE within each true-rating band.}
\label{2b_results_full_table}
\end{table}

\subsection{Output Quality \texorpdfstring{$\times$}{x} Collaboration Rating}
\begin{figure}
    \centering
    \includegraphics[width=0.6\linewidth]{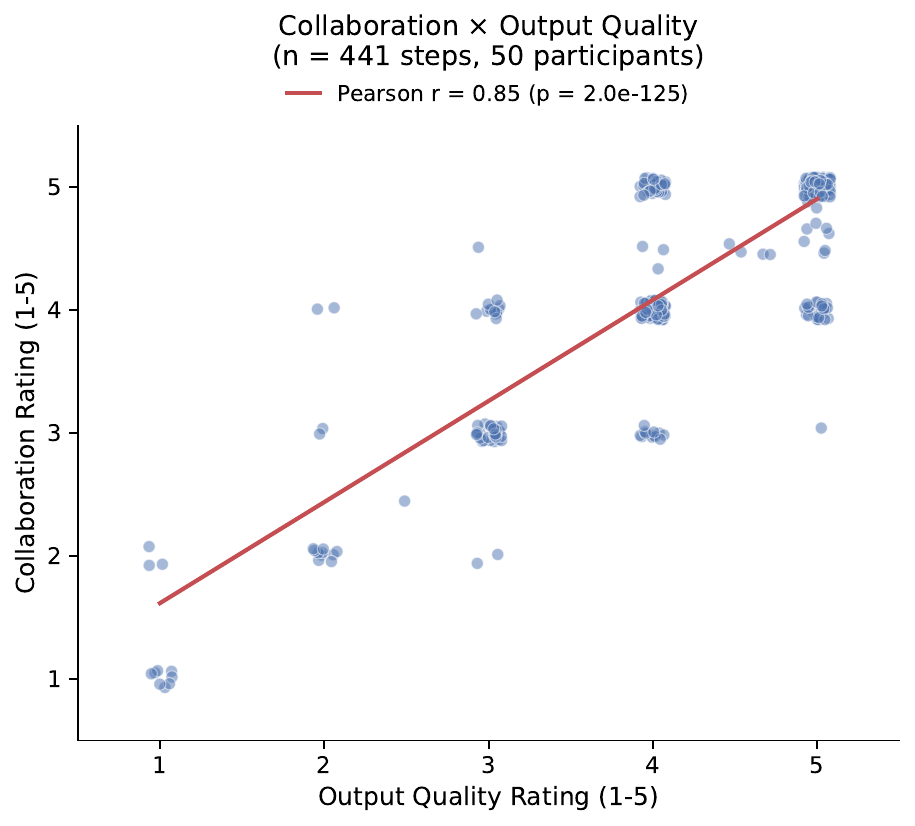}
\caption{Professionals' ratings of collaboration satisfaction and output quality across all steps.}
\label{fig:collab_x_output_quality}
\end{figure}

As shown in Figure~\ref{fig:collab_x_output_quality}, output quality and the collaboration experience perceived by professionals are correlated.

\clearpage
\subsection{Model Comparison on Common Instance Sets}
\label{app:common_instance_sets}

Due to resource limitations, GPT-6-Astra was evaluated on fewer instances than the other models. For a fair comparison, we also evaluate all models on the same instances used for GPT-6-Astra. Scores therefore differ from those in the full tables (Tables~\ref{1a_results_full_table} and~\ref{1b_results_full_table}).

\begin{table}[!htbp]
\centering
\begin{tabular}{lrr}
\toprule
\textbf{Model} & \textbf{F1} & \textbf{Runs} \\
\midrule
GPT-6-Astra & \textbf{0.713} & 1 \\
Claude & 0.699 & 1 \\
Gemini-3.1-Pro-Preview & 0.667 & 3 \\
Gemini-3.8-Flash & 0.620 & 3 \\
Qwen3-VL-30B-A3B-Instruct & 0.560 & 3 \\
Qwen3-VL-32B-Instruct & 0.374 & 3 \\
\bottomrule
\end{tabular}
\caption{F1 for Experiment 1a on the common set of 279 instances from 32 participants. The common set is limited by GPT-6-Astra, which has 279 valid outputs.}
\label{tab:exp1a_common_set}
\end{table}

\begin{table}[!htbp]
\centering
\begin{tabular}{lrr}
\toprule
\textbf{Model} & \textbf{F1} & \textbf{Runs} \\
\midrule
Claude & \textbf{0.533} & 1 \\
GPT-6-Astra & 0.511 & 1 \\
Gemini-3.1-Pro-Preview & 0.399 & 3 \\
Gemini-3.8-Flash & 0.392 & 3 \\
Qwen3-VL-30B-A3B-Instruct & 0.375 & 3 \\
Qwen3-VL-32B-Instruct & 0.221 & 3 \\
\bottomrule
\end{tabular}
\caption{F1 for Experiment 1b on the common set of 349 instances from 37 participants. The common set is limited by GPT-6-Astra, which has 349 valid outputs. GPT-6-Astra's score is based on its single run covering all 349 instances.}
\label{tab:exp1b_common_set}
\end{table}

\clearpage
\subsection{Post-interaction Questionnaire}
\label{sec:post-interaction}
\subsubsection*{Overview: All Behavioral Likert Items (Q13--Q38)}
\begin{figure}[H]
\centering
\includegraphics[width=0.85\linewidth]{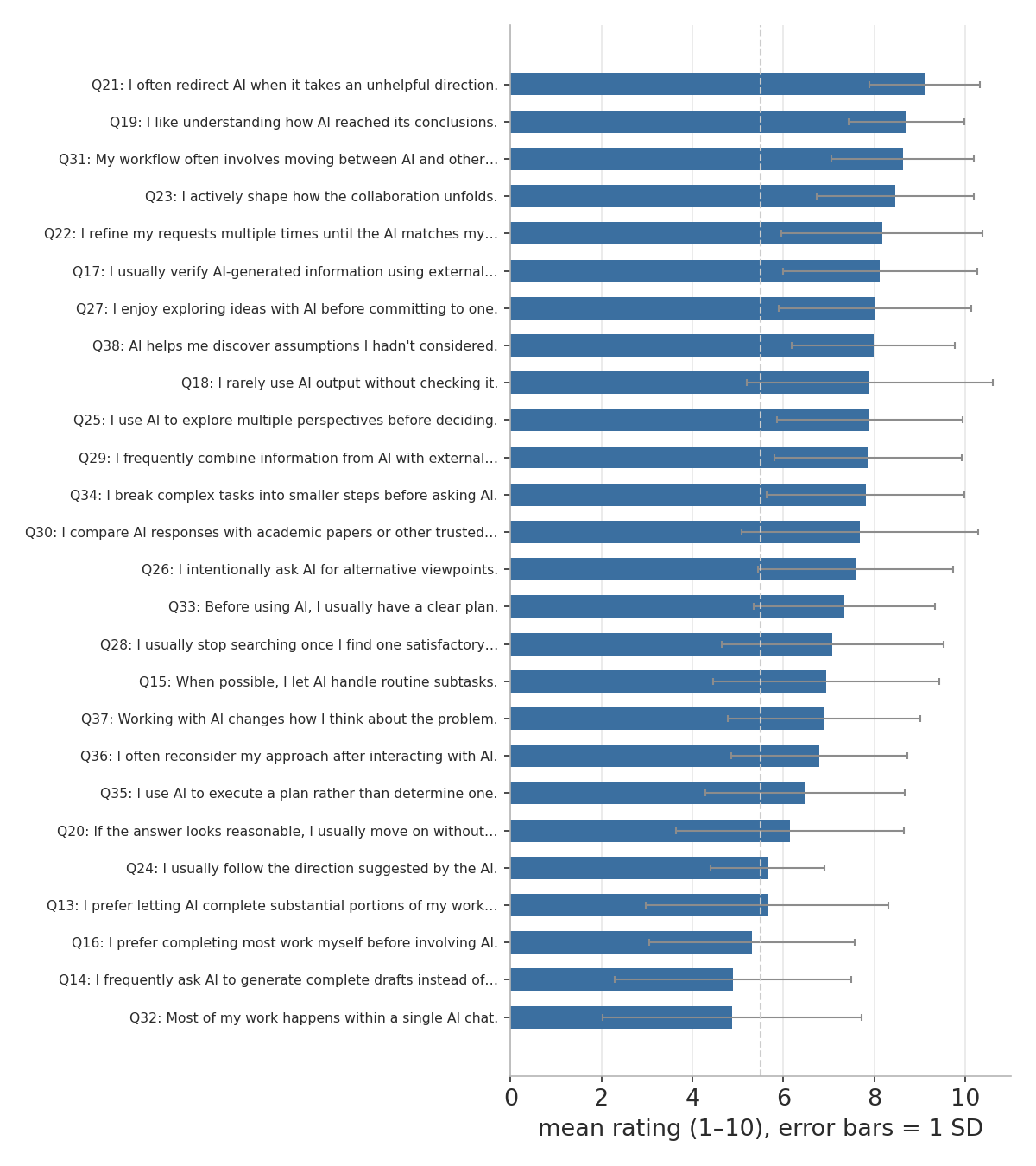}
\caption{Mean rating (1--10) for each of the 26 AI-collaboration behavior
items, sorted low to high (error bars: 1 SD). The dashed line marks the
scale midpoint.}
\end{figure}
\clearpage

\subsubsection*{Session 1 \& Session 2 Reflection}
\begin{figure}[H]
\centering
\begin{subfigure}[t]{0.48\textwidth}
\centering
\includegraphics[width=\linewidth]{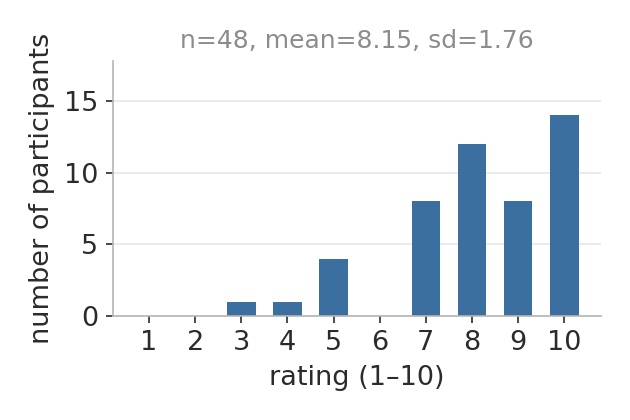}
\caption{\footnotesize Q1. After watching the session 1 recording again and looking at the extracted workflow, how natural do you think your workflow there compared to your general workflow?}
\end{subfigure}
\end{figure}

\paragraph{Q2. If you answered not natural at all in the previous question, why do you think so?}
\textit{Open-ended response}

\begin{figure}[H]
\centering
\begin{subfigure}[t]{0.48\textwidth}
\centering
\includegraphics[width=\linewidth]{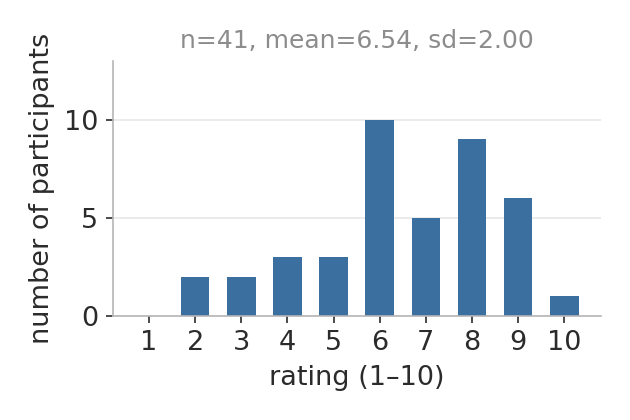}
\caption{\footnotesize Q3. How difficult was the problem you worked on in Session 1?}
\end{subfigure}
\hfill
\begin{subfigure}[t]{0.48\textwidth}
\centering
\includegraphics[width=\linewidth]{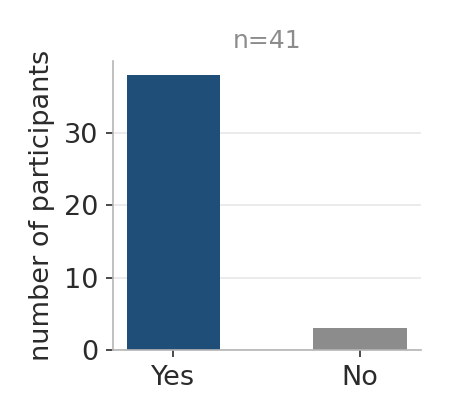}
\caption{\footnotesize Q4. Have you ever done problem 1 manually without agentic AI's help? }
\end{subfigure}
\end{figure}

\begin{figure}[H]
\centering
\begin{subfigure}[t]{0.48\textwidth}
\centering
\includegraphics[width=\linewidth]{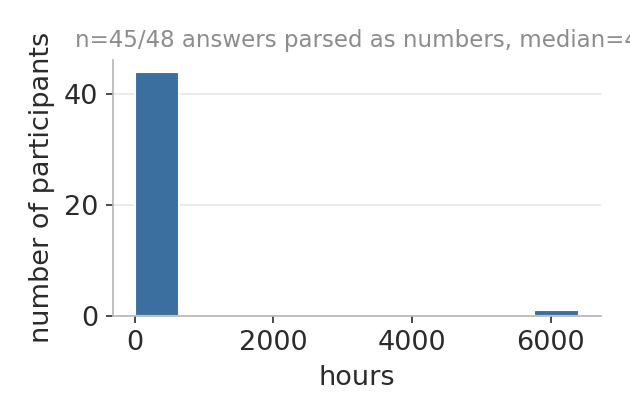}
\caption{\footnotesize Q5. How many hours on average did it take or do you think it takes for you to finish the problem for session 1 without agentic AI's help?}
\end{subfigure}
\hfill
\begin{subfigure}[t]{0.48\textwidth}
\centering
\includegraphics[width=\linewidth]{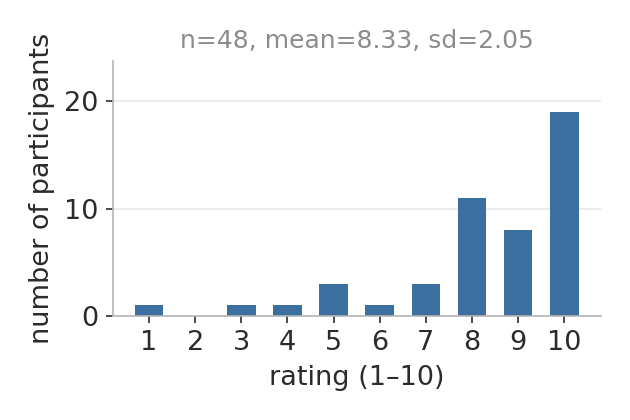}
\caption{\footnotesize Q6. After watching the session 2 recording again and looking at the extracted workflow, how natural do you think your workflow there compared to your general workflow?}
\end{subfigure}
\end{figure}

\paragraph{Q7. If you answered not natural at all in the previous question, why do you think so?}
\textit{Open-ended response}

\begin{figure}[H]
\centering
\begin{subfigure}[t]{0.48\textwidth}
\centering
\includegraphics[width=\linewidth]{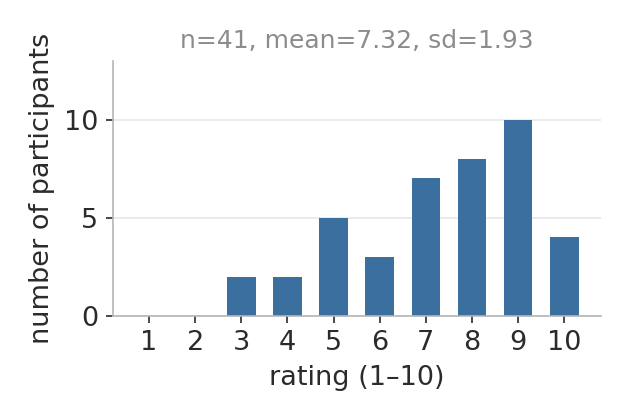}
\caption{\footnotesize Q8. How difficult was the problem you worked on in Session 2?}
\end{subfigure}
\hfill
\begin{subfigure}[t]{0.48\textwidth}
\centering
\includegraphics[width=\linewidth]{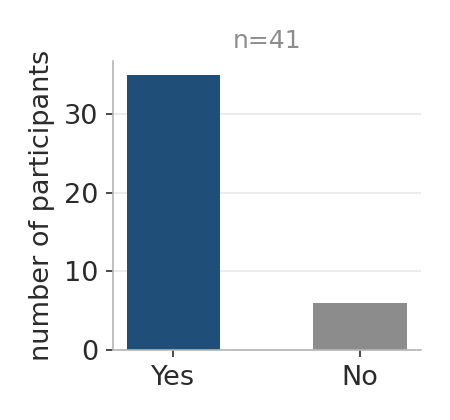}
\caption{\footnotesize Q9. Have you ever done problem 2 manually without agentic AI's help?}
\end{subfigure}
\end{figure}

\begin{figure}[H]
\centering
\begin{subfigure}[t]{0.48\textwidth}
\centering
\includegraphics[width=\linewidth]{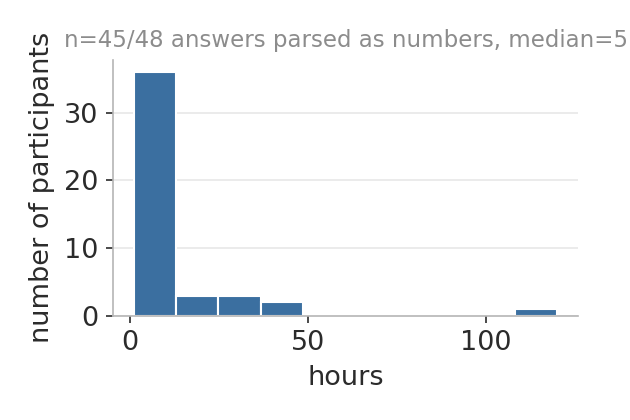}
\caption{\footnotesize Q10. How many hours on average did it take or do you think it takes for you to finish the problem for session 2 without agentic AI's help?}
\end{subfigure}
\end{figure}

\paragraph{Q11. For either or both session 1 and 2, were there any points where you changed your plan or workflow? What prompted that?}
\textit{Open-ended response}

\paragraph{Q12. If you did a similar task again, would you approach it differently? If so, what would you change?}
\textit{Open-ended response.}

\clearpage
\subsubsection*{AI Reliance / Delegation}
\begin{figure}[H]
\centering
\begin{subfigure}[t]{0.48\textwidth}
\centering
\includegraphics[width=\linewidth]{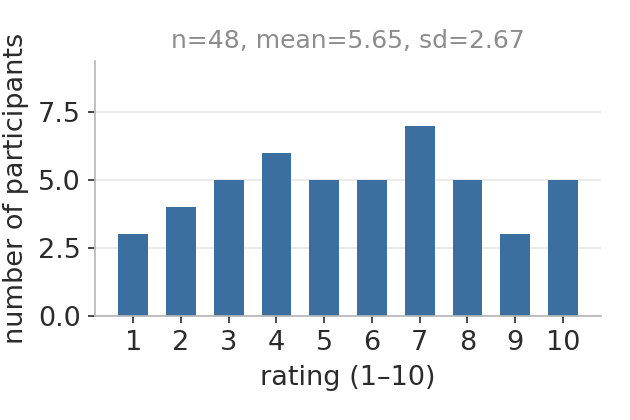}
\caption{\footnotesize Q13. I prefer letting AI complete substantial portions of my work before reviewing it.}
\end{subfigure}
\hfill
\begin{subfigure}[t]{0.48\textwidth}
\centering
\includegraphics[width=\linewidth]{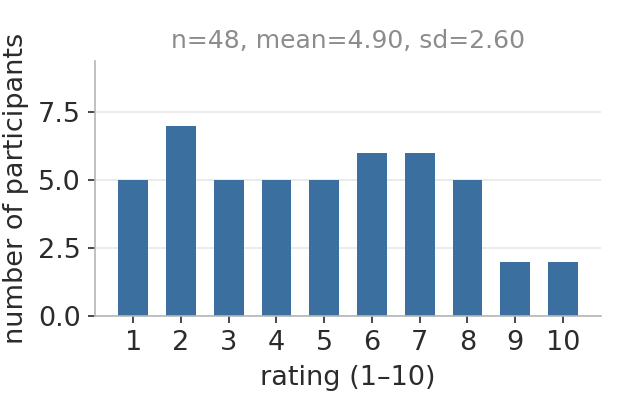}
\caption{\footnotesize Q14. I frequently ask AI to generate complete drafts instead of working incrementally.}
\end{subfigure}
\end{figure}

\begin{figure}[H]
\centering
\begin{subfigure}[t]{0.48\textwidth}
\centering
\includegraphics[width=\linewidth]{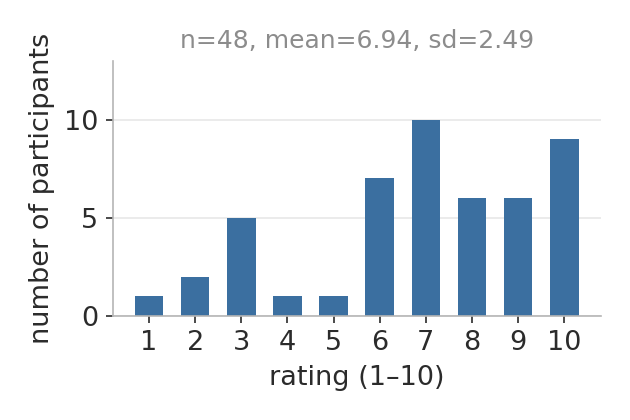}
\caption{\footnotesize Q15. When possible, I let AI handle routine subtasks.}
\end{subfigure}
\hfill
\begin{subfigure}[t]{0.48\textwidth}
\centering
\includegraphics[width=\linewidth]{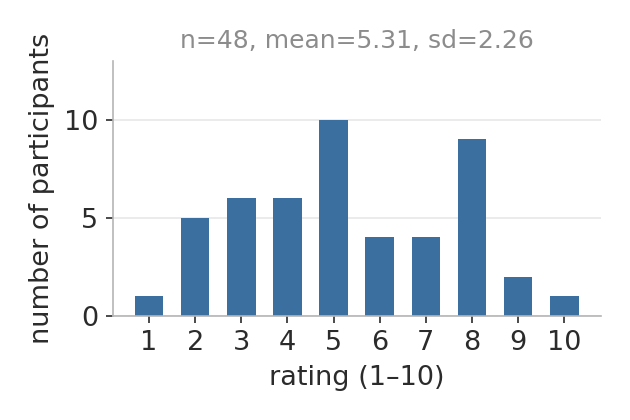}
\caption{\footnotesize Q16. I prefer completing most work myself before involving AI.}
\end{subfigure}
\end{figure}
\clearpage
\subsubsection*{Verification Behavior}
\begin{figure}[H]
\centering
\begin{subfigure}[t]{0.48\textwidth}
\centering
\includegraphics[width=\linewidth]{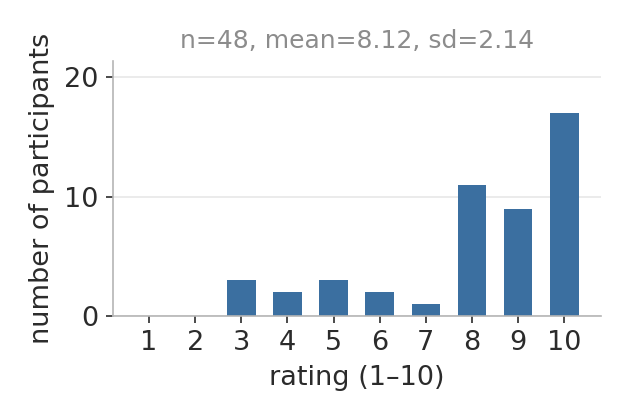}
\caption{\footnotesize Q17. I usually verify AI-generated information using external sources.}
\end{subfigure}
\hfill
\begin{subfigure}[t]{0.48\textwidth}
\centering
\includegraphics[width=\linewidth]{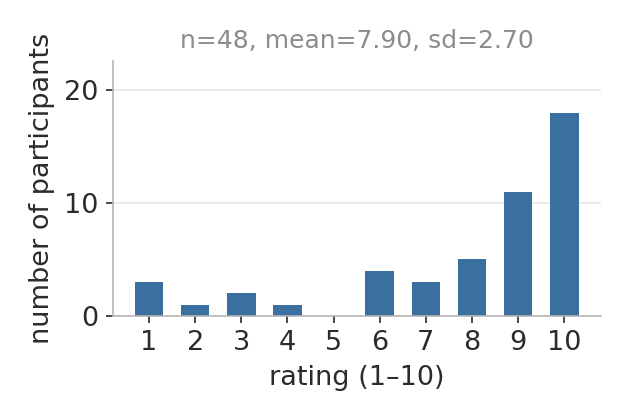}
\caption{\footnotesize Q18. I rarely use AI output without checking it.}
\end{subfigure}
\end{figure}

\begin{figure}[H]
\centering
\begin{subfigure}[t]{0.48\textwidth}
\centering
\includegraphics[width=\linewidth]{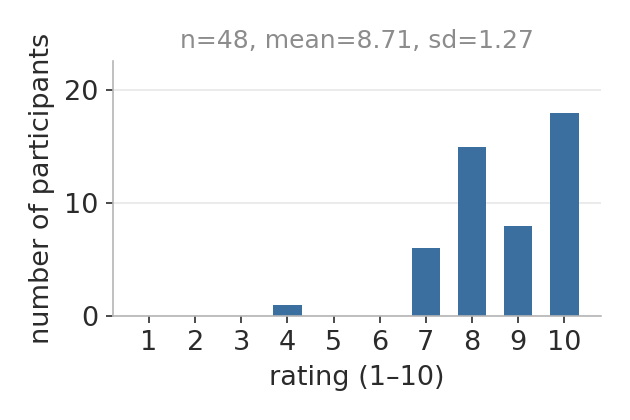}
\caption{\footnotesize Q19. I like understanding how AI reached its conclusions.}
\end{subfigure}
\hfill
\begin{subfigure}[t]{0.48\textwidth}
\centering
\includegraphics[width=\linewidth]{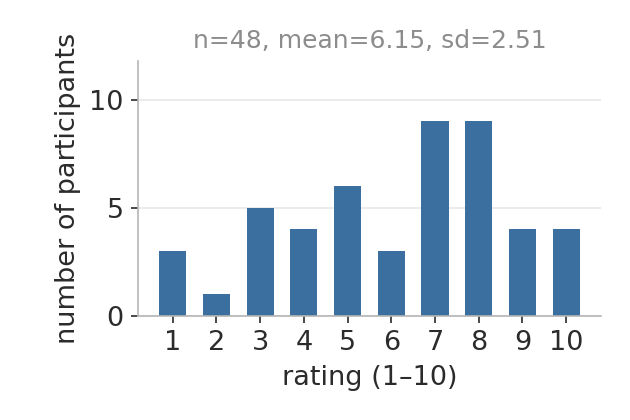}
\caption{\footnotesize Q20. If the answer looks reasonable, I usually move on without further verification.}
\end{subfigure}
\end{figure}
\clearpage
\subsubsection*{Control \& Redirection}
\begin{figure}[H]
\centering
\begin{subfigure}[t]{0.48\textwidth}
\centering
\includegraphics[width=\linewidth]{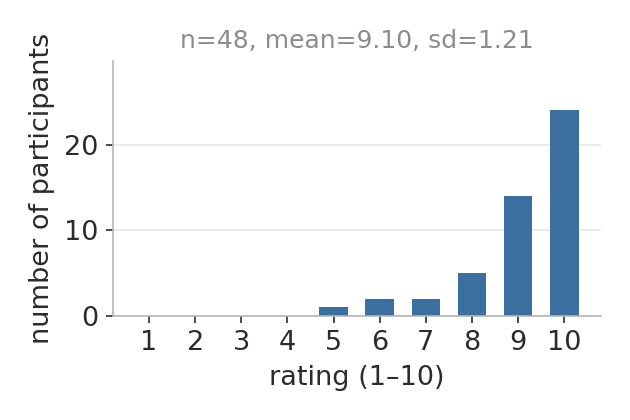}
\caption{\footnotesize Q21. I often redirect AI when it takes an unhelpful direction.}
\end{subfigure}
\hfill
\begin{subfigure}[t]{0.48\textwidth}
\centering
\includegraphics[width=\linewidth]{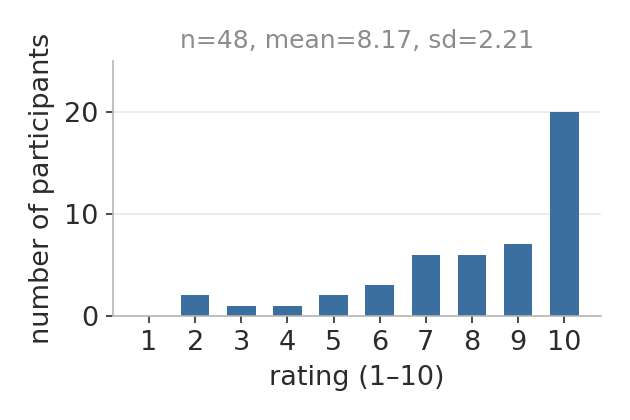}
\caption{\footnotesize Q22. I refine my requests multiple times until the AI matches my intent.}
\end{subfigure}
\end{figure}

\begin{figure}[H]
\centering
\begin{subfigure}[t]{0.48\textwidth}
\centering
\includegraphics[width=\linewidth]{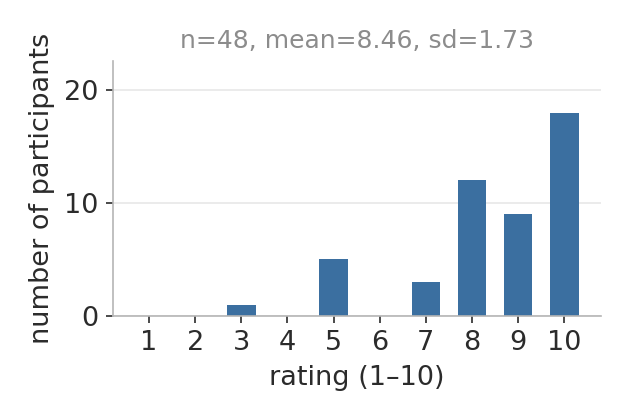}
\caption{\footnotesize Q23. I actively shape how the collaboration unfolds.}
\end{subfigure}
\hfill
\begin{subfigure}[t]{0.48\textwidth}
\centering
\includegraphics[width=\linewidth]{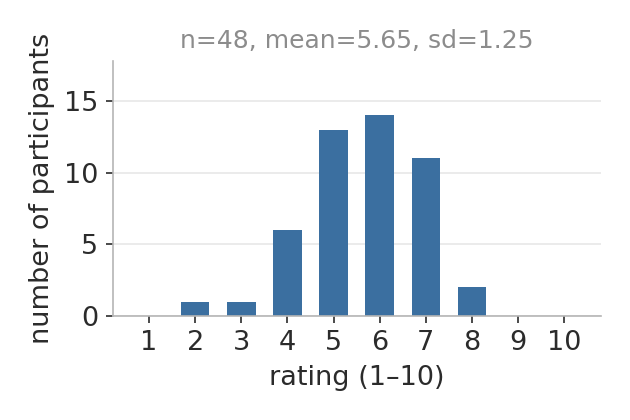}
\caption{\footnotesize Q24. I usually follow the direction suggested by the AI.}
\end{subfigure}
\end{figure}
\clearpage
\subsubsection*{Exploration of Alternatives}
\begin{figure}[H]
\centering
\begin{subfigure}[t]{0.48\textwidth}
\centering
\includegraphics[width=\linewidth]{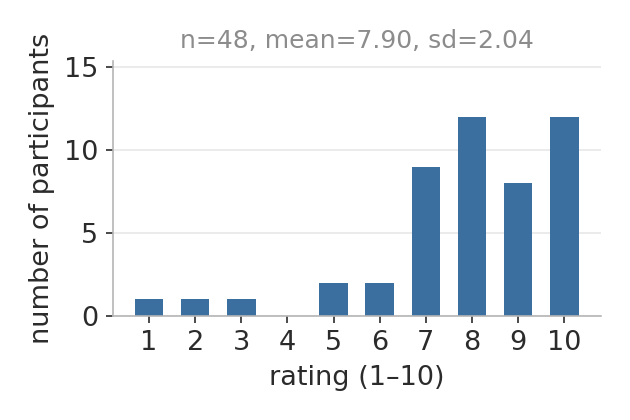}
\caption{\footnotesize Q25. I use AI to explore multiple perspectives before deciding.}
\end{subfigure}
\hfill
\begin{subfigure}[t]{0.48\textwidth}
\centering
\includegraphics[width=\linewidth]{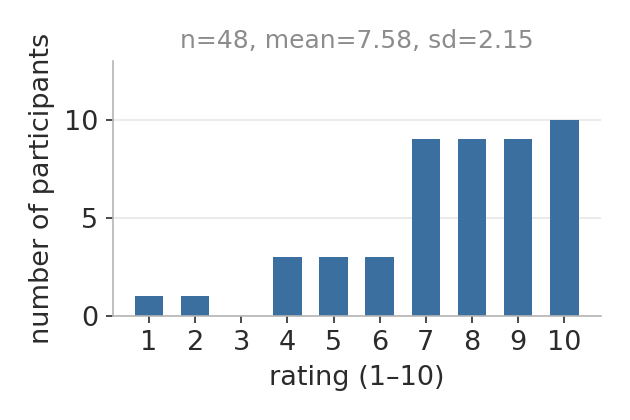}
\caption{\footnotesize Q26. I intentionally ask AI for alternative viewpoints.}
\end{subfigure}
\end{figure}

\begin{figure}[H]
\centering
\begin{subfigure}[t]{0.48\textwidth}
\centering
\includegraphics[width=\linewidth]{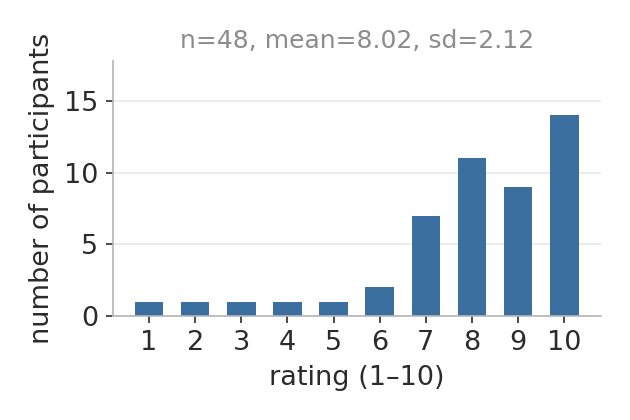}
\caption{\footnotesize Q27. I enjoy exploring ideas with AI before committing to one.}
\end{subfigure}
\hfill
\begin{subfigure}[t]{0.48\textwidth}
\centering
\includegraphics[width=\linewidth]{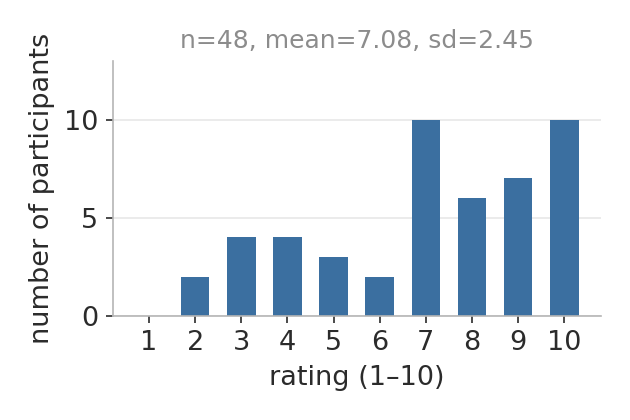}
\caption{\footnotesize Q28. I usually stop searching once I find one satisfactory answer.}
\end{subfigure}
\end{figure}
\clearpage
\subsubsection*{External / Cross-checking}
\begin{figure}[H]
\centering
\begin{subfigure}[t]{0.48\textwidth}
\centering
\includegraphics[width=\linewidth]{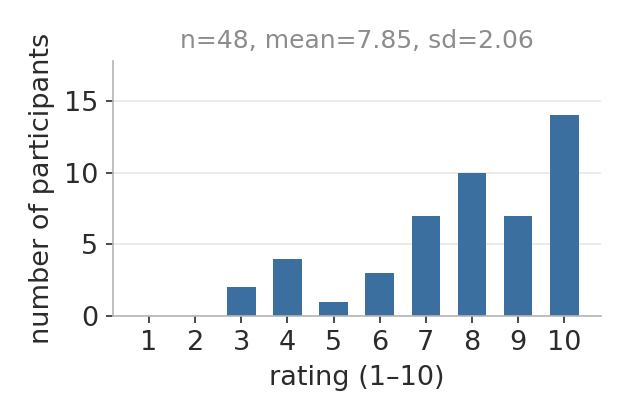}
\caption{\footnotesize Q29. I frequently combine information from AI with external documents.}
\end{subfigure}
\hfill
\begin{subfigure}[t]{0.48\textwidth}
\centering
\includegraphics[width=\linewidth]{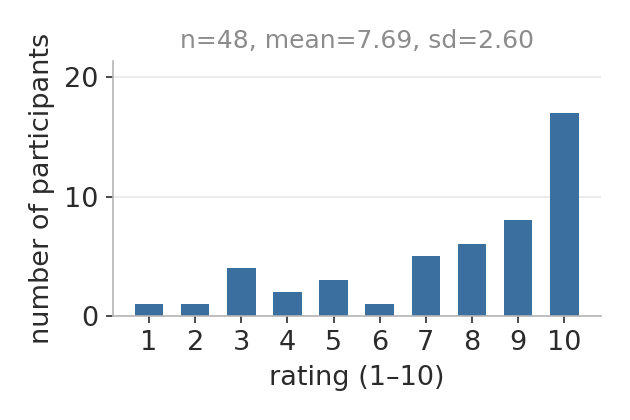}
\caption{\footnotesize Q30. I compare AI responses with academic papers or other trusted sources.}
\end{subfigure}
\end{figure}
\clearpage
\subsubsection*{Tool-Switching Behavior}
\begin{figure}[H]
\centering
\begin{subfigure}[t]{0.48\textwidth}
\centering
\includegraphics[width=\linewidth]{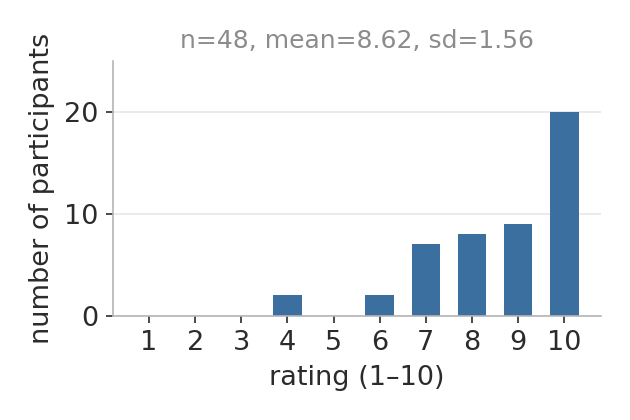}
\caption{\footnotesize Q31. My workflow often involves moving between AI and other tools.}
\end{subfigure}
\hfill
\begin{subfigure}[t]{0.48\textwidth}
\centering
\includegraphics[width=\linewidth]{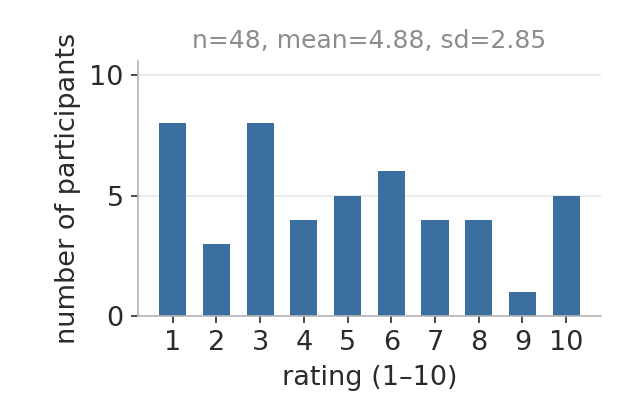}
\caption{\footnotesize Q32. Most of my work happens within a single AI chat.}
\end{subfigure}
\end{figure}
\clearpage
\subsubsection*{Planning Behavior}
\begin{figure}[H]
\centering
\begin{subfigure}[t]{0.48\textwidth}
\centering
\includegraphics[width=\linewidth]{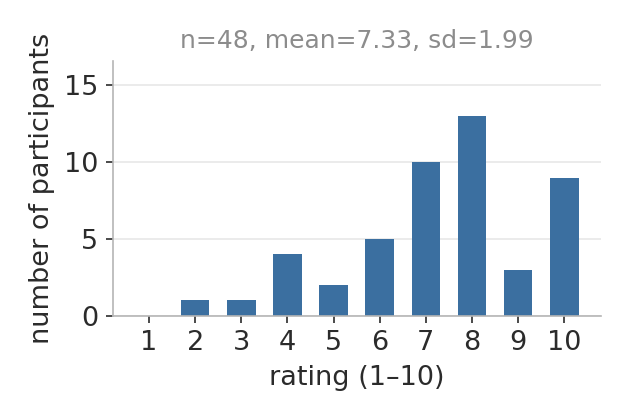}
\caption{\footnotesize Q33. Before using AI, I usually have a clear plan.}
\end{subfigure}
\hfill
\begin{subfigure}[t]{0.48\textwidth}
\centering
\includegraphics[width=\linewidth]{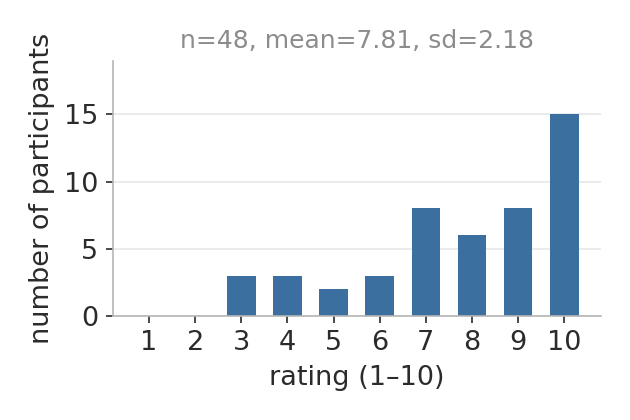}
\caption{\footnotesize Q34. I break complex tasks into smaller steps before asking AI.}
\end{subfigure}
\end{figure}

\begin{figure}[H]
\centering
\begin{subfigure}[t]{0.48\textwidth}
\centering
\includegraphics[width=\linewidth]{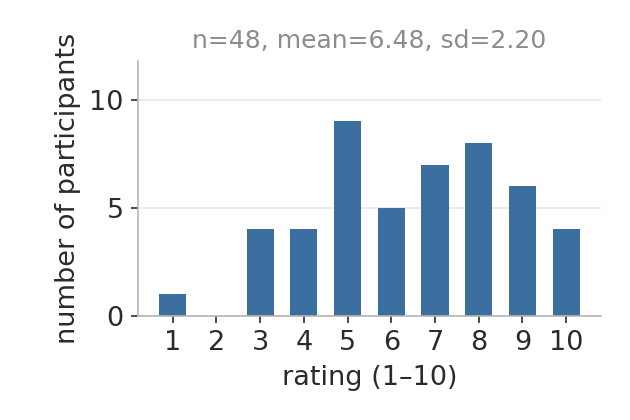}
\caption{\footnotesize Q35. I use AI to execute a plan rather than determine one.}
\end{subfigure}
\end{figure}
\clearpage
\subsubsection*{Reflection \& Reframing}
\begin{figure}[H]
\centering
\begin{subfigure}[t]{0.48\textwidth}
\centering
\includegraphics[width=\linewidth]{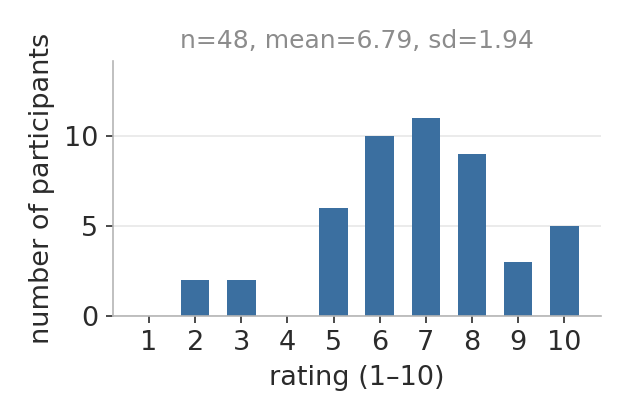}
\caption{\footnotesize Q36. I often reconsider my approach after interacting with AI.}
\end{subfigure}
\hfill
\begin{subfigure}[t]{0.48\textwidth}
\centering
\includegraphics[width=\linewidth]{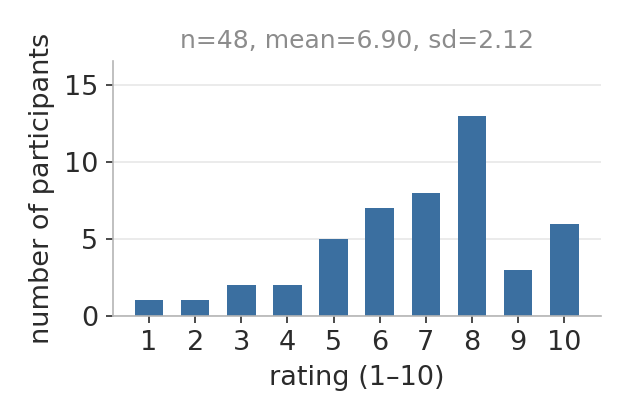}
\caption{\footnotesize Q37. Working with AI changes how I think about the problem.}
\end{subfigure}
\end{figure}

\begin{figure}[H]
\centering
\begin{subfigure}[t]{0.48\textwidth}
\centering
\includegraphics[width=\linewidth]{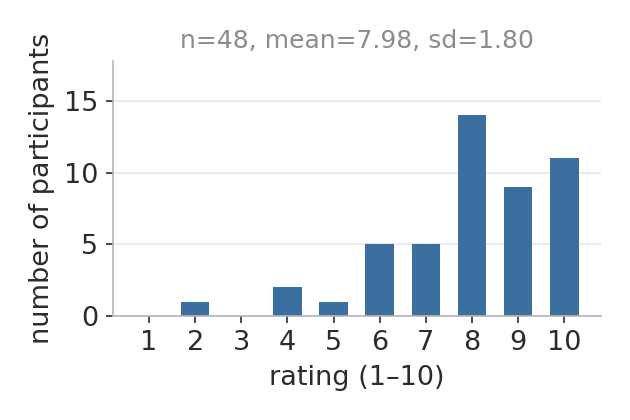}
\caption{\footnotesize Q38. AI helps me discover assumptions I hadn't considered.}
\end{subfigure}
\end{figure}
\clearpage
\subsubsection*{Error Cost \& Risk Perception}
\begin{figure}[H]
\centering
\begin{subfigure}[t]{0.48\textwidth}
\centering
\includegraphics[width=\linewidth]{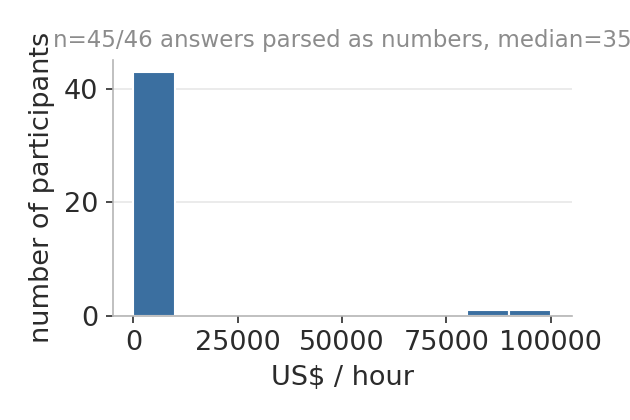}
\caption{\footnotesize Q39. What is the average hourly wage (in US Dollar) for a job similar to yours?}
\end{subfigure}
\end{figure}

\begin{figure}[H]
\centering
\includegraphics[width=0.75\linewidth]{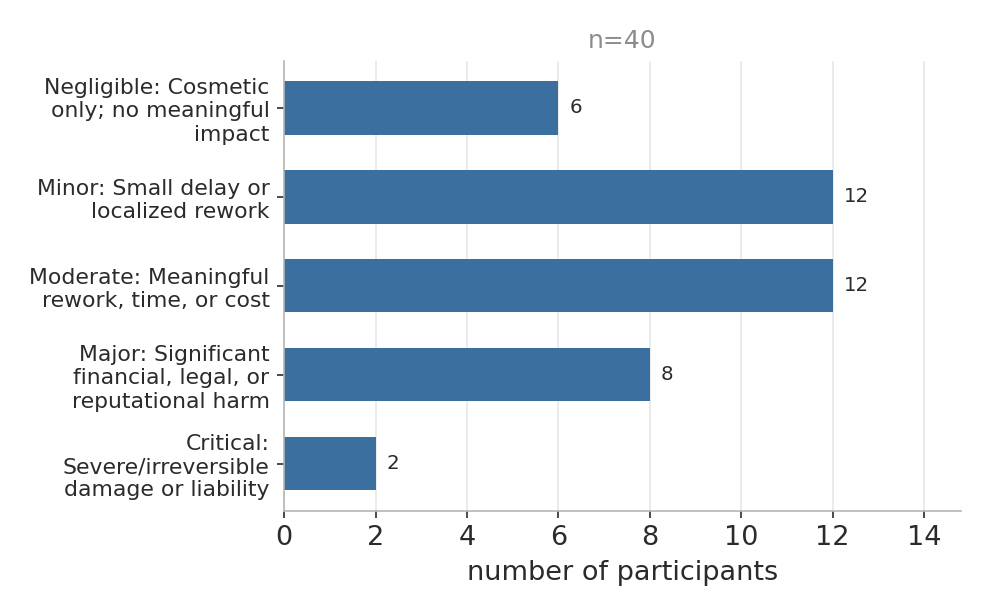}
\caption{Q40. How serious would an undetected material error be in the tasks you did for this study?}
\end{figure}

\begin{figure}[H]
\centering
\includegraphics[width=0.75\linewidth]{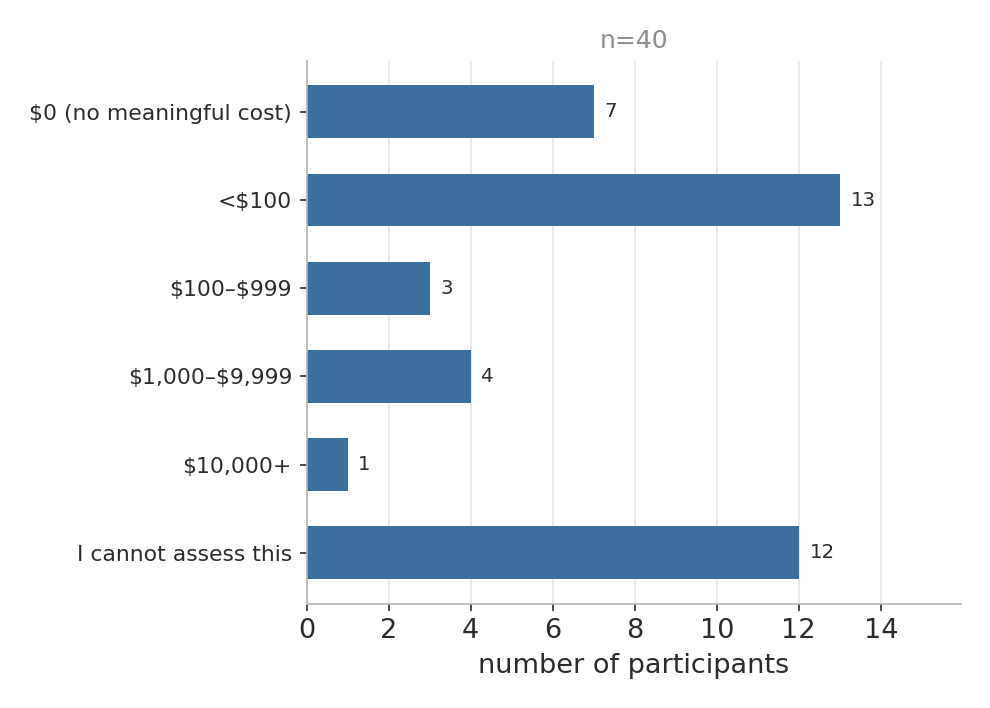}
\caption{Q41. Estimated financial/resource cost of an undetected error:}
\end{figure}

\begin{figure}[H]
\centering
\includegraphics[width=0.75\linewidth]{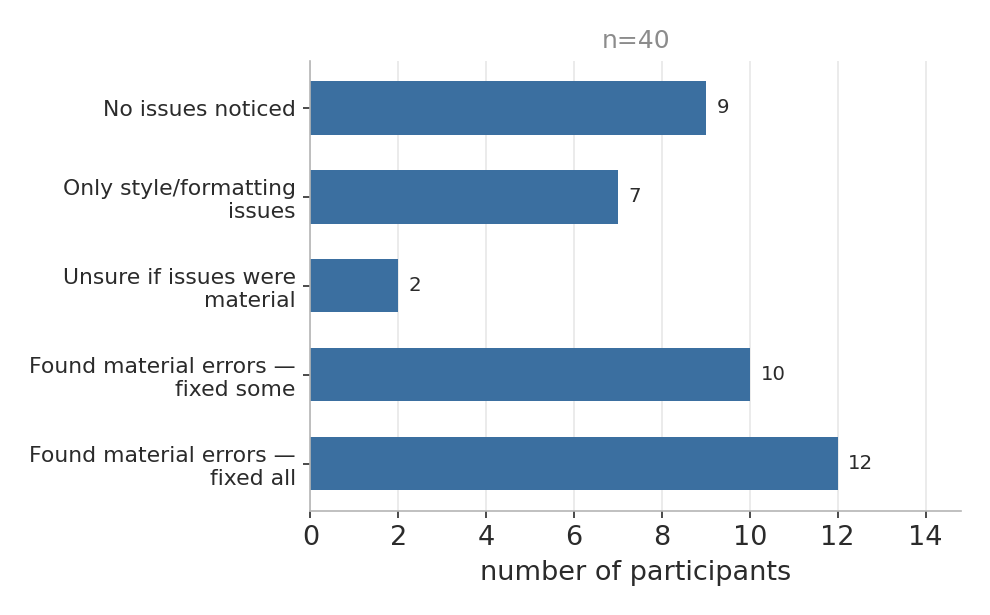}
\caption{Q42. Did you notice any issues in the AI content during the tasks you did for this study?}
\end{figure}

\begin{figure}[H]
\centering
\begin{subfigure}[t]{0.48\textwidth}
\centering
\includegraphics[width=\linewidth]{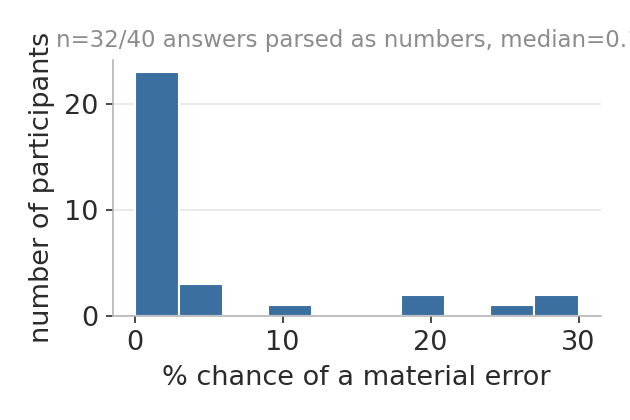}
\caption{\footnotesize Q43. Estimated chance your final output still contains a material error:}
\end{subfigure}
\end{figure}
\clearpage

\end{document}